\documentclass[10pt,twocolumn,letterpaper]{article}

\usepackage{cvpr}
\usepackage{times}
\usepackage{epsfig}
\usepackage{graphicx}
\usepackage{amsmath}
\usepackage{amssymb}
\usepackage{tabularx}
\usepackage{multirow}
\usepackage{caption}
\usepackage{subfig}
\usepackage{comment}
\usepackage{ctable}
\usepackage{boldline}
\usepackage{pifont}
\usepackage{tablefootnote}
\usepackage{colortbl}
\usepackage{hhline}
\usepackage{float}

\definecolor{viz_red}{RGB}{204, 0, 0}
\definecolor{viz_yellow}{RGB}{241, 194, 50}
\definecolor{viz_green}{RGB}{52, 183, 47}
\definecolor{viz_blue}{RGB}{17, 85, 204}
\definecolor{viz_purple}{RGB}{166, 77, 121}
\definecolor{hi_gray1}{gray}{0.9}
\definecolor{hi_yellow1}{RGB}{240, 240, 100}

\definecolor{hi_yellow2}{RGB}{250, 250, 150}

\newcommand{\uvd}{\textit{UVD}}
\newcommand{\xyz}{\textit{XYZ}}

\usepackage[pagebackref=true,breaklinks=true,colorlinks,bookmarks=false]{hyperref}

\cvprfinalcopy % *** Uncomment this line for the final submission

\def\cvprPaperID{5010} % *** Enter the CVPR Paper ID here

\ifcvprfinal\fi
\begin{document}

%%%%%%%%% TITLE
\title{KeyPose: Multi-View 3D Labeling and Keypoint Estimation \\ for Transparent Objects}

\author{Xingyu Liu$^1$\thanks{Work done as an intern at Google Research/Robotics at Google.} \qquad Rico Jonschkowski$^2$ \qquad Anelia Angelova$^2$ \qquad Kurt Konolige$^2$\\ $^1$Stanford University \qquad $^2$Robotics at Google}

\maketitle

% For some reason, this needs to be here, otherwise the first page is numbered.
\thispagestyle{empty}

%%%%%%%%% ABSTRACT
\begin{abstract}
Estimating the 3D pose of desktop objects is crucial for applications such as robotic manipulation. Many existing approaches to this problem require a depth map of the object for both training and prediction, which restricts them to opaque, lambertian objects that produce good returns in an RGBD sensor. In this paper we forgo using a depth sensor in favor of raw stereo input. We address two problems: first, we establish an easy method for capturing  and labeling 3D keypoints on desktop objects with an RGB camera; and second, we develop a deep neural network, called \emph{KeyPose}, that learns to accurately predict object poses using 3D keypoints, from stereo input, and works even for transparent objects.  To evaluate the performance of our method, we create a dataset of 15 clear objects in five classes, with 48K 3D-keypoint labeled images.  We train both instance and category models, and show generalization to new textures, poses, and objects. KeyPose surpasses state-of-the-art performance in 3D pose estimation on this dataset by factors of 1.5 to 3.5, even in cases where the competing method is provided with ground-truth depth. Stereo input is essential for this performance as it improves results compared to using monocular input by a factor of 2. We will release a public version of the data capture and labeling pipeline, the transparent object database, and the KeyPose models and evaluation code.  Project website: \url{https://sites.google.com/corp/view/keypose}.
\end{abstract}

%%%%%%%%% BODY TEXT
\section{Introduction}

% \newpage

\begin{figure}[!ht] 
  \subfloat{% 
    \includegraphics[height=0.335\linewidth]{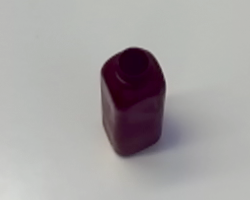} 
  } 
  \hspace{3.5ex} 
  \subfloat{% 
    \includegraphics[height=0.335\linewidth]{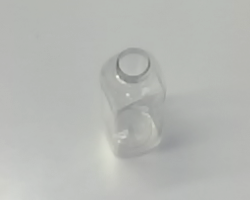} 
  }
  \\ [-1.6ex]
  \subfloat{% 
    \includegraphics[height=0.34\linewidth]{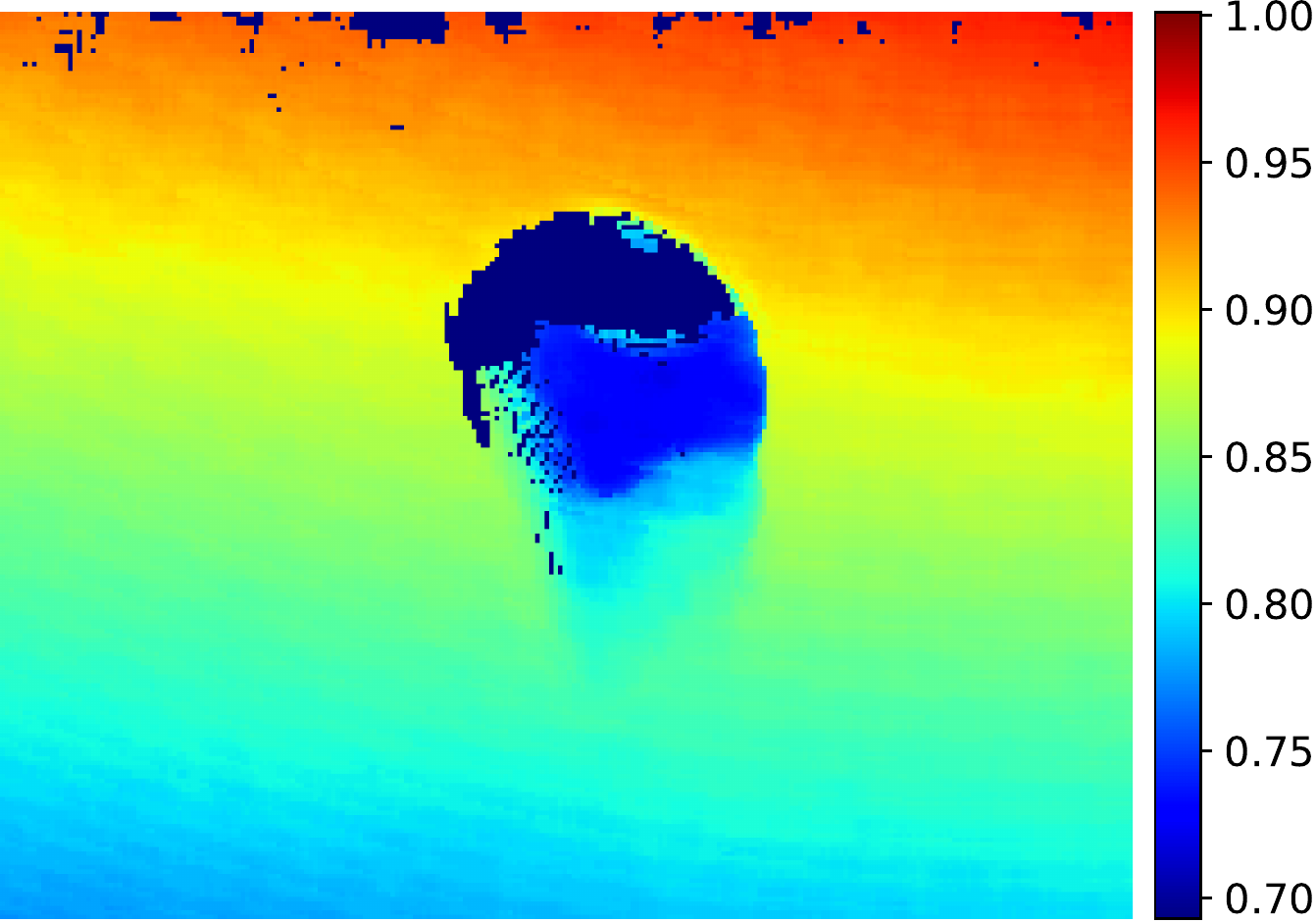} 
  }
  \hfill 
  \subfloat{% 
    \includegraphics[height=0.34\linewidth]{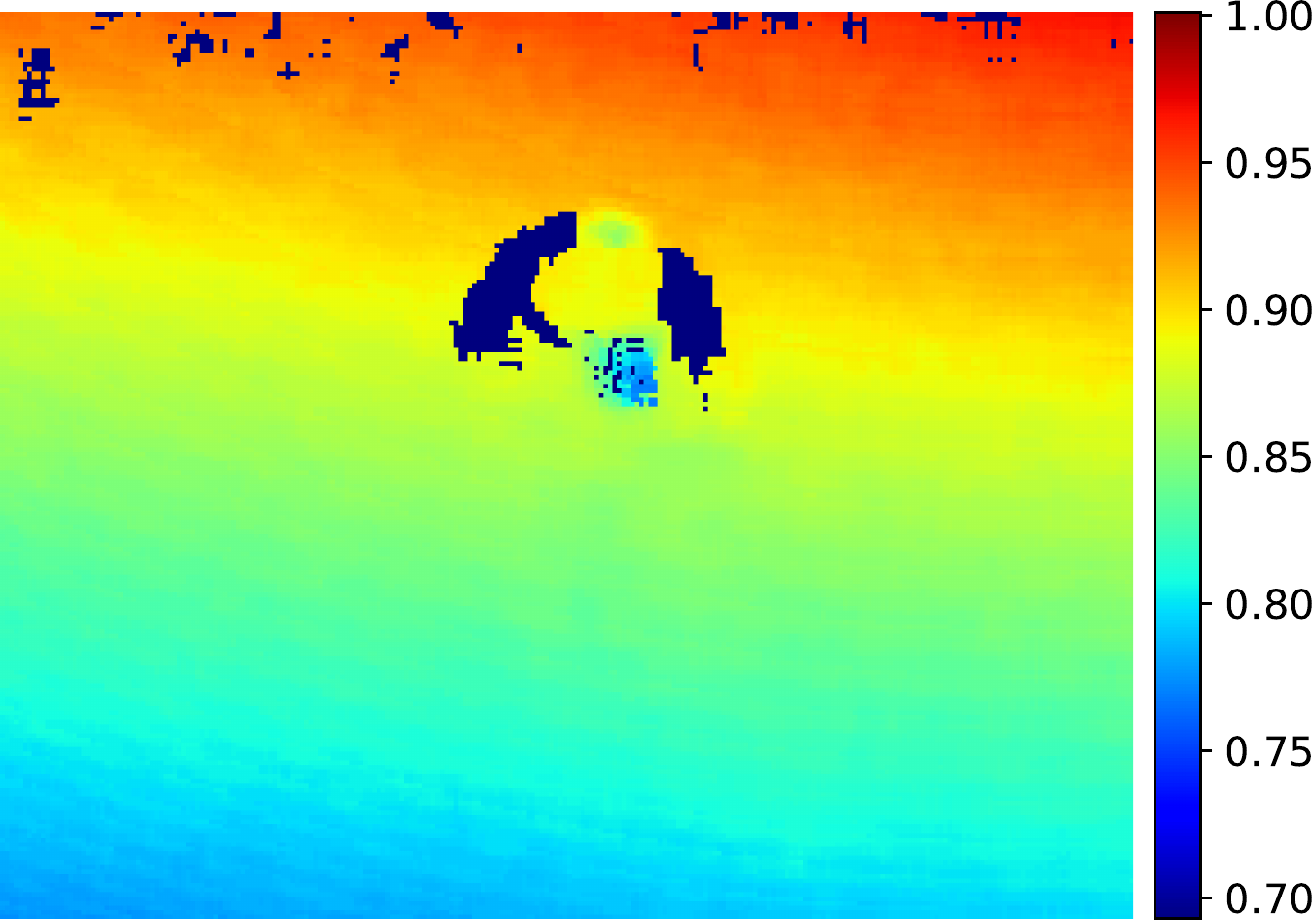} 
  }
  \\ [-1.6ex]
  \subfloat{% 
    \includegraphics[height=0.24\linewidth]{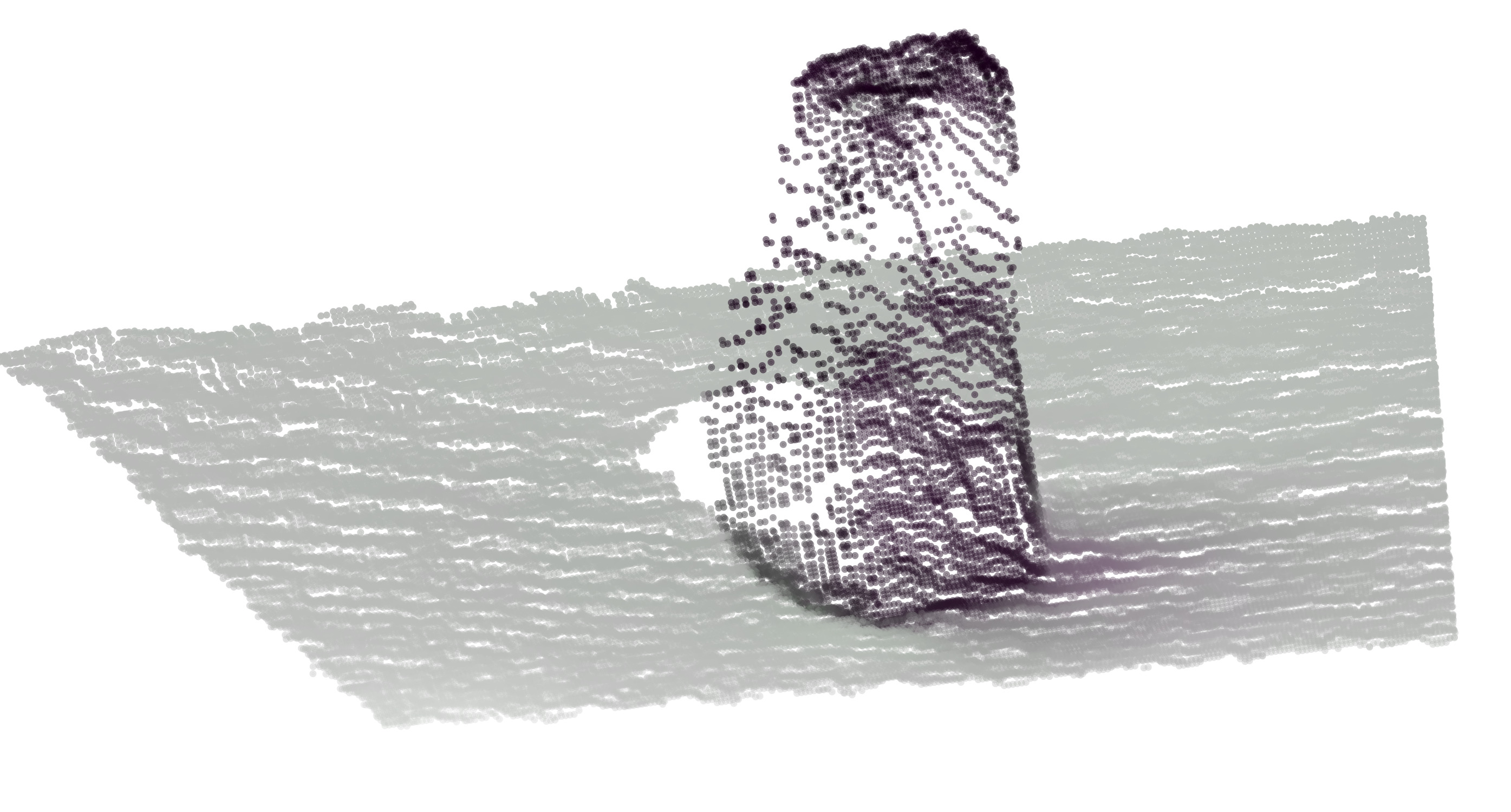} 
  }
  \hspace{1.7ex} 
  \subfloat{% 
    \includegraphics[height=0.24\linewidth]{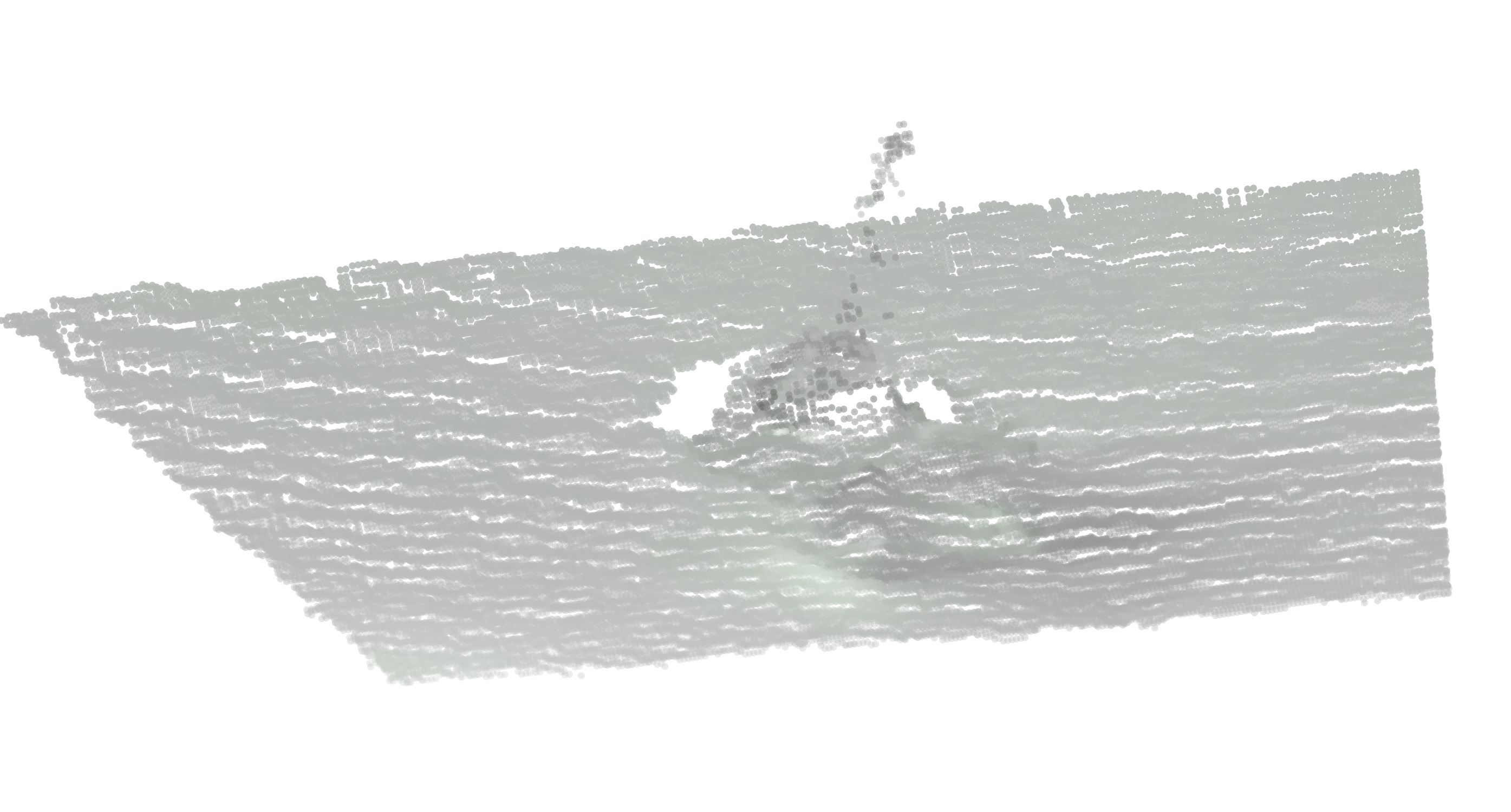} 
  } 
  \caption{RGB image (top), depth map (middle), and point cloud (bottom) of an opaque bottle (left) and its transparent twin (right). The opaque bottle returns reasonable depth while the transparent one returns invalid depth values using a Microsoft Azure Kinect sensor.} 
  \label{fig:teaser}
  % \vspace{-1ex}
\end{figure}

\newcolumntype{P}[1]{>{\centering\arraybackslash}p{#1}}

Estimating the position and orientation of 3D objects is one of the core problems in computer vision
applications that involve object-level perception such as augmented reality (AR) and robotic manipulation. 
Rigid objects with a known model can be described by 4D pose (e.g., vehicles  \cite{FaF,pointpillars}), 6D pose \cite{posecnn,poserbpf}, and 9D pose where scale is predicted \cite{nocs}.  A more flexible method uses 3D \emph{keypoints} \cite{labelfusion,keypointnet}, which can handle articulated and deformable objects such as the human hand or body \cite{integral:net,Li2019}. While some of these methods predict 3D keypoints from a single RGB image, others use RGBD data collected by a depth sensor \cite{densefusion,labelfusion,ycb} to achieve better accuracy.
Unfortunately, existing commercial depth sensors, such as projected light or time-of-flight (ToF) sensors, assume that objects have opaque, lambertian surfaces that can support diffuse reflection from the sensor. Depth sensing fails when these conditions do not hold, e.g., for \emph{transparent} or \emph{shiny metallic} objects. Figure \ref{fig:teaser} shows such an example.

In this paper, we present the first method of \textbf{keypoint-based pose estimation for (transparent) 3D objects from stereo RGB images}.
There are several challenges:
first, there is no available large-scale dataset for transparent 3D object pose estimation from stereo images with annotated keypoints. 
Datasets such as NYUDepth v2 \cite{nyudepthv2} lack annotations for precise pose of each individual objects, while other datasets such as LabelFusion \cite{labelfusion}, YCB dataset \cite{ycb} and REAL275 \cite{nocs} annotate monocular RGBD images of opaque objects.
The second challenge is the annotation of pose of transparent 3D objects. Existing datasets such as \cite{labelfusion,ycb,nocs} require accurate depth information as well as an object CAD model so that alignment algorithms such as iterative closest point (ICP) \cite{icp} can be applied. The third challenge is how to leverage only RGB images for 3D keypoint estimation, thus obviating the need for a depth sensor. 

To address the challenges regarding data acquisition and annotation, we introduce an efficient method of capturing and labeling stereo RGB images for transparent (and other) objects.  Although our method does not need them, we also capture and register depth maps of the object, for both the transparent object and its opaque twin, registered with the stereo images; we use a robotic arm to help automate this process. The registered opaque depth allows us to compare to methods that require depth maps as input such as DenseFusion \cite{densefusion}. Following the proposed data capturing and labeling method, we constructed a large dataset consisting of 48k images from 15 transparent object instances.  We call this dataset TOD (Transparent Object Dataset).

To reduce the requirement on reliable depth, we propose a deep model, KeyPose, that predicts 3D keypoints on transparent objects from cropped stereo RGB input. The crops are obtained from a detection stage that we assume can loosely bound objects (see \cite{cleargrasp} for an appropriate method for transparent objects). The model determines depth implicitly by combining information from the image pair, and predicting the 3D positions of keypoints for object instances and classes. After training on TOD, we compare KeyPose to the best existing RGB and RGBD methods and find that it vastly outperforms them on this dataset.  In summary, we make the following contributions:
\begin{itemize}
\setlength{\itemsep}{1pt}
\setlength{\parskip}{0pt}
\setlength{\parsep}{0pt}
\item 
A pipeline to label 3D keypoints on real-world objects, including transparent objects that does not require depth images, thus making learning-based 3D estimation of previously unknown objects possible without simulation data or accurate depth images. This pipeline supports a twin-opaque technique to enable comparison with models that require depth input.
\item A dataset of 15 transparent objects in 6 classes, labeled with relevant 3D keypoints, and comprising 48k stereo and RGBD images with both transparent and opaque depth. This dataset can also be used in other transparent 3D object applications.
\item A deep model, KeyPose, that predicts 3D keypoints on these objects with high accuracy using RGB stereo input only, and even outperforms methods which use ground-truth depth input.
\end{itemize}

\section{Related Work}

\textbf{4D/6D/9D Pose Representation. } The assumption behind these representations is the rigidity of the object, so that translation, rotation and size is sufficient to describe its configuration.
Existing techniques for 4D/6D/9D pose estimation can generally be categorized by whether a 3D CAD model is used in training or inference.
The first type of technique aligns the observed RGB images with rendered CAD model images \cite{poserbpf, deepim}, or aligns the observed 3D point clouds with 3D CAD model point clouds with algorithms such as ICP \cite{densefusion}, or renders mixed reality data from 3D CAD models as additional training data \cite{nocs}.
While it is possible to render high-quality RGB scenes of transparent objects using ray-tracing, there has been no work done on rendering depth images that faithfully reproduces the degraded depth seen in real-world RGBD data (see Figure \ref{fig:teaser}).

The second type of technique regresses the object coordinate values from the RGB image or 3D point clouds
\cite{posecnn, FaF, pointpillars, frustum:pointnet, bb8}.
Our method does not assume object rigidity, and the object pose is based the locations of 3D keypoints, which can be used on articulated or deformable objects.  Our method also does not rely on prior knowledge about each individual object, such as a 3D CAD model.

\textbf{Keypoint Based Pose Representation. } 
Previous work has explored deep learning methods for detecting 3D keypoints of an object given a monocular RGB image \cite{keypointnet} or RGBD image \cite{kpam}.
The core is to predict probability maps for the 2D keypoint locations, and then use the given or predicted depth image for 3D.
Other works proposed similar methods for monocular pose estimation \cite{integral:net,monocular:3d:human:pose:estimation,H+O}.  Though estimating 3D positions from a single RGB image is an ill-conditioned problem, these methods implicitly learn the prior of object size during training, or rely on the known object 3D model.
Our method is inspired by these works and focuses on 3D keypoint location estimation from stereo instead of single images, and is well-conditioned even for similar objects that vary in scale.  Recently, a method similar to ours was proposed for hand tracking using raw strereo \cite{Li2019}.  For rigid objects with a known model, the 6D pose can be recovered using the Procrustes algorithm (see the Supplementary materials).

\textbf{Stereo for Disparity Estimation. }
Estimating disparity and therefore depth from stereo has been a long-standing problem in computer vision. 
The success of deep-learning methods in computer vision inspired research in this area, using end-to-end deep networks equipped with a correlation cost volume \cite{dispnet, end:to:end:stereo:regression, amnet, segstereo}, or point-based depth representation and iterative refinement \cite{point:mvsnet}.
Here, instead of generating a dense disparity field, we focus on estimating the 3D location of sparse keypoints directly from stereo images.

\textbf{3D Object Pose Estimation Datasets. } Directly labeling 3D object pose in real RGB images is costly. 
All existing real (non-synthetic) datasets for 3D object pose estimation rely on capturing RGBD images and annotating pose by either constructing a 3D mesh \cite{kpam}, or fitting 3D CAD models to 3D point clouds \cite{labelfusion, ycb, nocs, linemod}, neither of which is possible for transparent objects.
On the contrary, we build a data capturing pipeline where ground-truth depth of transparent object keypoints
can be efficiently obtained, without relying on depth or 3D CAD models.

\textbf{Estimation of transparent and reflective objects. } Objects that are transparent or reflective present significant challenges for all  
camera-based depth estimation. 
Works on estimating transparent object pose and geometry might assume knowing object 3D model \cite{seeing:glassware, pose:estimation:transparent:clutter} or rely on synthetic data to train vision models \cite{saxena2008robotic, cleargrasp}.
Our data capturing and labeling enables generating large-scale real dataset for training and testing transparent object pose and geometry, so synthetic data are not needed.

\begin{figure}[t] 
  \centering
    \includegraphics[width=\linewidth]{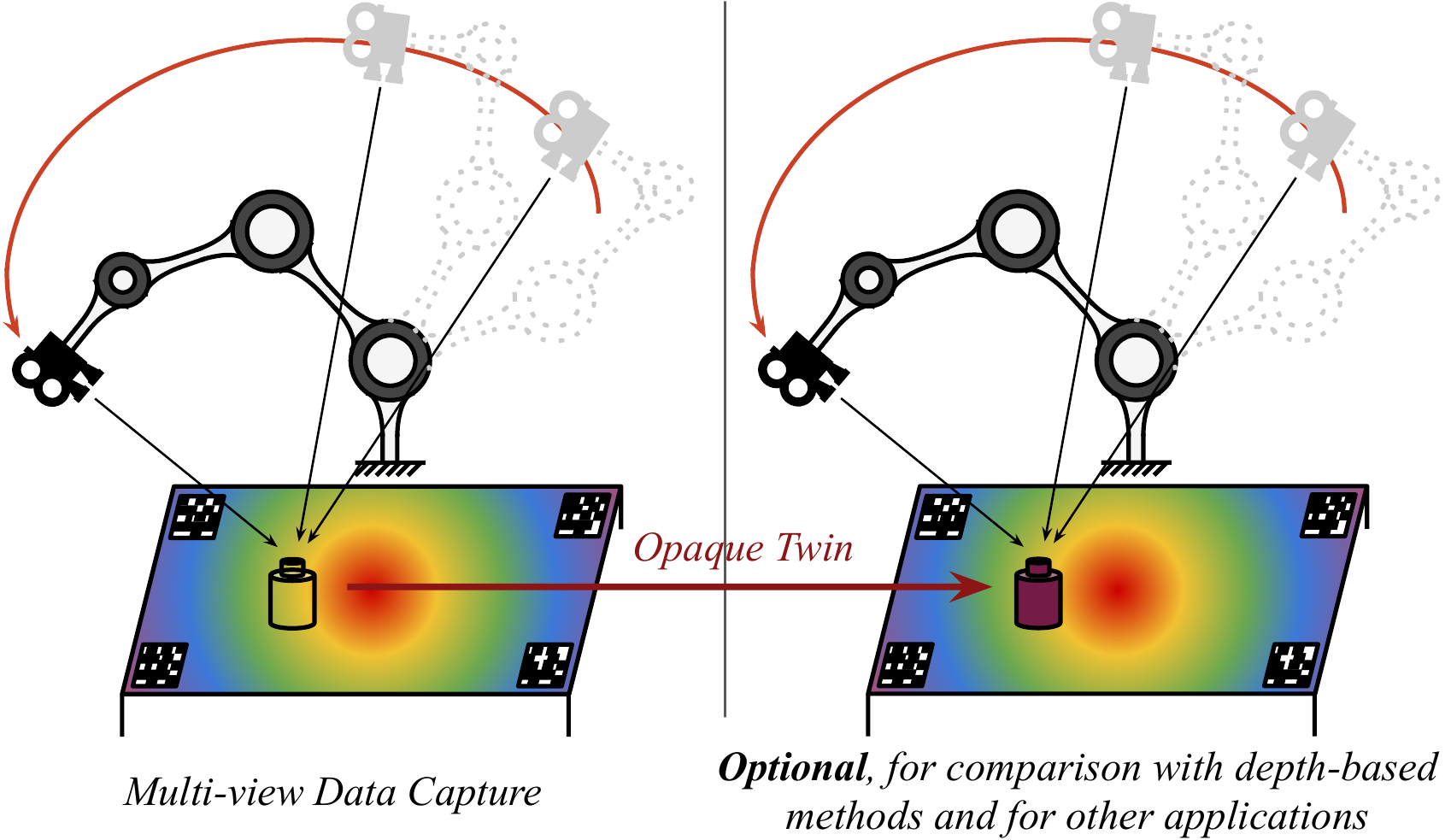} 
  \caption{Data capturing pipeline.  We mount both the stereo RGB camera and RGBD camera on the end-effector of the robot. We then use the robot arm to perform similar paths to scan both the opaque lambertian object (left) and its transparent twin placed at the same location of a textured surface (right). AprilTags~\cite{apriltag2} are used as global pose indicator for the cameras.
  }
  \label{fig:data:capture}
\end{figure}

\section{Transparent Object Dataset (TOD)}

In this section, we describe the data capturing pipeline that enables efficient capture and labeling of 3D keypoints for a large number of samples without requiring a depth sensor.

\subsection{Data Collection with a Robot}

Hand-labeling 3D keypoints in individual RGB images is difficult or impossible due to uncertainty about keypoint depth.  Instead, we leverage multi-view geometry to raise 2D labels from a small number of images into 3D labels for a set of images where the object has not moved.  The general idea is illustrated in Figure \ref{fig:data:capture}.

\begin{figure}[t] 
\centering
  \subfloat{% 
    \includegraphics[width=0.33\linewidth]{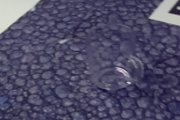} 
  } 
  \subfloat{% 
    \includegraphics[width=0.33\linewidth]{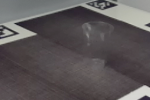} 
  } 
  \subfloat{% 
    \includegraphics[width=0.33\linewidth]{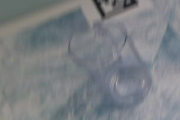} 
  } 
  \vspace{-1ex}
  \caption{Challenging cases in our dataset, including dark background textures (left), thin handles of mugs (middle) and motion blur (right). Accurately locating these objects is a difficult task even for human.} 
  \label{fig:challenging:cases}
  % \vspace{-1ex}
\end{figure}

We use a stereo camera with known parameters to capture images in a sequence, moving the camera with a robot arm (we could also move it by hand).
To estimate the pose of the camera relative to the world, we set up a planar form with AprilTags \cite{apriltag2} that can be be recognized in an image, and from their known locations estimate the camera pose.  From a small subset of widely-separated poses, we label 2D keypoints on the object. Optimization from multi-view geometry gives the 3D position of the keypoints, which can be reprojected to all images in the sequence.
To increase diversity, we place various textures under the object.  Figure \ref{fig:challenging:cases} shows some challenging data examples.

The resultant labeled stereo samples are sufficient to train and evaluate the KeyPose model.  We can collect and label data for a new object in a few hours. 
In addition to the stereo data, we also capture and register depth data using the Microsoft Kinect Azure RGBD device.  This data is purely ancillary to our model, but it lets us compare KeyPose to methods that require depth data.  We collect two depth images, one during the initial scan with co-mounted stereo and RGBD devices, and one with the transparent object replaced by its opaque (painted) twin during a second scan (Figure \ref{fig:data:capture}, right).  Although the RGBD images are captured at slightly different poses from the stereo (due to variations in the trajectories and camera capture times), we can leverage the calculated pose of the RGBD camera (using AprilTags in the RGB image), and the known offset of the depth sensor from the RGB sensor, to warp the depth image to align precisely with the left stereo image (see Figure \ref{fig:teaser}).

\subsection{Keypoint Labeling and Automatic Propagation}

To accurately construct this dataset, we need to address different sources of error. First, since AprilTag detection is imperfect in finding tag positions, we spread out these tags on the target to produce large baselines for camera pose estimation. Second, since human labeling of keypoints on 2D images introduces error, we use a farthest-point algorithm on the camera poses to ensure that annotated images used in going from 2D to 3D have a large baseline. 

We want to know the accuracy of the manual annotation.  While the absolute ground truth of the 3D keypoints is unknown, we can estimate the labeling error, given the known reprojection errors of the AprilTags and 2D annotations.  Using a Monte Carlo simulation based on the reprojection errors, we calculate the random error of the labeled 3D keypoints to be around 3.4 mm RMSE, which is quite accurate.  Details of the simulation are in the supplementary material.

\section{Predicting 3D Keypoints from RGB Stereo}

In this section, we describe the KeyPose method of estimating the pose of 3D objects from stereo input, using supervised training of 3D keypoints. 
We first introduce patching cropping from bounding box and then describe our CNN architecture. Finally, we present the choice of loss functions used in training, which significantly affect the performance.

\subsection{Data Input to the Training Process}

\begin{figure}[t] 
  \centering
\includegraphics[width=0.9\linewidth]{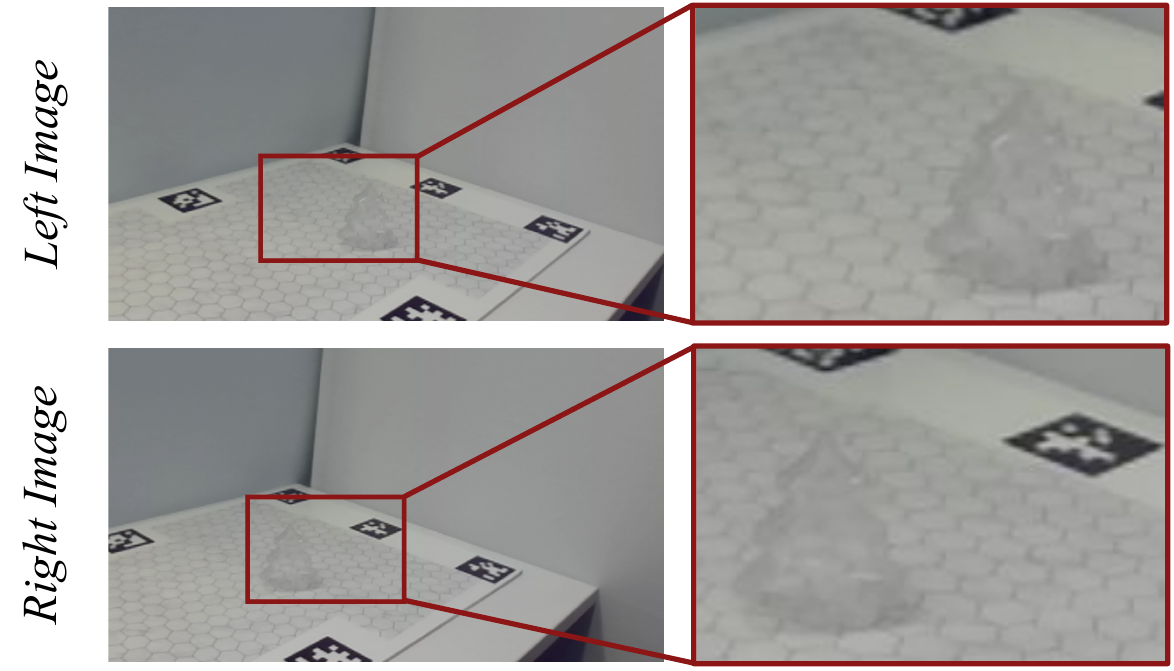}
\vspace{-1ex}
\caption{Example of cropping with bounding box for left and right images.}
\vspace{-1ex}
\label{fig:cropping:left:right}
\end{figure}

We assume a detection stage that approximately determines the location of an object (see \cite{cleargrasp} for a method to detect and segment transparent objects; or, the \textit{UV} heatmap of Figure \ref{fig:early} could be used).  From this bounding box we crop a fixed-size rectangle from the left image, and a corresponding rectangle at the same height from the right image, preserving the epipolar geometry (Figure \ref{fig:cropping:left:right}).

Since the right object image is offset from the left -- in our case, by 48 to 96 pixels, given the stereo device and assuming an object distance from 0.5m to 1m -- the rectangle must extend far enough to encompass the right object no matter where it might appear.  To limit the rectangle extension, we offset the right crop horizontally by 30 pixels, changing the apparent disparity to 18-66 pixels.  The input size for each crop is 180$\times$120 pixels.

The input images are processed by the model to produce, for each keypoint, a UV (2D) image location of the keypoint and a disparity D that encodes depth and is the offset of the left and right keypoints (in pixels).  The \uvd\ triplet encodes the 3D \xyz\ coordinates by:
$Q := \uvd \mapsto \xyz$, where $Q$ is a \textit{reprojection matrix} determined by the camera parameters \cite{opencv:reproj}. We use these \xyz\ positions as labels to generate training errors, by projecting back to the camera image and comparing \uvd\ differences.  Reprojected pixel errors are a stable, physically-realizable error method widely used in multiple-view geometry \cite{kendall:cvpr2017}.  Comparing 3D errors directly introduces a large bias, as they grow quadratically with distance, overwhelming the errors of closer objects.

To encourage generalization, we perform geometric and photometric augmentation of the input images.  More details are in the supplementary material. Note that geometric augmentation must be limited to transformations that do not violate epipolar constraints, i.e. scaling, $Y$-axis shear, mirroring, and rotation of the view around the $X$-axis.

\subsection{Architecture for 3D Pose Estimation }

The KeyPose model combines the following principles:
\begin{description}
\setlength{\itemsep}{0pt}
\setlength{\parskip}{0pt}
\setlength{\parsep}{0pt}
    \item[Stereo for Implicit Depth.] Use stereo images to introduce depth information to the model.
    \item[Early Fusion.] Combine information from the two image crops as early as possible.  Let the deep neural network determine disparity implicitly, rather than forming explicit correlations (as in \cite{amnet}).
    \item[Broad Context.] Extend the spatial context of each keypoint as broadly as possible, to take advantage of any relevant shape information of the object.
\end{description}

\begin{figure}[t]
\centering
\includegraphics[width=\linewidth]{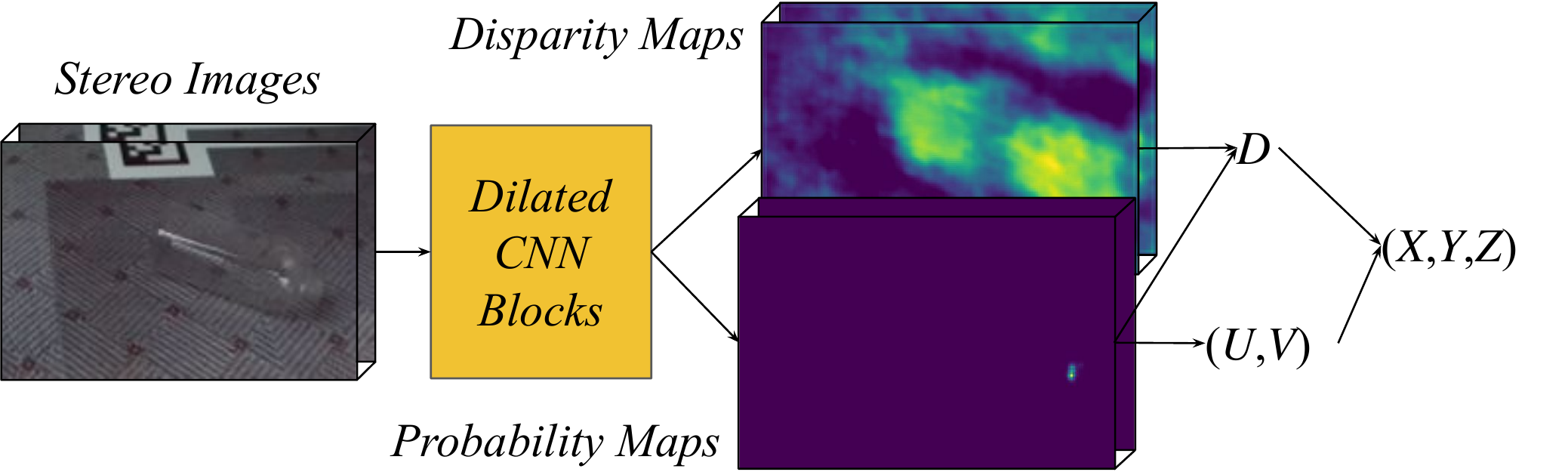}
\caption{Early fusion architecture.}
\label{fig:early}
\vspace{1ex}
\centering
\includegraphics[width=\linewidth]{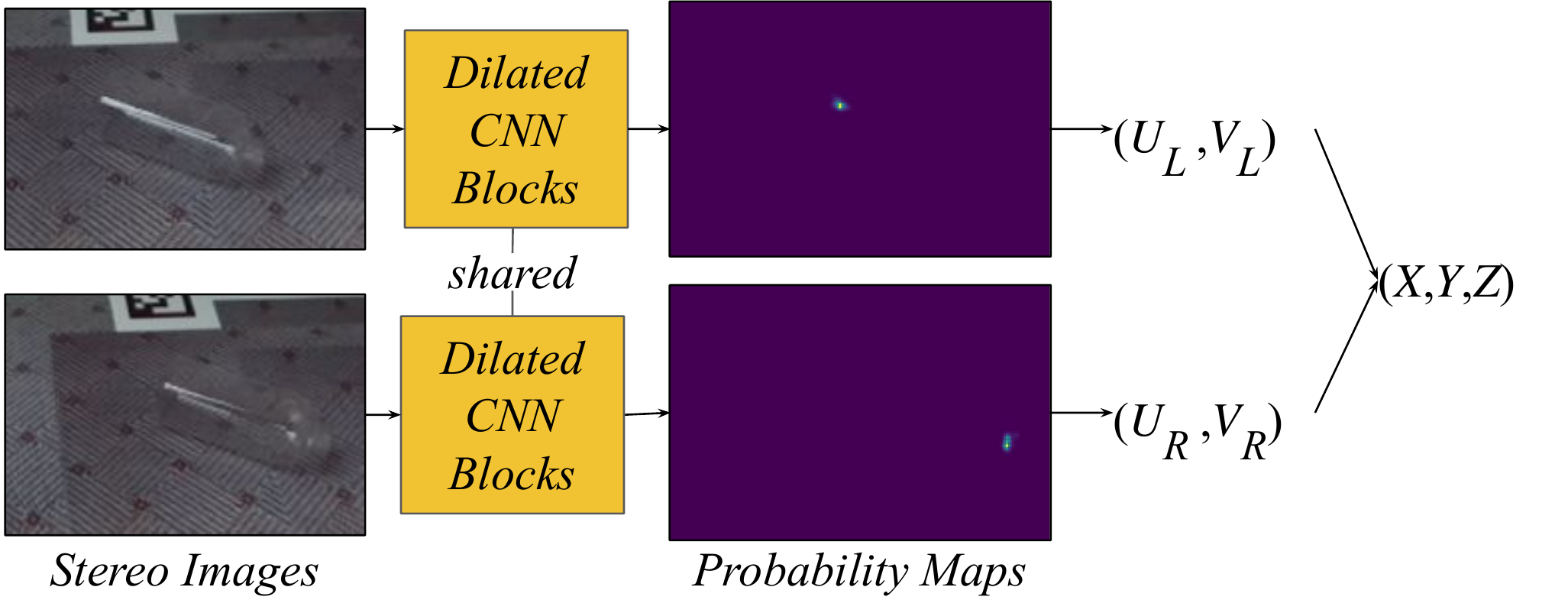}
\caption{Late fusion architecture.}
\label{fig:late}
\vspace{-1ex}
\end{figure}

\noindent Figure \ref{fig:early} shows the basic structure of the model, which was adapted from \cite{keypointnet}.  Stereo images are stacked and fed into a set of exponentially-dilated 3x3 convolutions \cite{dcnn} that expands the context for predicting keypoints, while keeping the resolution constant.  Two such groupings ensure that the context for each keypoint is thoroughly mixed.  The number of features is kept constant at 48 (for instance models) and 64 (for category models) throughout the CNN blocks.  After this, \textit{projection heads}, one per keypoint, extract \uvd\ coordinates.  We investigate two projection methods:
\begin{enumerate}
    \item Direct regression.  Three 1x1 convolutional layers produce $N \times 3$ numeric \uvd\ coordinates, where $N$ is the number of keypoints.
    \item Heatmaps.  For each keypoint $i$, a CNN layer produces a heatmap, followed by spatial softmax to generate a probability map $\mathit{prob_i}$, and then integrated to get \textit{UV} coordinates, as in IntegralNet \cite{integral:net}.  A disparity heatmap is also computed, convolved with the probability map, and integrated to produce disparity (Figure \ref{fig:early}).  This method is also useful for visualization.
\end{enumerate}

To test the efficacy of early fusion, we also implemented a \emph{late fusion} model (Figure \ref{fig:late}), in which siamese dilated CNN blocks separately predict \textit{UV} keypoints for the left and right images.  Then standard stereo geometry is used to generate a 3D keypoint prediction.

\subsection{Losses}
\newcommand{\UVD}{\mathit{UVD}}
\newcommand{\La}{{\mathcal{L}}}
We use three losses: a direct keypoint \uvd\  loss, a projection loss, and a locality loss.  We also permute the total loss and take a minimum for symmetric keypoints.

    \noindent \textbf{Keypoint loss.}  The predicted ($\UVD$) and labeled ($\UVD ^*$) pixel values are compared via squared error
    \begin{equation}
        \La_{kp} = \sum_{i \in \mathit{kps}} \| \UVD_i - \UVD^*_i  \| ^ 2
    \end{equation}
    We tried a direct 3D loss, but the errors grow quadratically with distance, overwhelming the errors of closer objects. This introduces a large bias in the model performance.\\
    \noindent\textbf{Projection loss.} Predicted \uvd\ values are converted to a 3D point, then re-projected to the widely-separated views that were used to create the 3D points.  The difference between the predicted and labeled \textit{UV} re-projections is squared for the loss.  Let $P_j$ be the projection function, and $Q := \UVD \mapsto \mathit{XYZ}$.  Then
     \begin{equation}
        \La_\mathit{proj} = \sum_{i \in \mathit{keypts}} \sum_{j \in \mathit{views}}
        \| P_j Q(\UVD_i) - P_j Q(\UVD^*_i)  \| ^ 2
     \end{equation}   
    In the same way that the wide viewpoints pinpoint the 3D coordinates of a keypoint in generating labels, here they recreate the same conditions for constraining the predicted keypoint.  This loss is critical for good performance (\cite{kendall:cvpr2017}, and see Section \ref{sec:ablation}). 
     
    \noindent\textbf{Locality loss.} Although the keypoint location is estimated from the \textit{UV} probability map, that map might not be unimodal and might have high probabilities away from the true keypoint location. This loss encourages the probability map to localize around the keypoint.
      \begin{equation}
        \La_\mathit{loc} = \sum_{i \in \mathit{kps}} \sum_{uv} \mathit{prob}_i[uv] \cdot \widetilde{\mathcal{N}}(UV_i^*, \sigma)[uv]
     \end{equation}   
    $\mathcal{N}$ is a circular normal distribution centered on the labeled $\mathit{UV}_i^*$ coordinates for keypoint $i$, with standard deviation $\sigma$.  $\widetilde{\mathcal{N}}$ is a normalized inverse
     \begin{equation}
      1 - \mathcal{N} / \max(\mathcal{N}).
     \end{equation}
    This loss gives a very low value when the predicted UV probability is concentrated near the UV label.  We use a $\sigma$ of 10 pixels.
     
The total loss is defined as the weighted sum
\begin{equation}
    \La_\mathit{total} = \La_\mathit{kp} + \alpha\La_\mathit{proj} + 0.001 \La_\mathit{loc}
\end{equation}
A small weight on $\La_\mathit{loc}$ nudges the probability distribution into the correct local form, while allowing room to spread out if necessary.  For stability, it is important to apply a curriculum to $\La_\mathit{proj}$.  The weight $\alpha$ ramps from 0 to 2.5 over the interval $[1/3, 2/3]$ of the training steps, to allow the predicted \uvd\ values to stabilize.  Otherwise, convergence can be difficult because the re-projection error gradients might initially be very large.

\textbf{Permutation for symmetric keypoints.}
Symmetric objects can cause aliasing among keypoints ids.  For example, the tree object in Figure \ref{fig:permutation} is indistinguishable when rotated $180^\circ$ around its vertical axis.  A keypoint placed on the object may thus obtain different, indistinguishable positions from the point of view of the pose estimator.

\begin{table*}[h]
\small
\setlength{\tabcolsep}{1.5pt}
\centering
\resizebox{\textwidth}{!}{
\begin{tabular}{l|c|c V{2} ccccccccccccccc|c}
\hline
method & input modality & metrics & ball & bottle$_0$ & bottle$_1$ & bottle$_2$ & cup$_0$ & cup$_1$ & mug$_0$ & mug$_1$ & mug$_2$ & mug$_3$ & mug$_4$ & mug$_5$ & mug$_6$ & heart & tree & mean \\
\hline
DenseFusion & mono RGB & AUC$\uparrow$ & 90.0 & 88.6  & 69.1  & 56.0  & 84.0  & 80.7  &  67.8  & 66.3  & 71.4  & 70.0  & 69.0  & 76.8  & 51.2 & 61.7 & 75.5 & 71.9 \\
\cite{densefusion} & + \textbf{opaque} depth & $<$2cm$\uparrow$ & 94.4 & 97.8 & 9.1 & 28.4 & 79.1 & 65.3 & 12.5 & 10.3 & 28.1 & 20.3 & 4.7 & 41.9 & 3.1 & 17.2 & 50.9 & 37.5 \\
 & & MAE$\downarrow$ & 9.9 & 11.3 & 57.6 & 77.8 & 16.0 & 37.5 & 32.2 & 33.7 & 28.6 & 30.0 & 31.0 & 23.2 & 75.2 & 38.3 & 24.5 & 35.1
 \\
\hline
DenseFusion & mono RGB & AUC$\uparrow$ & 84.7 & 81.6 & 72.3 & 47.5 & 59.4 & 77.8 & 54.5 & 51.3 & 60.4 & 67.3 & 48.1 & 70.6 & 64.9 & 61.2 & 55.6 & 63.8 \\
\cite{densefusion} & + \textbf{real} depth & $<$2cm$\uparrow$ & 78.8 & 67.5 & 18.1 & 9.1 & 5.6  & 54.4 & 4.6 & 0.3 & 12.2 & 8.1 & 0.0 & 20.0 & 4.7 & 0.0 & 0.0 & 18.9 \\
& & MAE$\downarrow$& 15.3 & 18.4 & 27.6 & 65.6 & 40.5 & 22.1 & 45.5 & 48.7 & 39.5 & 32.7 & 54.9 & 29.4 & 35.9 & 38.8 & 44.4 & 37.2
\\
\hline
\multirow{3}{*}{\textbf{Ours}} & \multirow{3}{*}{stereo RGB} & AUC$\uparrow$ & \textbf{96.1} &  \textbf{95.4} & \textbf{94.9}  & \textbf{90.7}  & \textbf{93.1}  & \textbf{92.0}  & \textbf{91.0}  & \textbf{78.1} & \textbf{89.7} & \textbf{88.6}  & \textbf{87.8}  & \textbf{91.0} & \textbf{90.3} & \textbf{84.3} & \textbf{87.1} & \textbf{90.0}\\
& & $<$2cm$\uparrow$ & \textbf{100} &  \textbf{99.8} & \textbf{99.7} & \textbf{91.4} & \textbf{97.8} & \textbf{95.3} & \textbf{94.6} & \textbf{63.6} &  \textbf{90.1} & \textbf{87.2} & \textbf{87.1} &  \textbf{93.1} & \textbf{92.2} & \textbf{77.2} & \textbf{82.5} & \textbf{90.1} \\
& & MAE$\downarrow$ & \textbf{3.8} & \textbf{4.6} & 
\textbf{5.1} & \textbf{9.3} & \textbf{6.8} &
\textbf{7.1} & \textbf{8.9} & \textbf{21.9}
& \textbf{10.1} & \textbf{11.3} & 
\textbf{12.1} & \textbf{9.0} & \textbf{9.7}
& \textbf{15.6} & \textbf{12.8} &
\textbf{9.9} \\
\hline
\end{tabular}
} 
\vspace{-1ex}
\caption{Instance-level pose estimation results. For each object instance, the model is trained on nine background textures and evaluated on unseen textures.
Higher is better for AUC and $<2$cm, lower for MAE.}
\label{tab:instance}
\vspace{-1ex}
\end{table*}

\begin{figure}[t]
  \vspace*{-3ex}
  \subfloat{% 
    \includegraphics[width=0.49\linewidth]{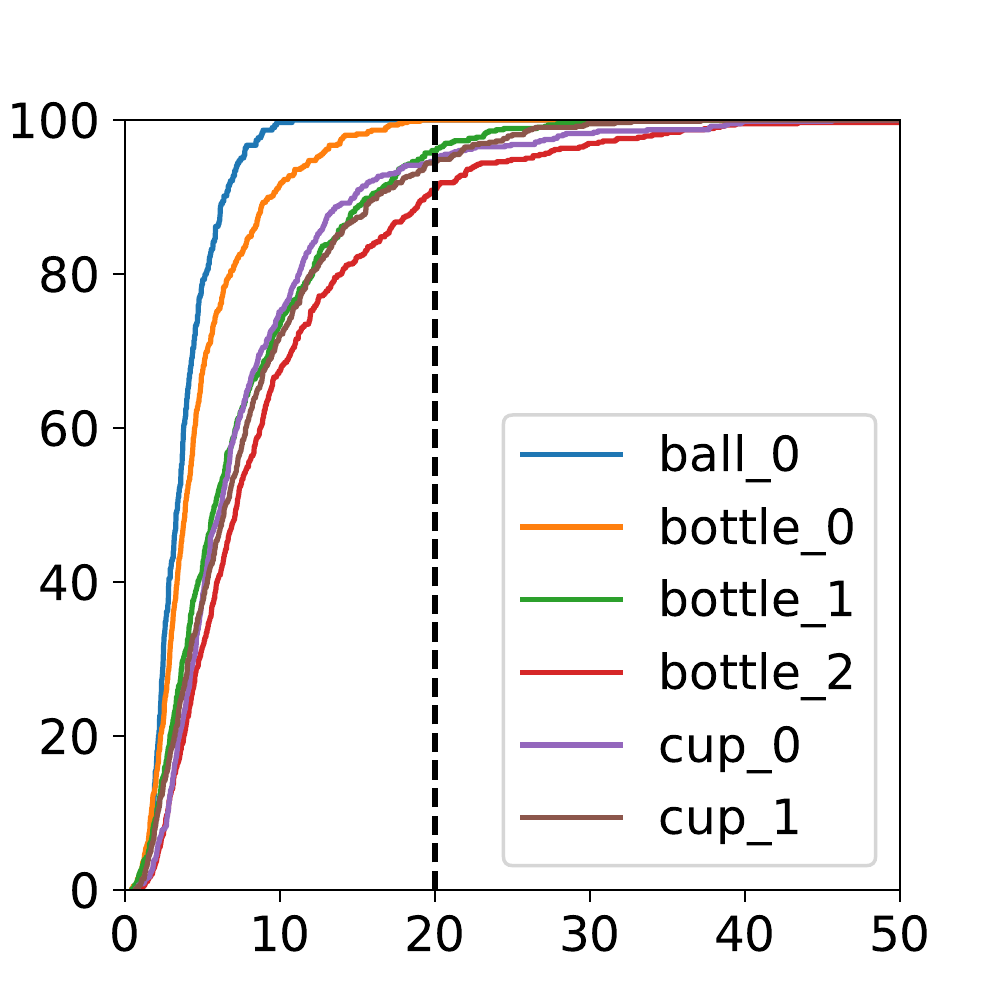} 
  }
  \subfloat{% 
    \includegraphics[width=0.49\linewidth]{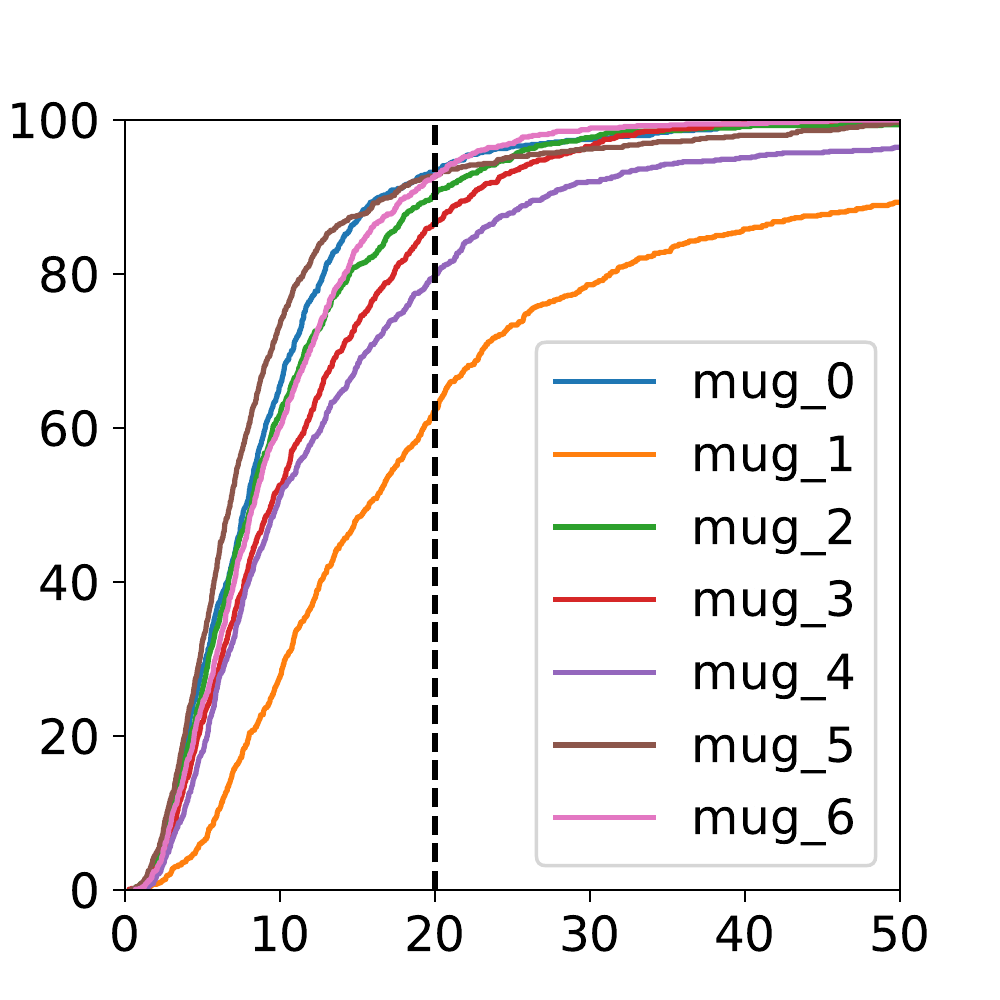} 
  }
  \vspace{-1ex}
  \caption{Precision curves for object instances.  $Y$ axis is cumulative percent.  $X$ axis is 3D MAE in mm; note it is limited to 50 mm instead of the normal 100 mm, to magnify the curves.}
  \vspace{-1ex}
  \label{fig:auc}
\end{figure}

We deal with keypoint symmetry by allowing permutation of the relevant keypoint ids in the loss function.  For example, in the case of the tree, there are two allowed permutations of the keypoint ids, $[1,2,3,4]$ and $[1,2,4,3]$.  $\La_\mathit{total}$ is evaluated for each of these permutations, and the minimum is chosen as the final loss.

Keypoints handle some symmetries without any permutations.  These are illustrated by the ball, bottle, and cup objects.  For the ball, a single keypoint at the center confers full rotational symmetry.  For bottles and cups, two keypoints along the cylindrical axis confer cylindrical symmetry.  Note that we may choose to use fewer keypoints than necessary -- for example, if we do not care where the handle of a mug is, we could use only the top and bottom keypoints.

\subsection{Training}
We trained the KeyPose model with a batch size of 32 and a constant number of steps, around 300 epochs.  For DenseFusion, we re-implemented the algorithm in TensorFlow and trained until convergence, around 80 epochs.  Since DenseFusion does not return keypoints, we added layers to regress to 3D positions for each keypoint.
More training details are in the supplementary material.

\section{Experiments}

We conducted experiments to test the KeyPose model and DenseFusion \cite{densefusion}, on the TOD dataset.  We compared two input variants for DenseFusion, with depth from opaque and transparent (real) versions of the object.  Remember that in the case of opaque depth, we still use the RGB image of the transparent object.  We trained both instance and category models, and derived test sets by holding out all sequences of a texture, and all sequences of an object.  We also performed ablation studies to understand the effects of stereo and the various losses.

Two error measures that are standard in the literature \cite{poserbpf,densefusion,posecnn} are Area Under the Curve (AUC) and percentage of 3D keypoint errors $<$2cm.  AUC  percentage is calculated based on an $X$-axis range from 0 to 10 cm, where the curve shows the cumulative percentage of errors under that metric value (Figure \ref{fig:auc}).  These measures were developed for lower-accuracy methods, and we prefer a more precise measure, Mean Absolute Error (MAE) of the 3D keypoints.

\begin{table}[t]
\small
\setlength{\tabcolsep}{1.5pt}
\centering
\resizebox{\linewidth}{!}{
\begin{tabular}{l| P{0.9cm} P{1.0cm} P{0.8cm} | P{0.8cm} P{1.0cm} P{0.8cm} | P{0.9cm} P{1.0cm} P{0.8cm} }
\hline
method &  
\multicolumn{3}{c|}{DenseFusion \cite{densefusion} } & 
\multicolumn{3}{c|}{DenseFusion \cite{densefusion} } & 
\multicolumn{3}{c}{ \textbf{Ours} } \\ 
\hline
input &
\multicolumn{3}{c|}{\multirow{2}{*}{\begin{tabular}[c]{@{}c@{}}monocular RGBD \\ + \textbf{opaque} depth \end{tabular}}} &
\multicolumn{3}{c|}{\multirow{2}{*}{\begin{tabular}[c]{@{}c@{}}monocular RGBD \\ + \textbf{real} depth \end{tabular}}} &
\multicolumn{3}{c}{\multirow{2}{*}{\begin{tabular}[c]{@{}c@{}} stereo \\ RGB  \end{tabular}}} \\
modality & 
\multicolumn{3}{c|}{} & 
\multicolumn{3}{c|}{} &  
\multicolumn{3}{c}{} \\ \hline
metrics & AUC$\uparrow$ & $<$2cm$\uparrow$ & MAE$\downarrow$ & AUC$\uparrow$ & $<$2cm$\uparrow$ & MAE$\downarrow$ & AUC$\uparrow$ & $<$2cm$\uparrow$ & MAE$\downarrow$ \\
\specialrule{.1em}{.05em}{.05em}
bottles & 83.4 & 88.4 &  34.2 & 76.9 & 71.0 &  26.4 &
\textbf{94.2} & \textbf{97.8} &  \textbf{5.8} \\ 
bots+cups & 90.0 & 93.4 &  10.5 & 77.2 & 70.3 &  24.5 &
\textbf{93.4} & \textbf{97.8} &  \textbf{6.6} \\ 
mugs & 82.4 & 72.8 &  17.6 & 73.5 & 41.5 &  26.5 &
\textbf{90.1} & \textbf{92.6} &  \textbf{9.9} \\ 
\hline
\end{tabular}
}
\vspace{-1ex}
\caption{
Category-level pose estimation results. Evaluate on unseen textures.  Boldface are best results. 
}
\label{tab:single:unseen:texture}
\vspace{-2ex}
\end{table}

\begin{table}[t]
\small
\setlength{\tabcolsep}{1.5pt}
\centering
\resizebox{\linewidth}{!}{
\begin{tabular}{l| P{0.9cm} P{1.0cm} P{0.9cm} | P{0.9cm} P{1.0cm} P{0.9cm} | P{0.9cm} P{1.0cm} P{0.9cm} }
\hline
method &  
\multicolumn{3}{c|}{DenseFusion \cite{densefusion} } & 
\multicolumn{3}{c|}{DenseFusion \cite{densefusion} } & 
\multicolumn{3}{c}{ \textbf{Ours} } \\ 
\hline
input &
\multicolumn{3}{c|}{\multirow{2}{*}{\begin{tabular}[c]{@{}c@{}}monocular RGBD \\ + \textbf{opaque} depth \end{tabular}}} &
\multicolumn{3}{c|}{\multirow{2}{*}{\begin{tabular}[c]{@{}c@{}}monocular RGBD \\ + \textbf{real} depth \end{tabular}}} &
\multicolumn{3}{c}{\multirow{2}{*}{\begin{tabular}[c]{@{}c@{}} stereo \\ RGB  \end{tabular}}} \\
modality & 
\multicolumn{3}{c|}{} & 
\multicolumn{3}{c|}{} &  
\multicolumn{3}{c}{} \\ \hline
metrics & AUC$\uparrow$ & $<$2cm$\uparrow$ & MAE$\downarrow$ & AUC$\uparrow$ & $<$2cm$\uparrow$ & MAE$\downarrow$ & AUC$\uparrow$ & $<$2cm$\uparrow$ & MAE$\downarrow$ \\
\specialrule{.1em}{.05em}{.05em} 
mugs & 76.4 & 40.7 & 23.5 & 74.3 & 43.4 & 25.7 & 
\textbf{84.7} & \textbf{78.6} & \textbf{15.6} \\ 
\hline
\end{tabular}
}
\vspace{-1ex}
\caption{Pose estimation for the mug category.  Evaluate on unseen instance mug$_0$.}
\label{tab:single:category}
\vspace{-2ex}
\end{table}

\subsection{Instance-Level Pose Estimation}

Each of the 15 objects was trained separately, and statistics computed for a held-out texture.  There were approximately 3000 training samples and 320 test samples. This experiment captures how well an instance-level model can generalize in a new setting. The results are illustrated in Table \ref{tab:instance}.
Not surprisingly, DenseFusion(opaque) performed better than DenseFusion(real) in almost every case, with the exception of cup$_1$ and mug$_6$.  These latter may be due to the errors in depth from the depth device, which even in the opaque case can have significant errors -- see the Supplemental.  For both, the 3D errors were large, averaging over 35 mm across the dataset.

KeyPose out-performed DenseFusion(real) across-the-board, often by large amounts.  Surprisingly, it also performed better than DenseFusion(opaque) on all objects.  This is despite the large premium offered by good depth information for the latter.  KeyPose MAE was 9.9 mm, averaged over all objects, more than 3.5 times more accurate than DenseFusion.
These results demonstrate that KeyPose with stereo input works remarkably well for transparent objects. Given its performance relative to DenseFusion(opaque), it is capable of surpassing state-of-the-art results for pose estimation of desktop object instances.

\newcommand{\picheight}{0.123}
\begin{figure*}[t]
\centering
\subfloat{% 
    \includegraphics[height=\picheight\linewidth]{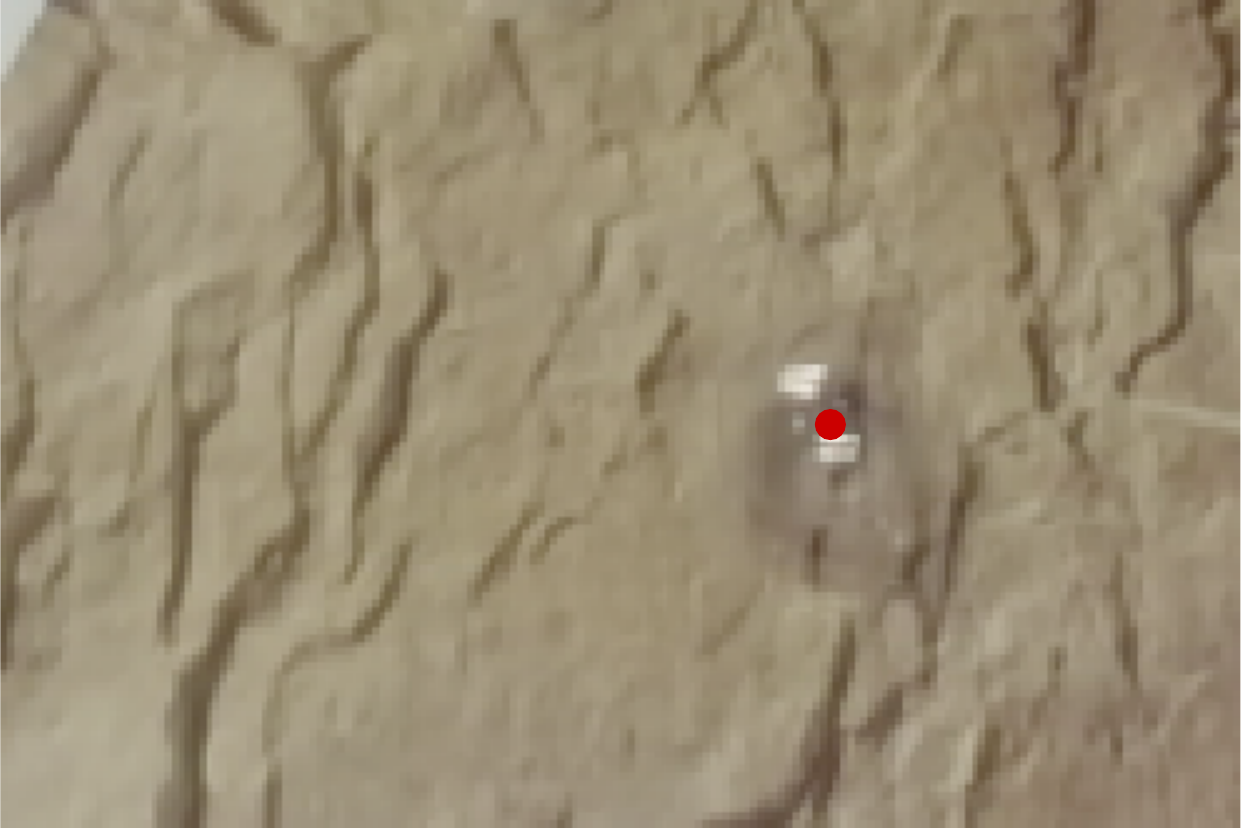}
  } 
  \subfloat{% 
    \includegraphics[height=\picheight\linewidth]{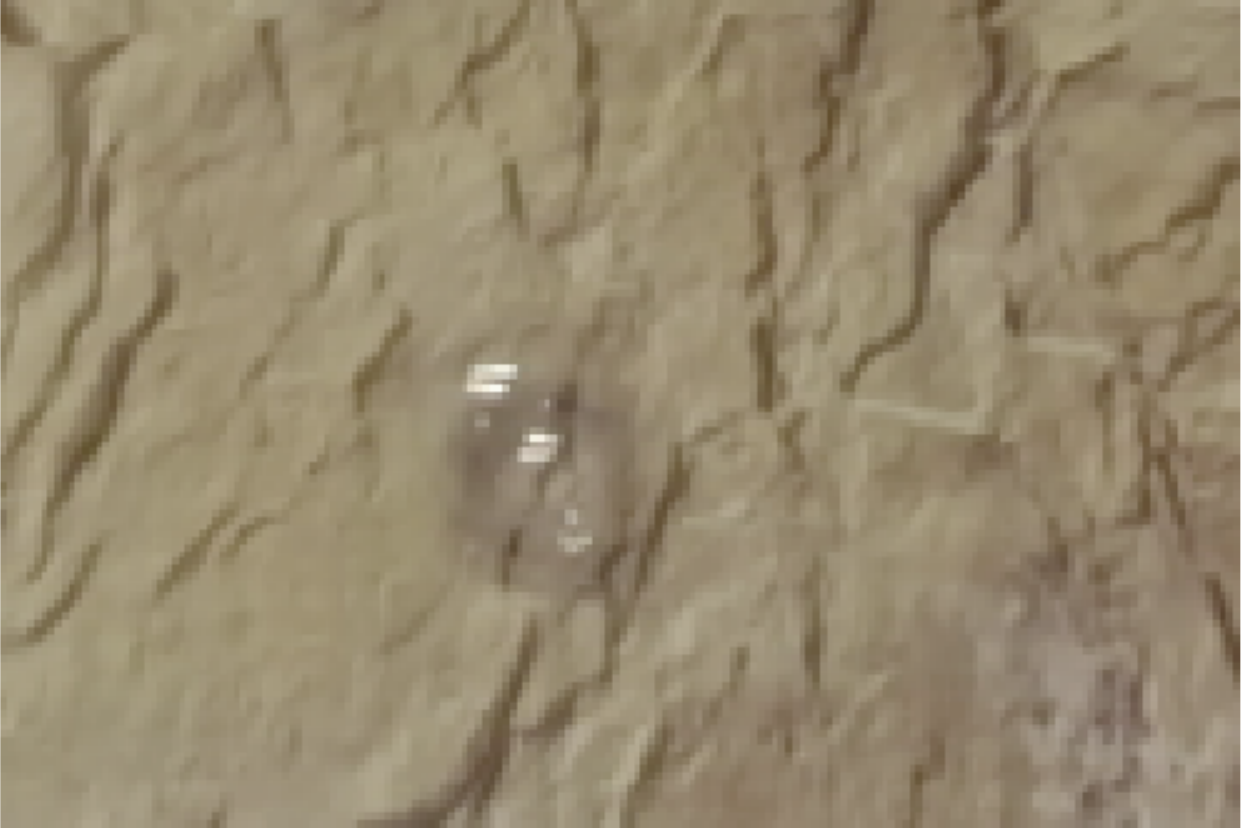} 
  }
  \subfloat{% 
    \includegraphics[height=\picheight\linewidth]{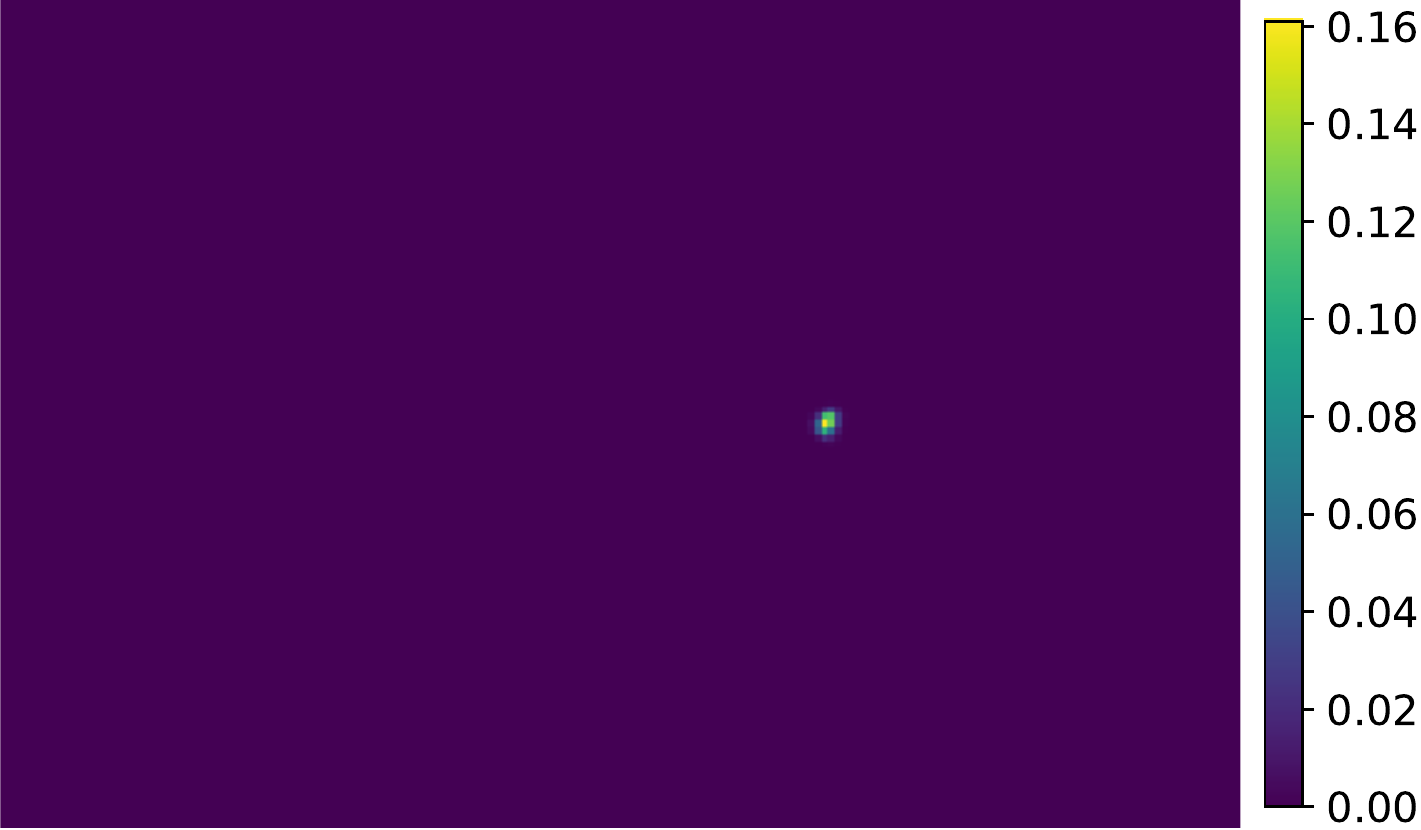} 
  }
  \hspace{0.21\linewidth} 
  \subfloat{% 
    \includegraphics[height=\picheight\linewidth]{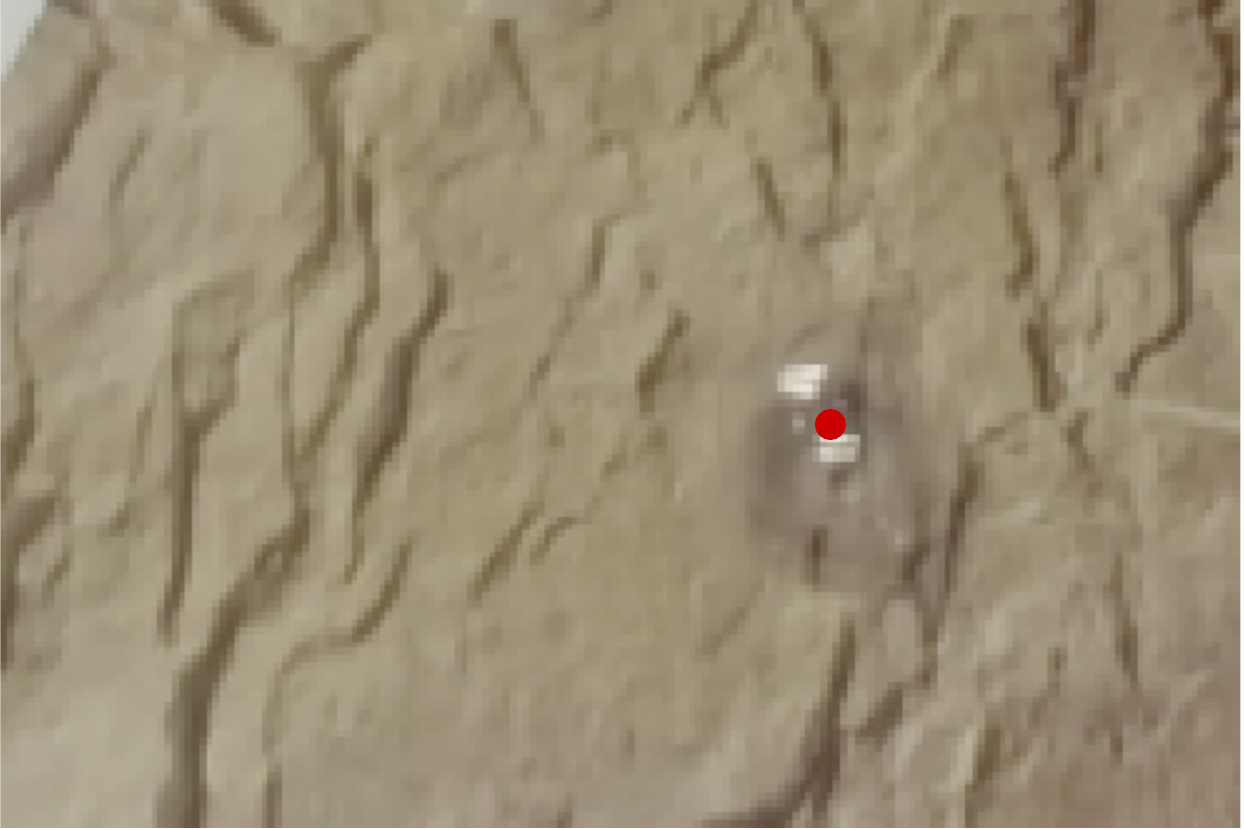} 
  }
  \\
  \vspace{-1.8ex}
  \subfloat{% 
    \includegraphics[height=\picheight\linewidth]{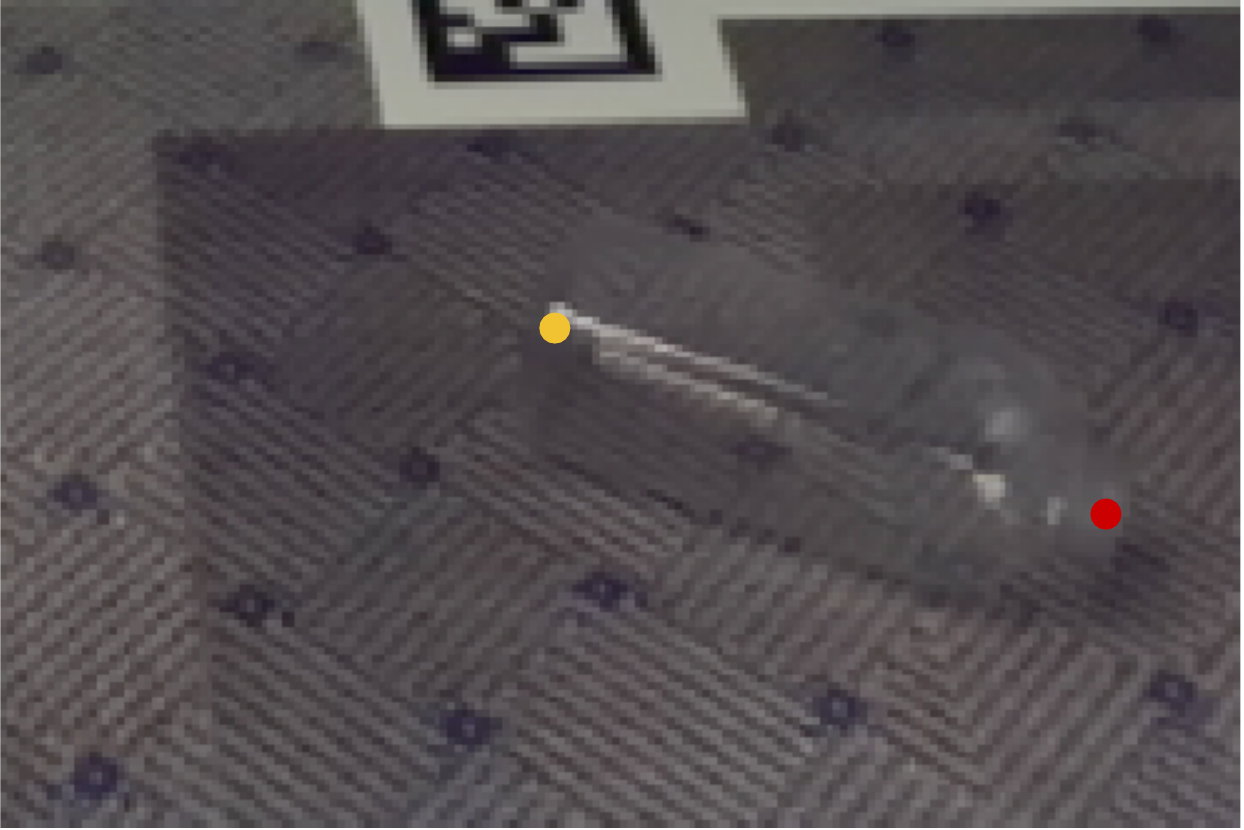}
  } 
  % \hspace{3.5ex} 
  \subfloat{% 
    \includegraphics[height=\picheight\linewidth]{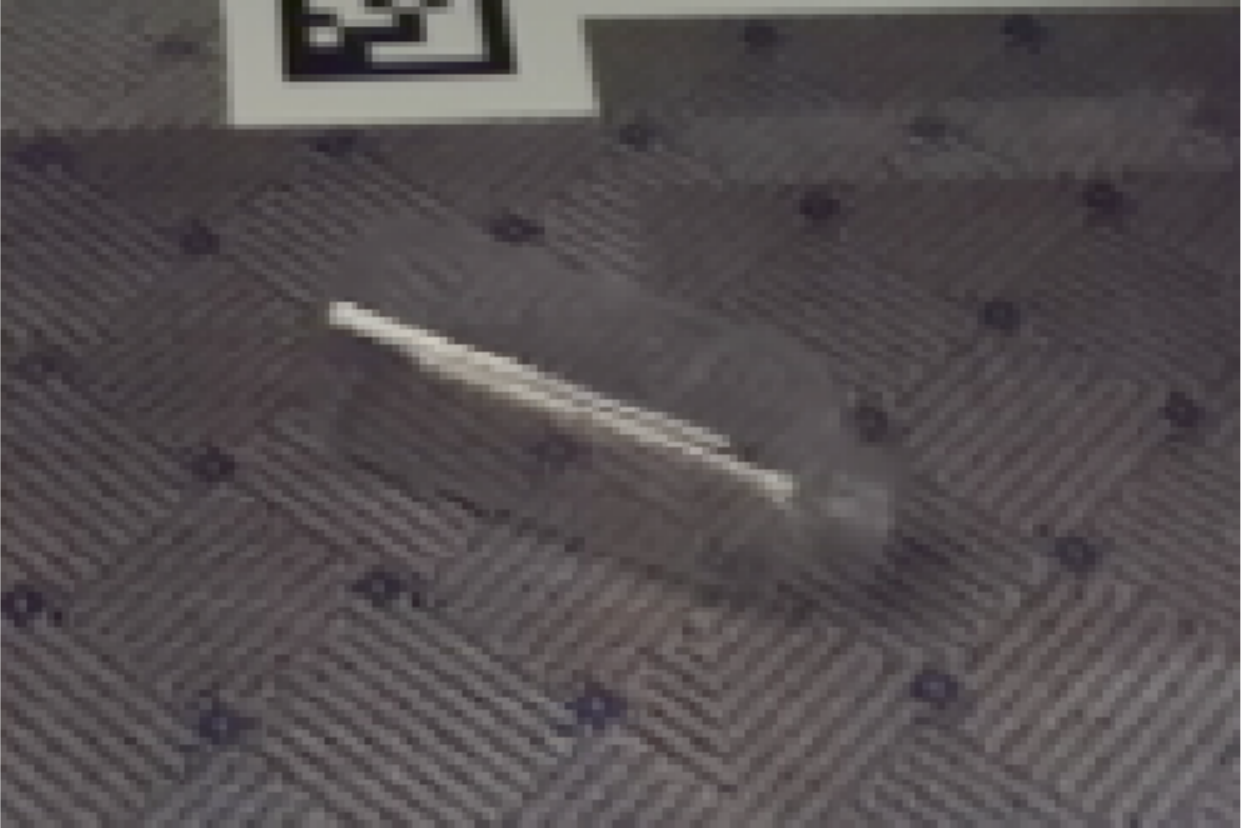} 
  }
  \subfloat{% 
    \includegraphics[height=\picheight\linewidth]{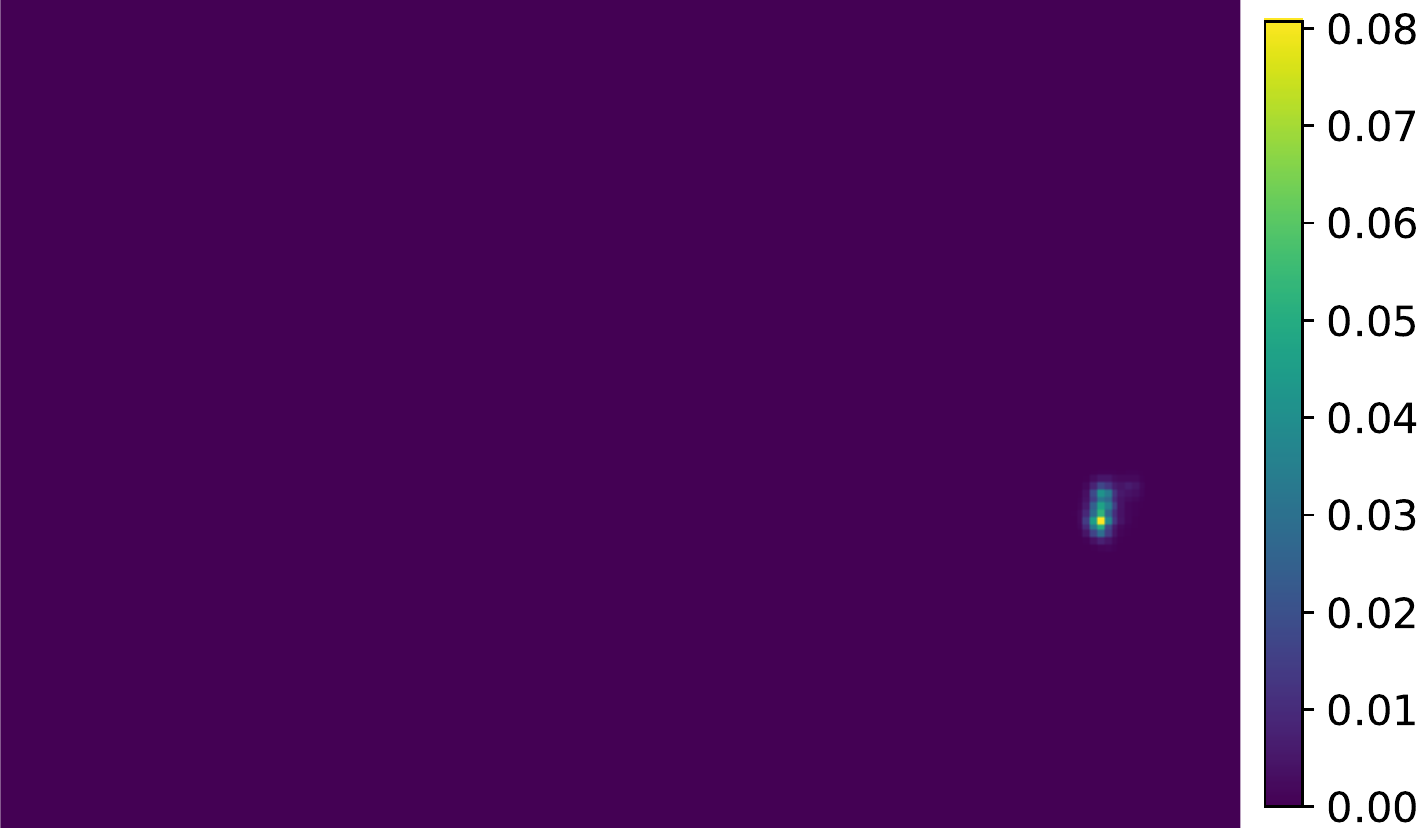} 
  }
  \subfloat{% 
    \includegraphics[height=\picheight\linewidth]{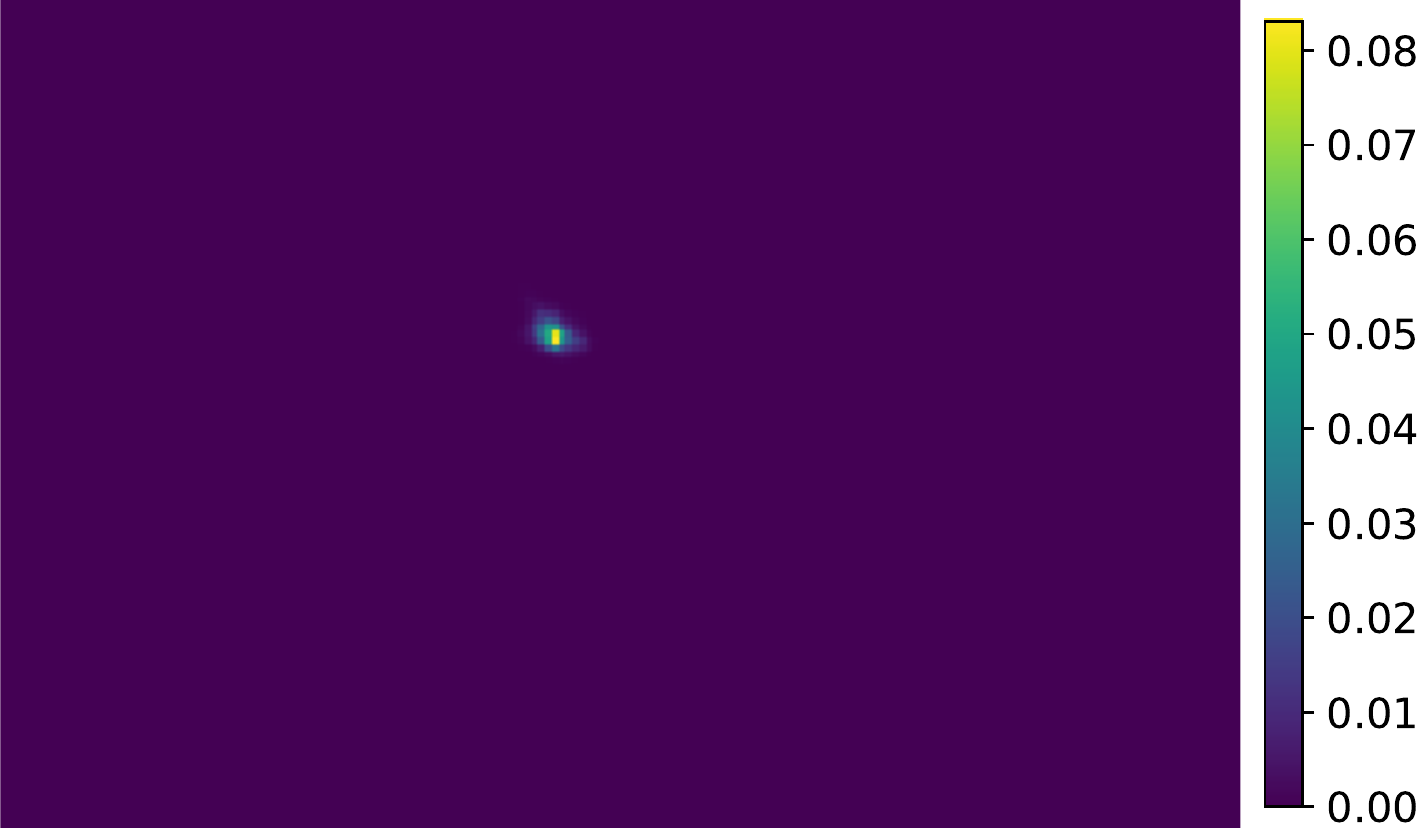} 
  }
  \subfloat{% 
    \includegraphics[height=\picheight\linewidth]{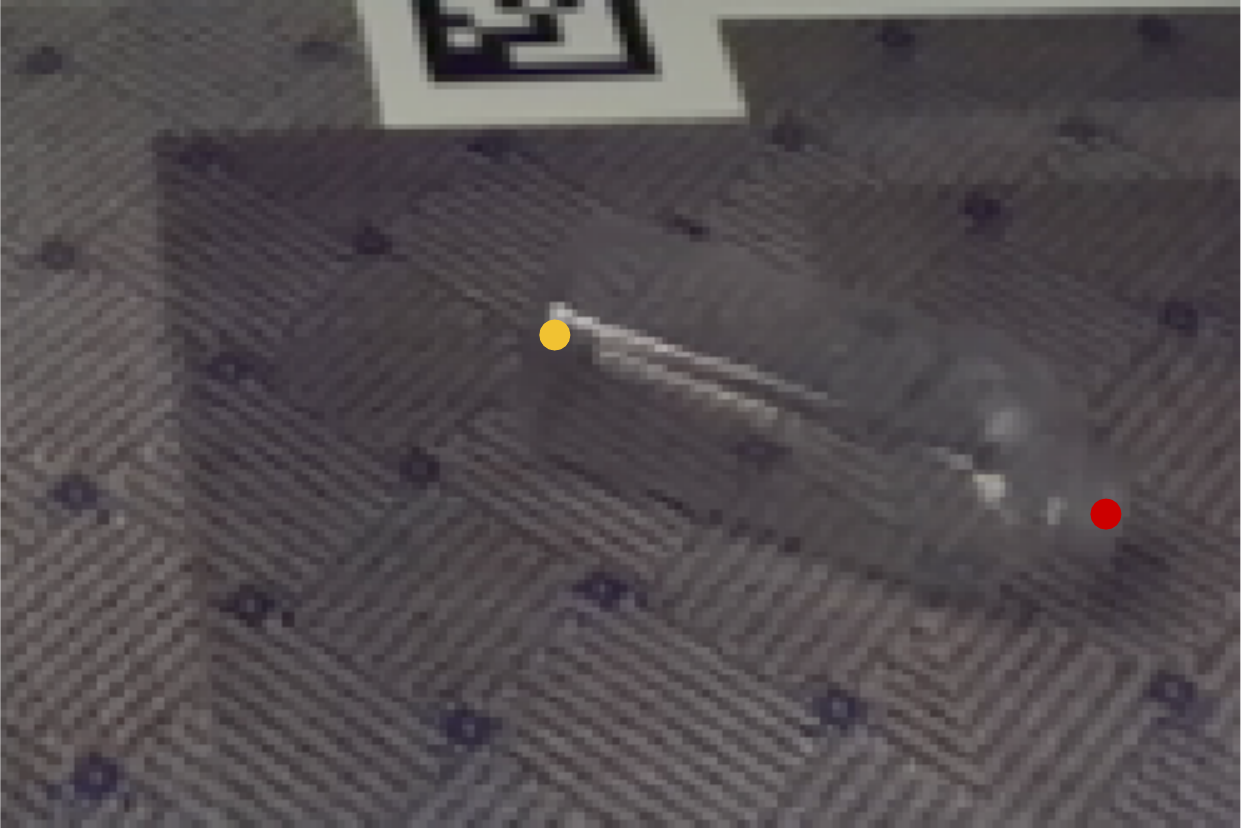} 
  }
  \\
  \vspace{-1.8ex}
  \subfloat{% 
    \includegraphics[height=\picheight\linewidth]{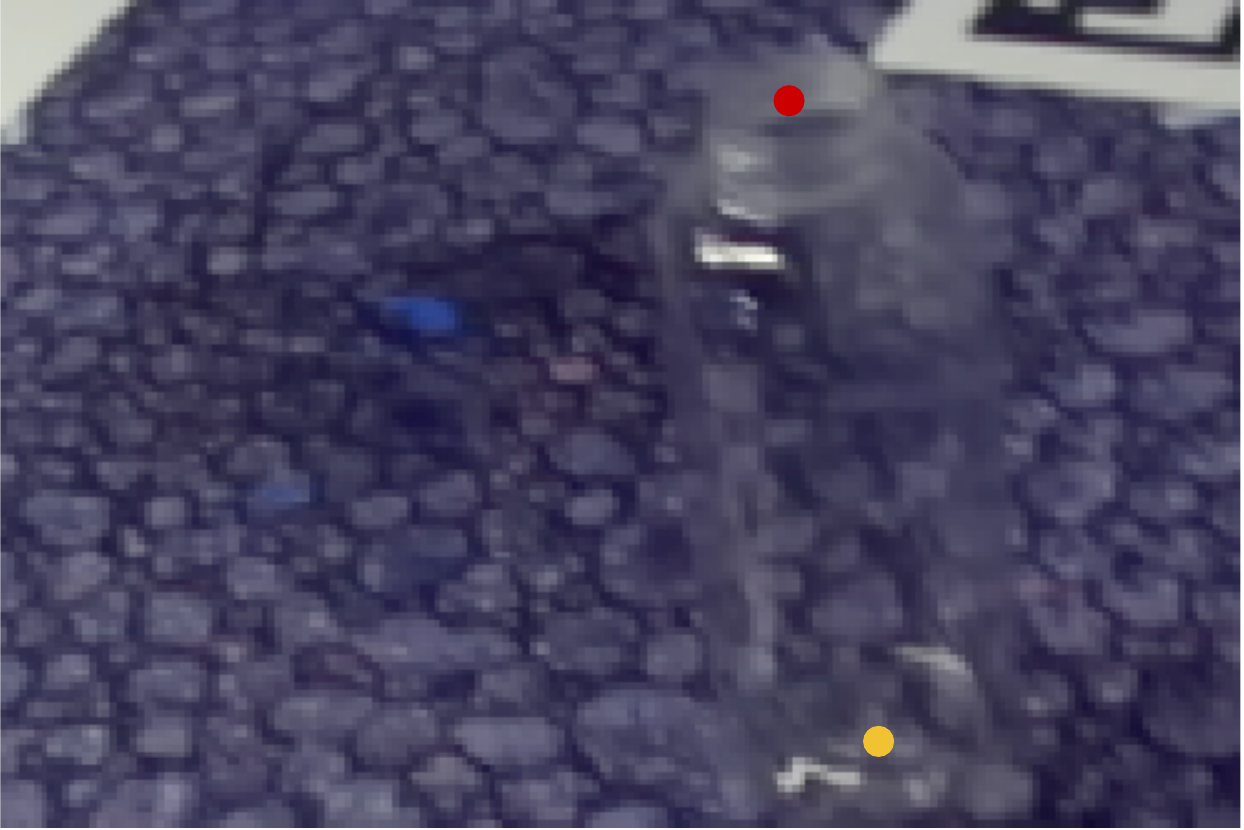}
  } 
  % \hspace{3.5ex} 
  \subfloat{% 
    \includegraphics[height=\picheight\linewidth]{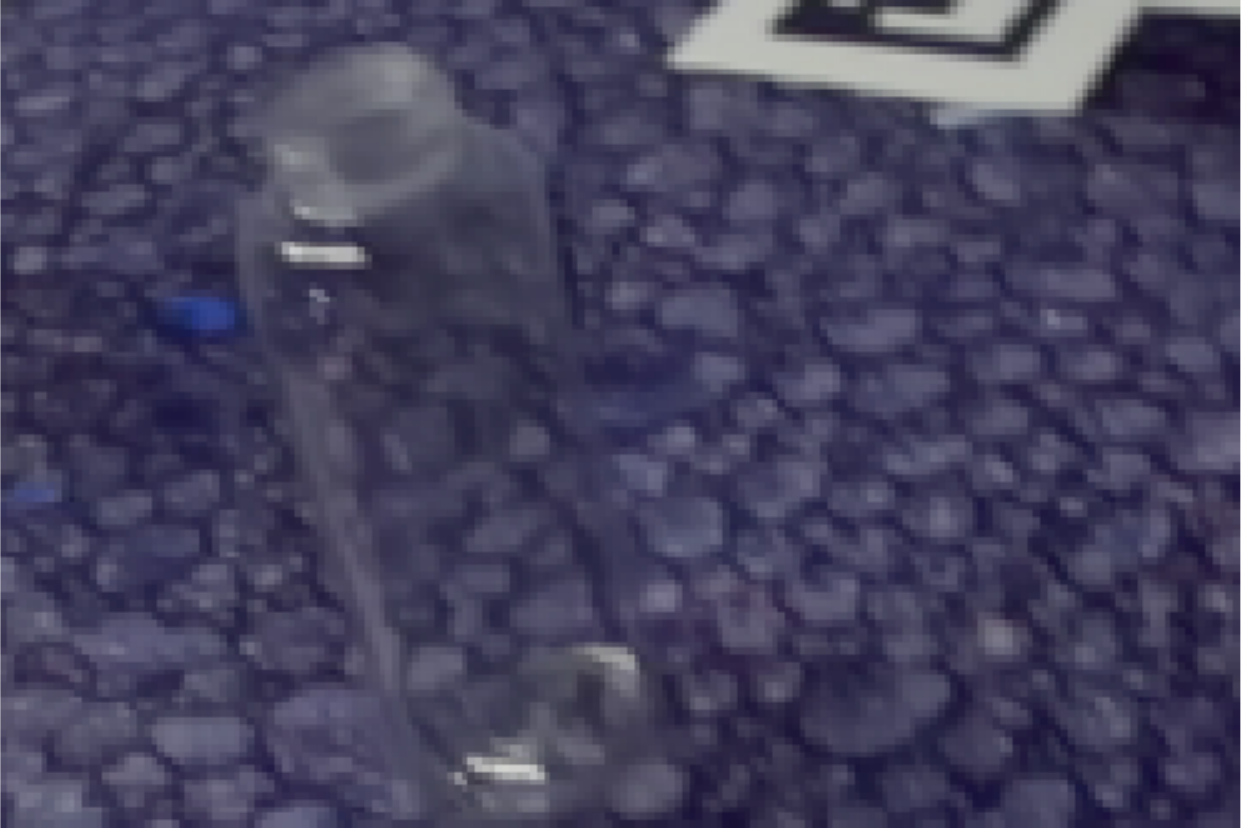} 
  }
  \subfloat{% 
    \includegraphics[height=\picheight\linewidth]{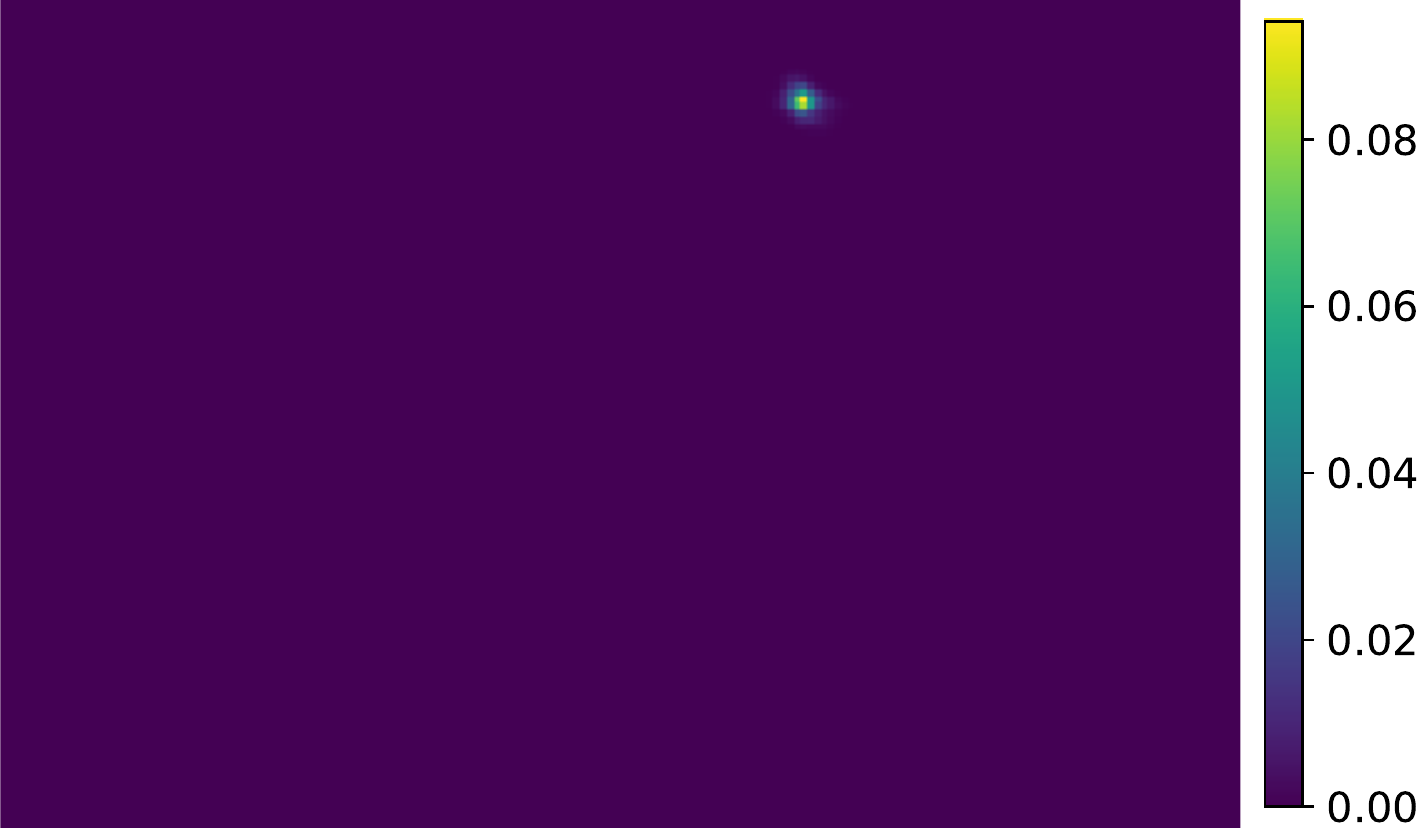} 
  }
  \subfloat{% 
    \includegraphics[height=\picheight\linewidth]{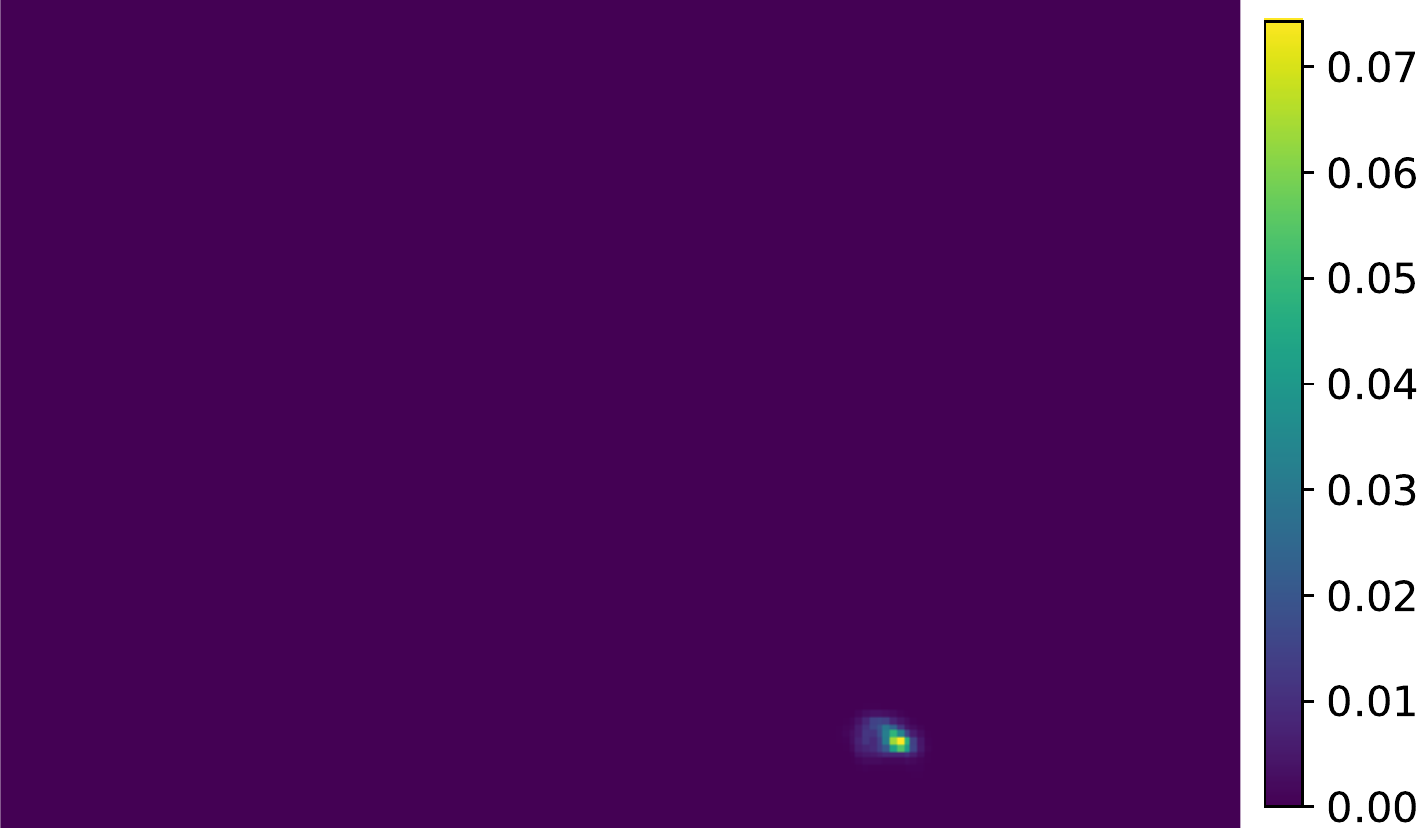} 
  }
  \subfloat{% 
    \includegraphics[height=\picheight\linewidth]{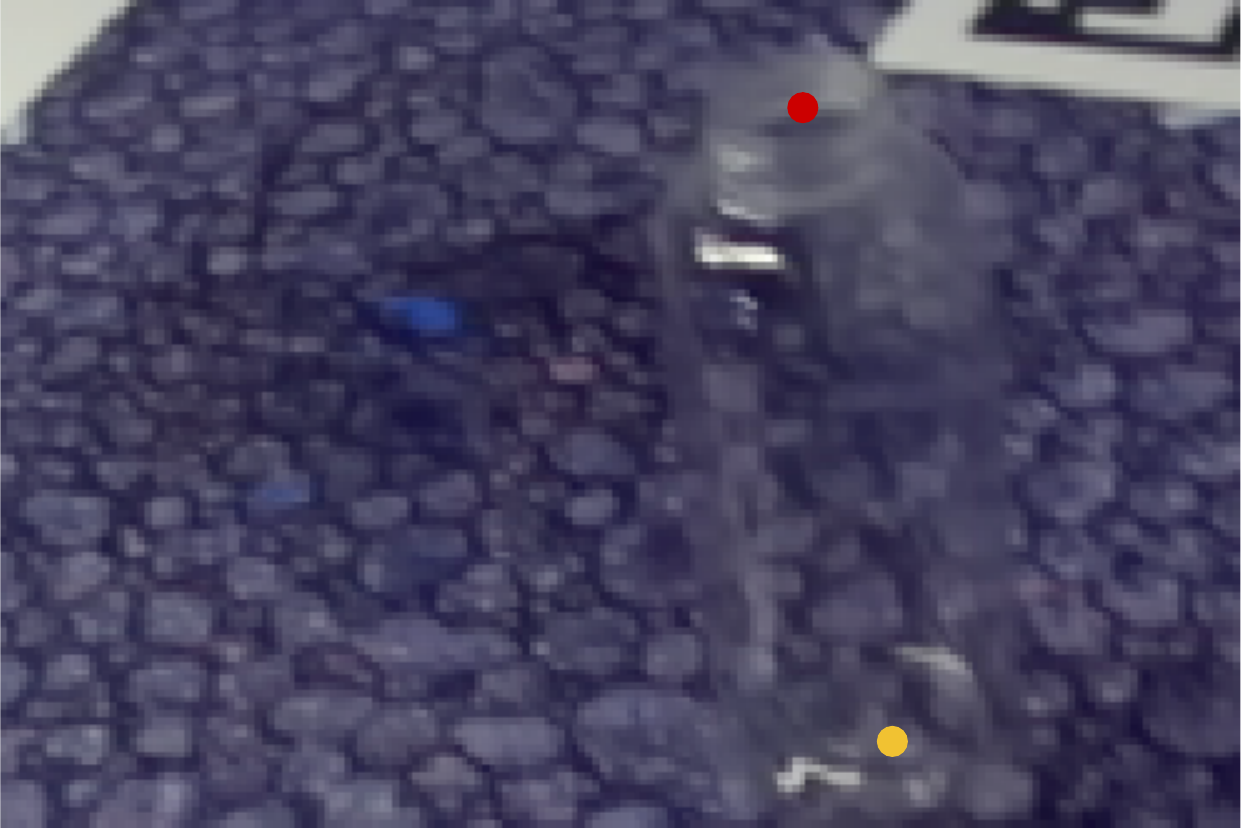} 
  }
  \\
  \vspace{-1.8ex}
  \subfloat{% 
    \includegraphics[height=\picheight\linewidth]{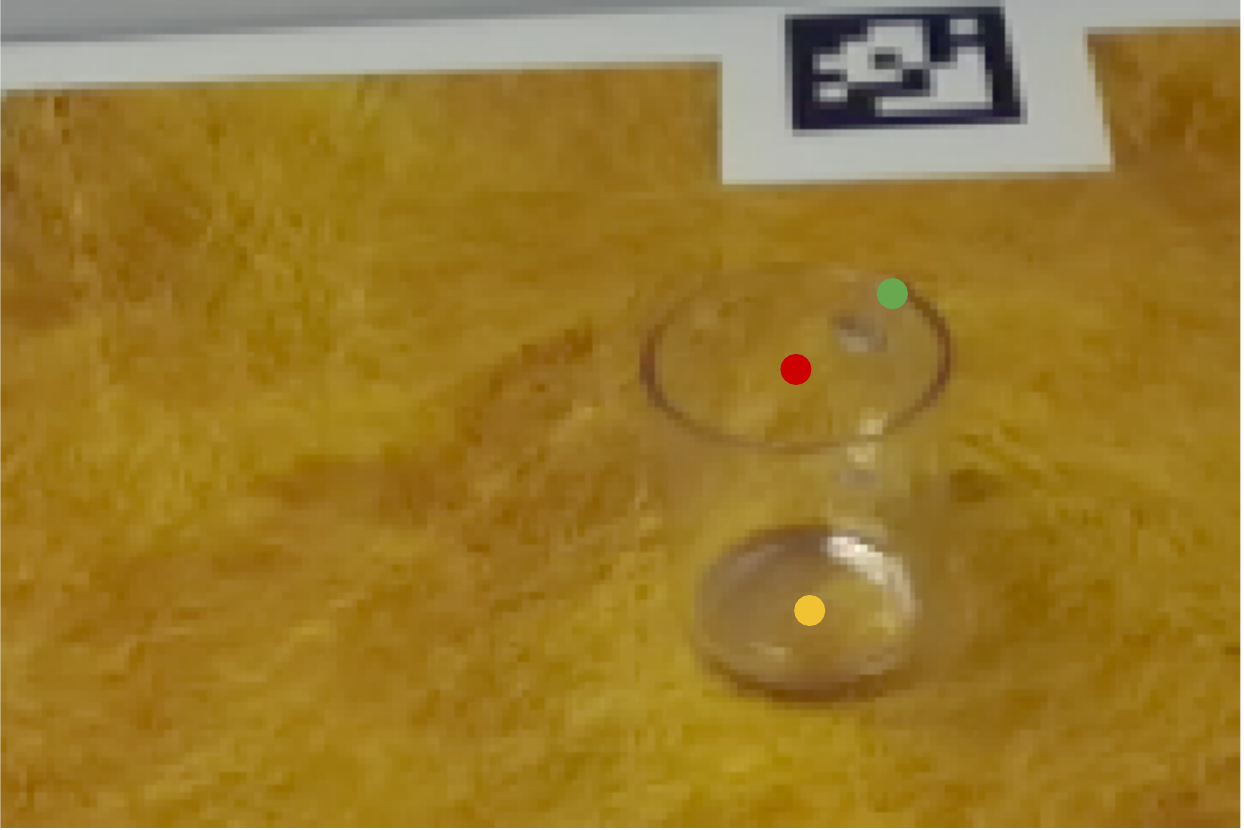}
  } 
  % \hspace{3.5ex} 
  \subfloat{% 
    \includegraphics[height=\picheight\linewidth]{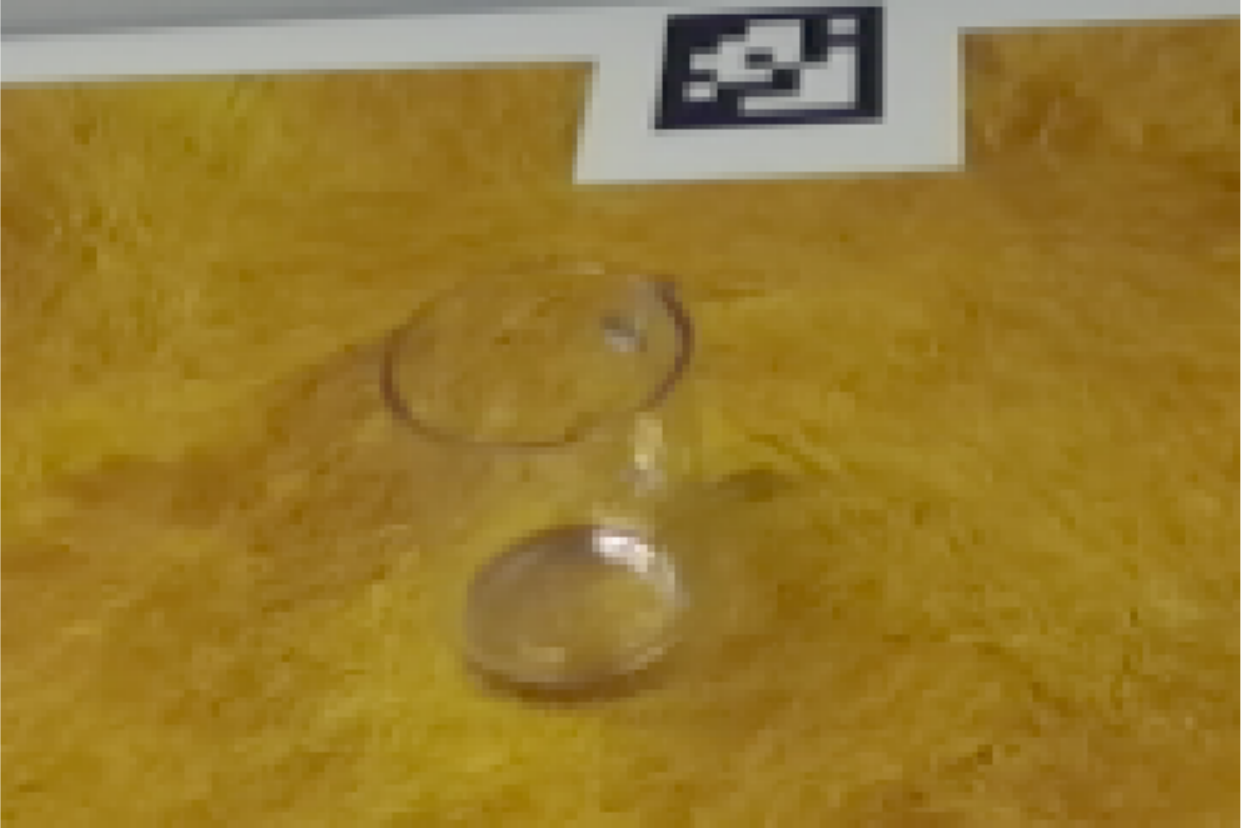} 
  }
  \subfloat{% 
    \includegraphics[height=\picheight\linewidth]{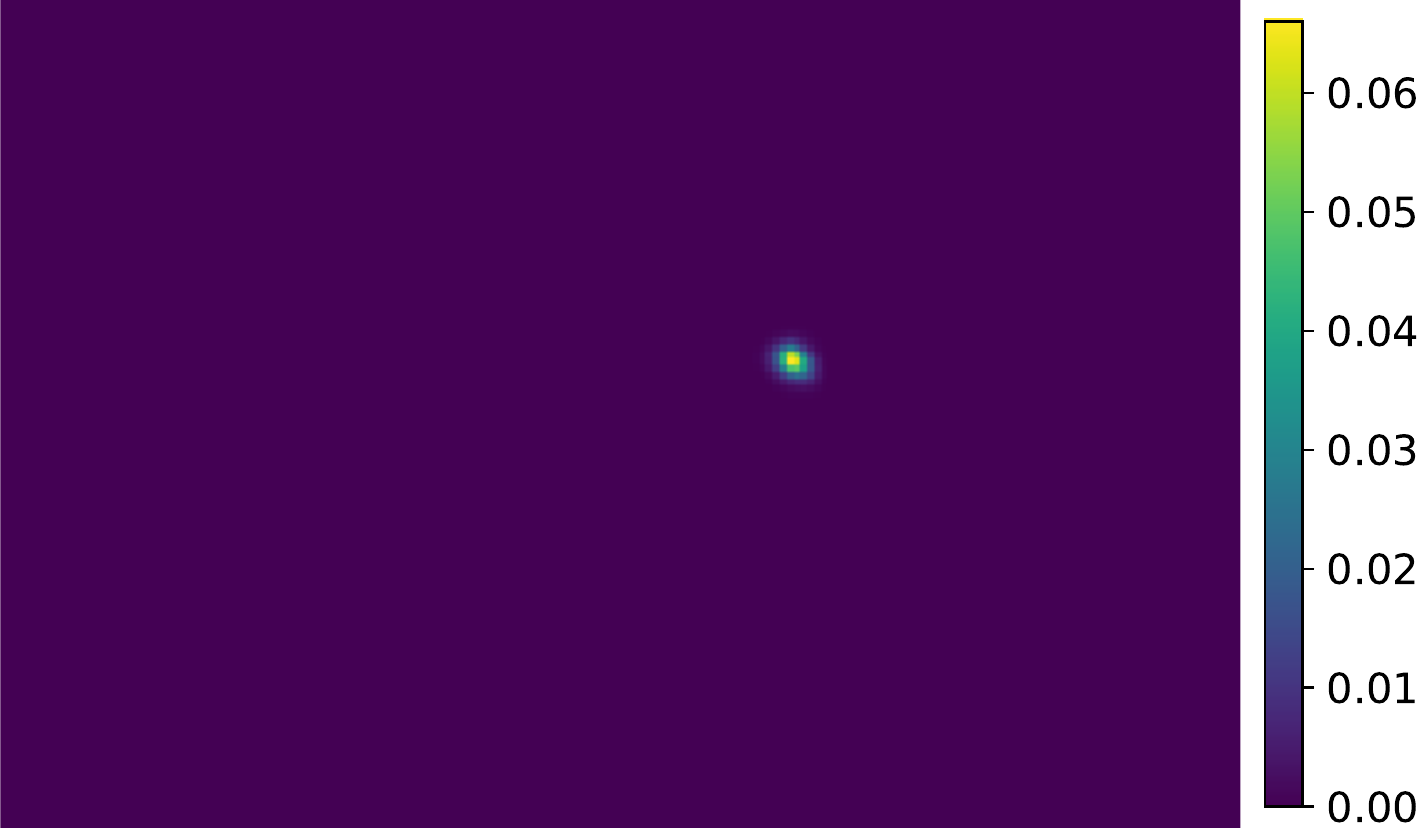} 
  }
  \subfloat{% 
    \includegraphics[height=\picheight\linewidth]{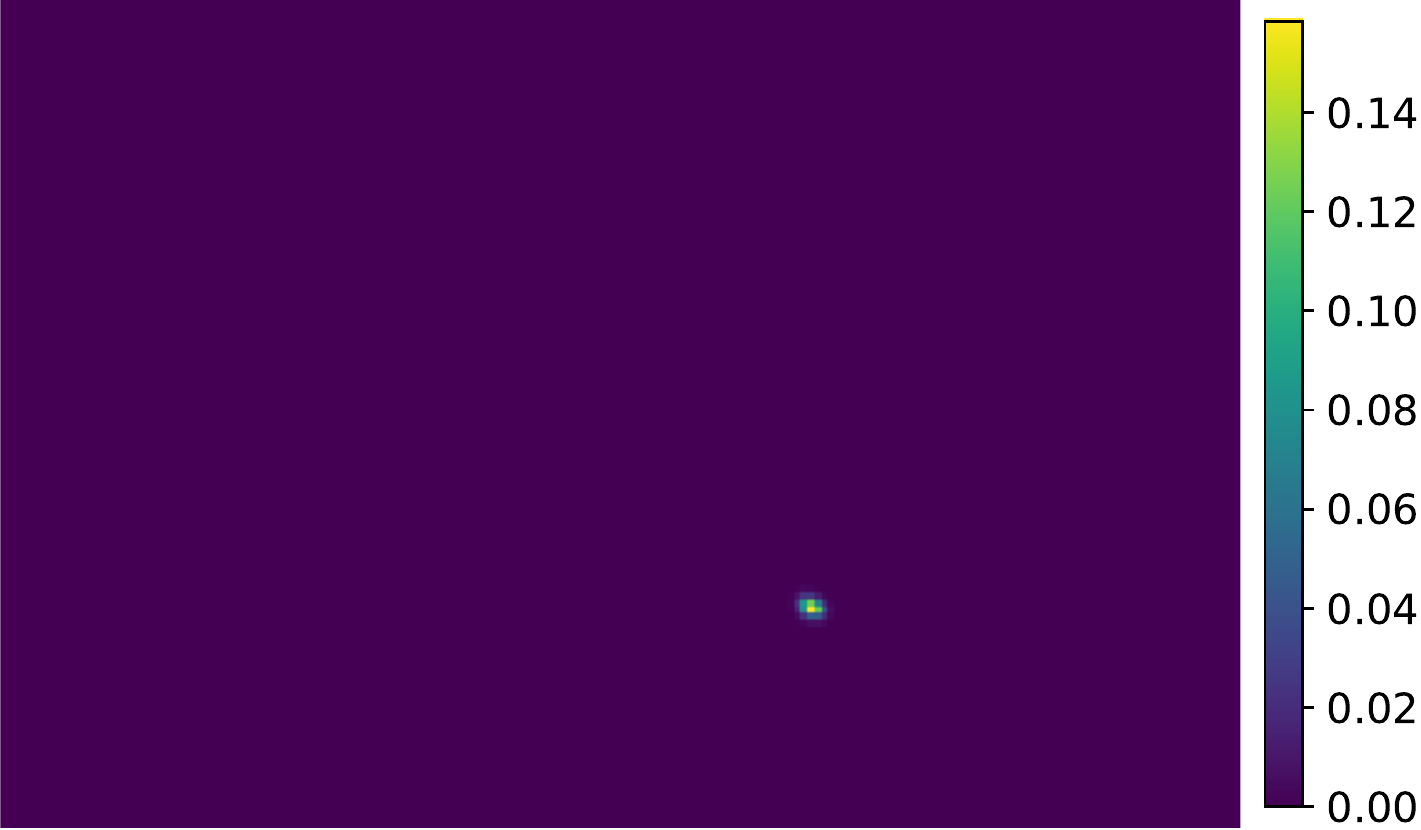} 
  }
  \subfloat{% 
    \includegraphics[height=\picheight\linewidth]{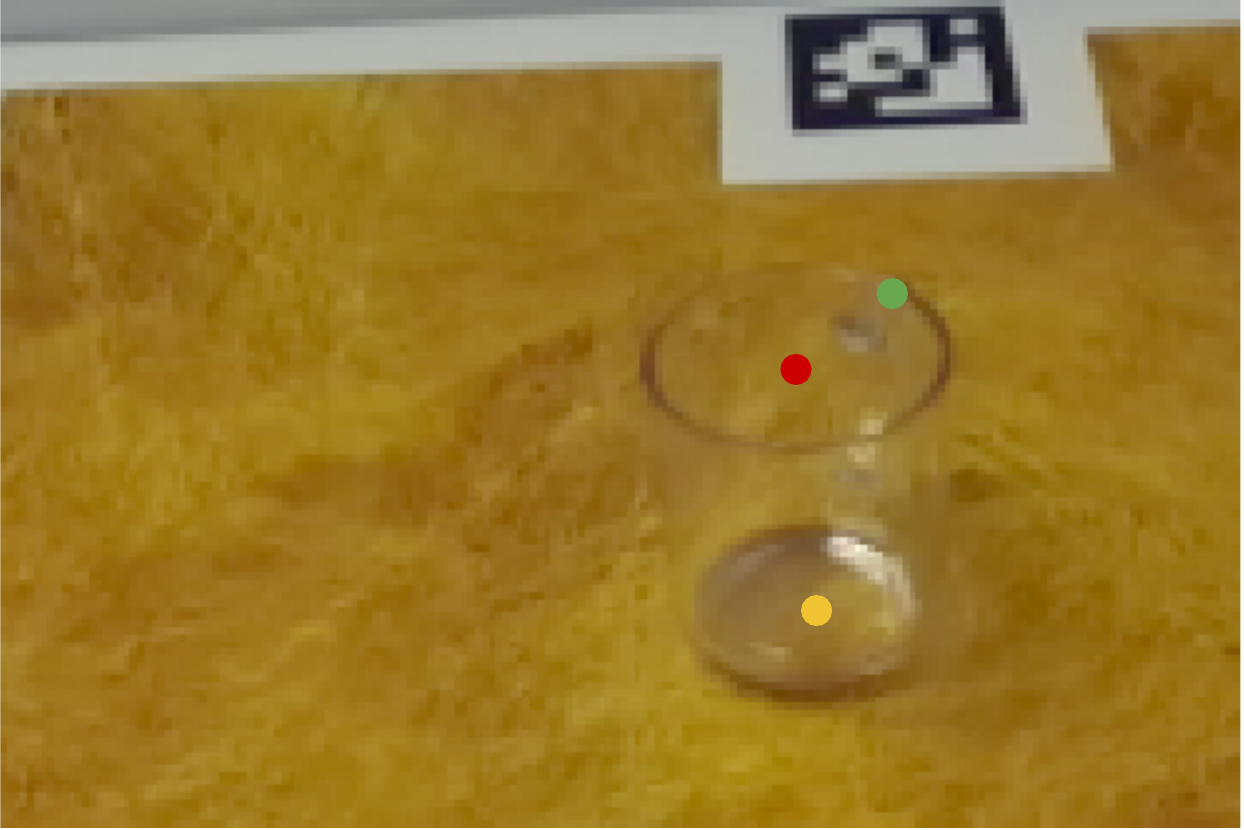} 
  }
  \\
  \vspace{-1.8ex}
  \subfloat{% 
    \includegraphics[height=\picheight\linewidth]{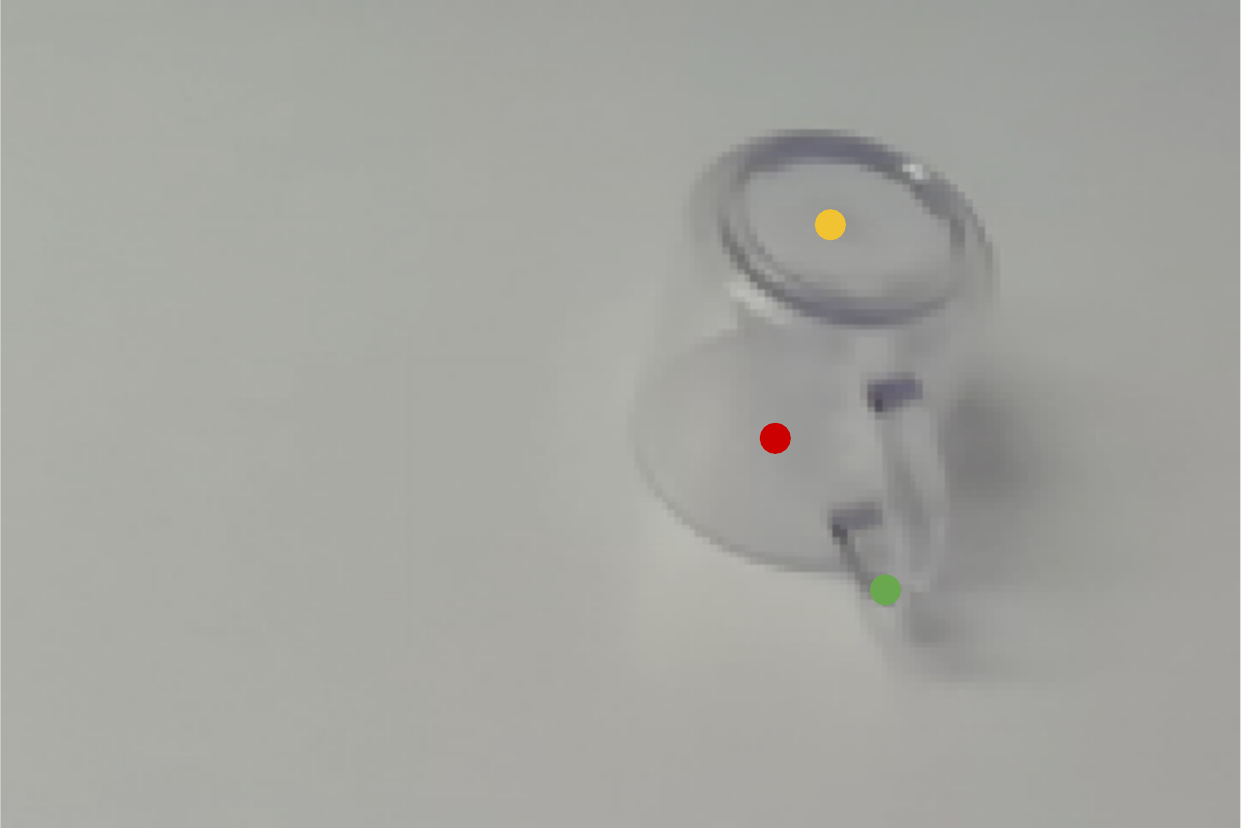}
  } 
  % \hspace{3.5ex} 
  \subfloat{% 
    \includegraphics[height=\picheight\linewidth]{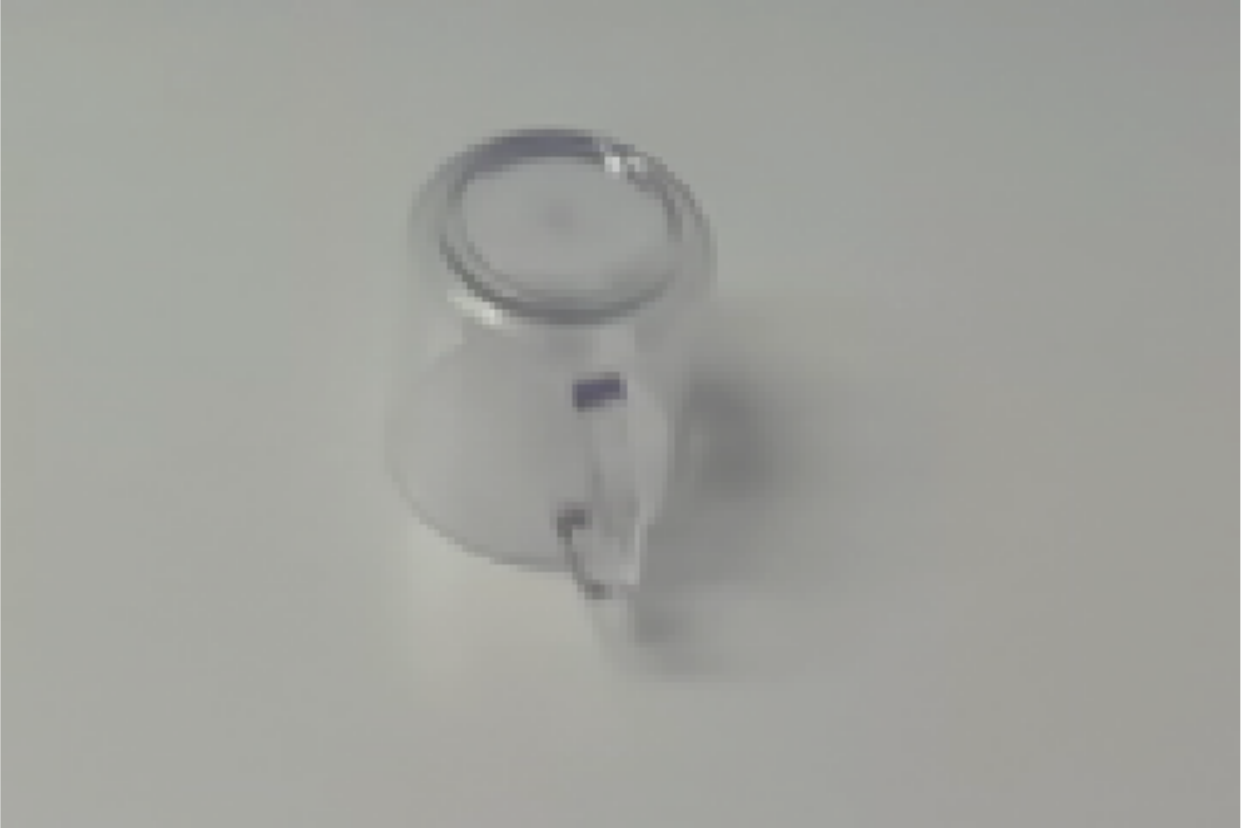} 
  }
  \subfloat{% 
    \includegraphics[height=\picheight\linewidth]{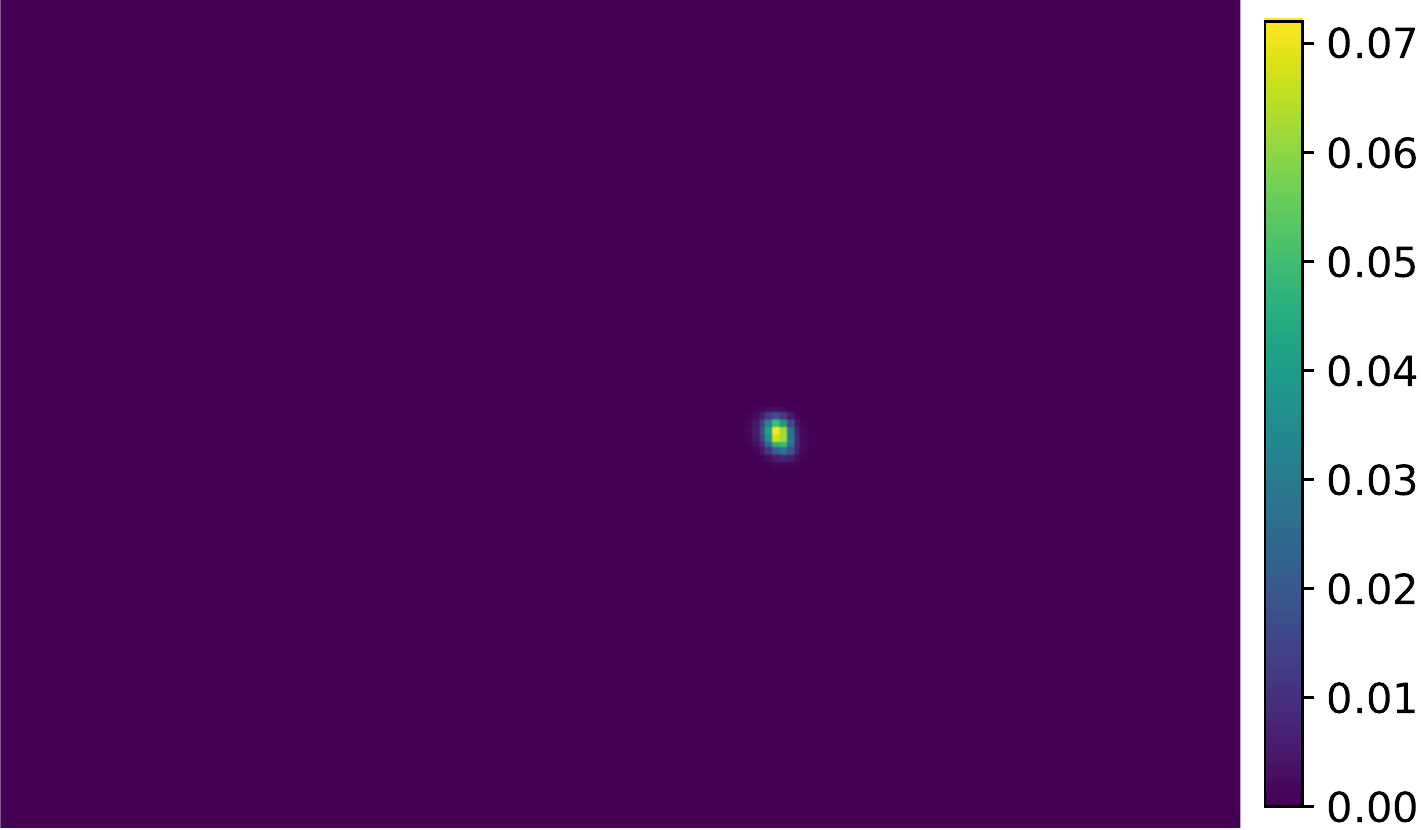} 
  }
  \subfloat{% 
    \includegraphics[height=\picheight\linewidth]{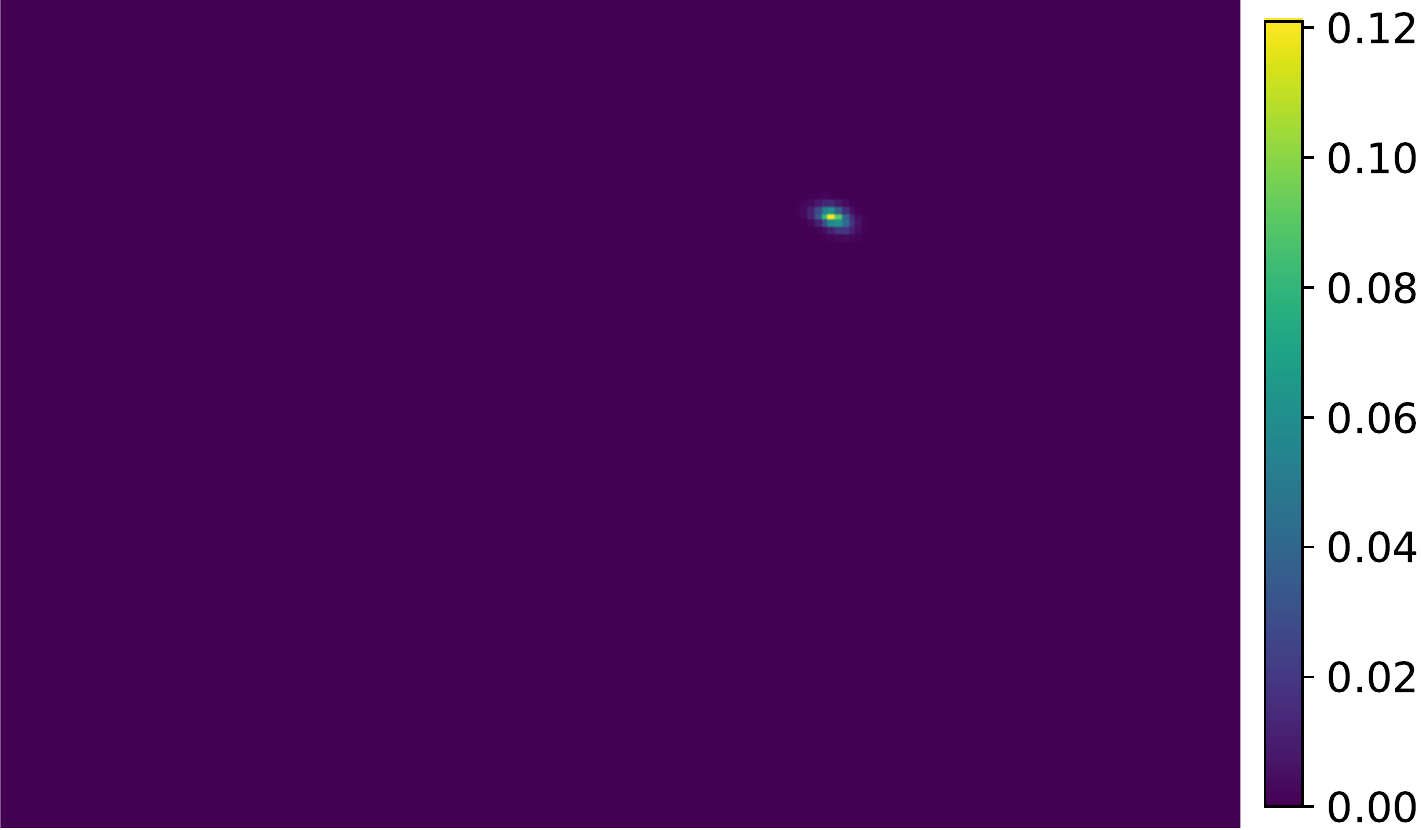} 
  }
  \subfloat{% 
    \includegraphics[height=\picheight\linewidth]{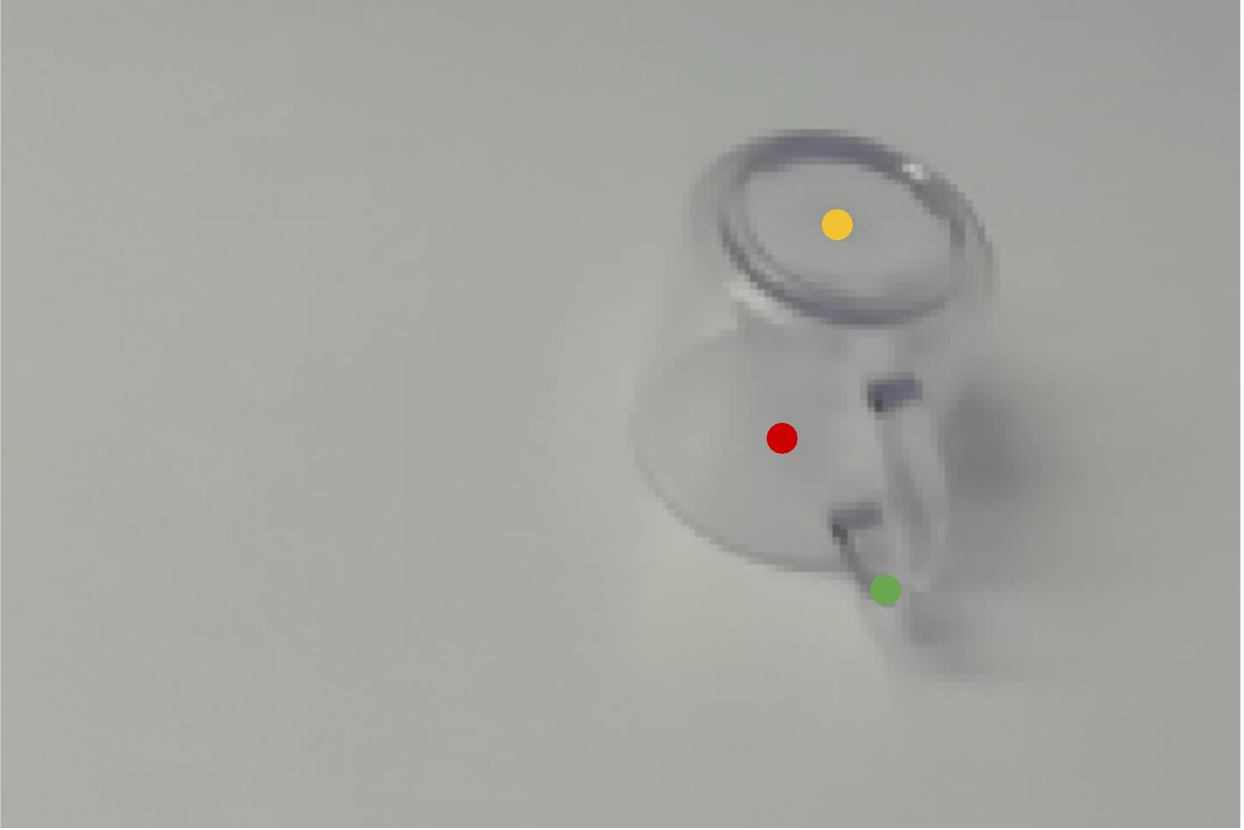} 
  }
  \\
  \vspace{-1.8ex}
  \subfloat{% 
    \includegraphics[height=\picheight\linewidth]{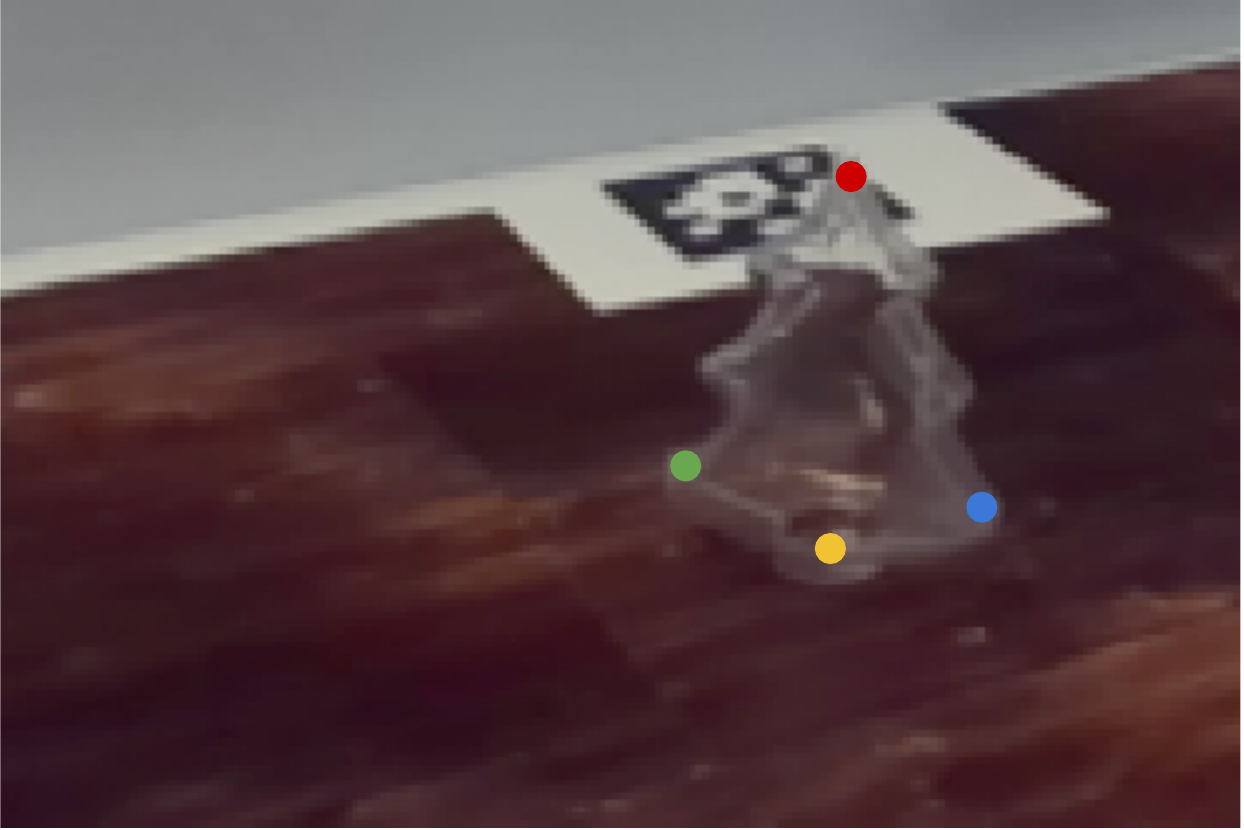}
  } 
  % \hspace{3.5ex} 
  \subfloat{% 
    \includegraphics[height=\picheight\linewidth]{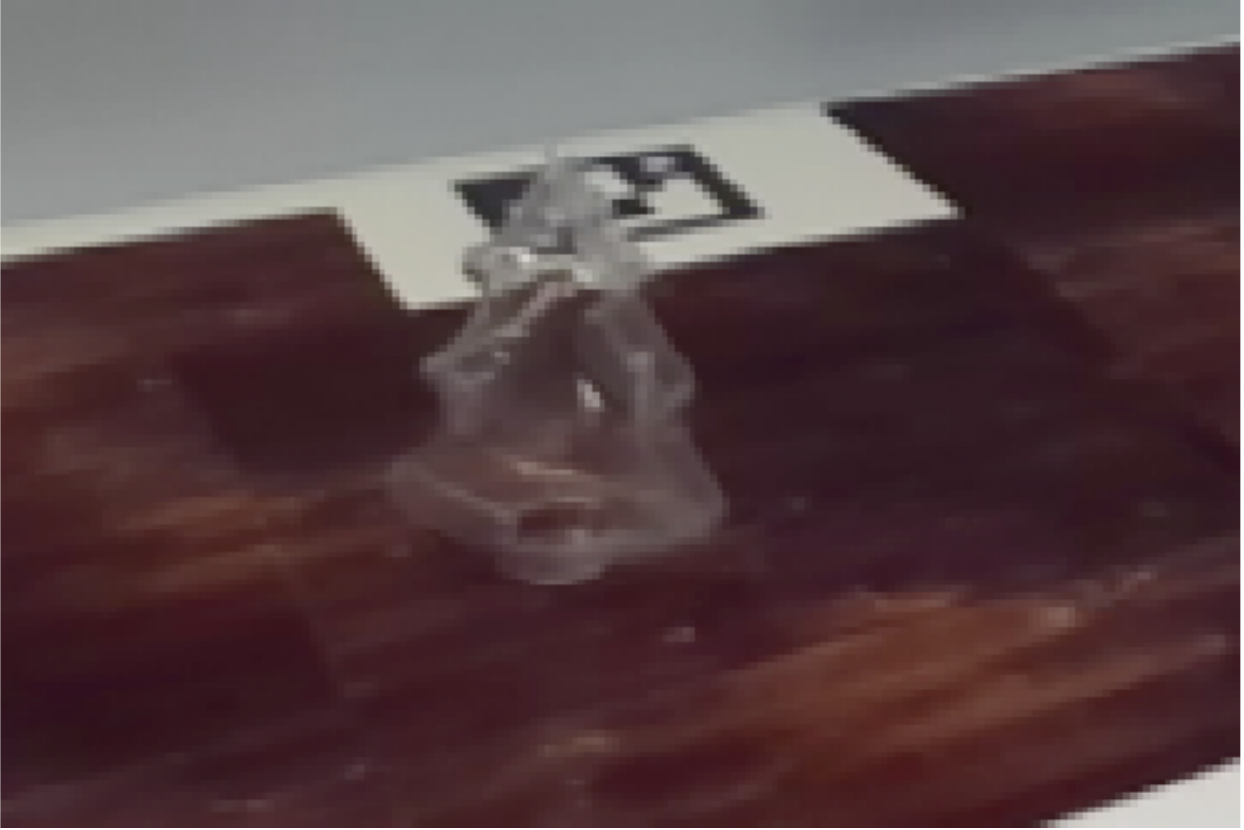} 
  }
  \subfloat{% 
    \includegraphics[height=\picheight\linewidth]{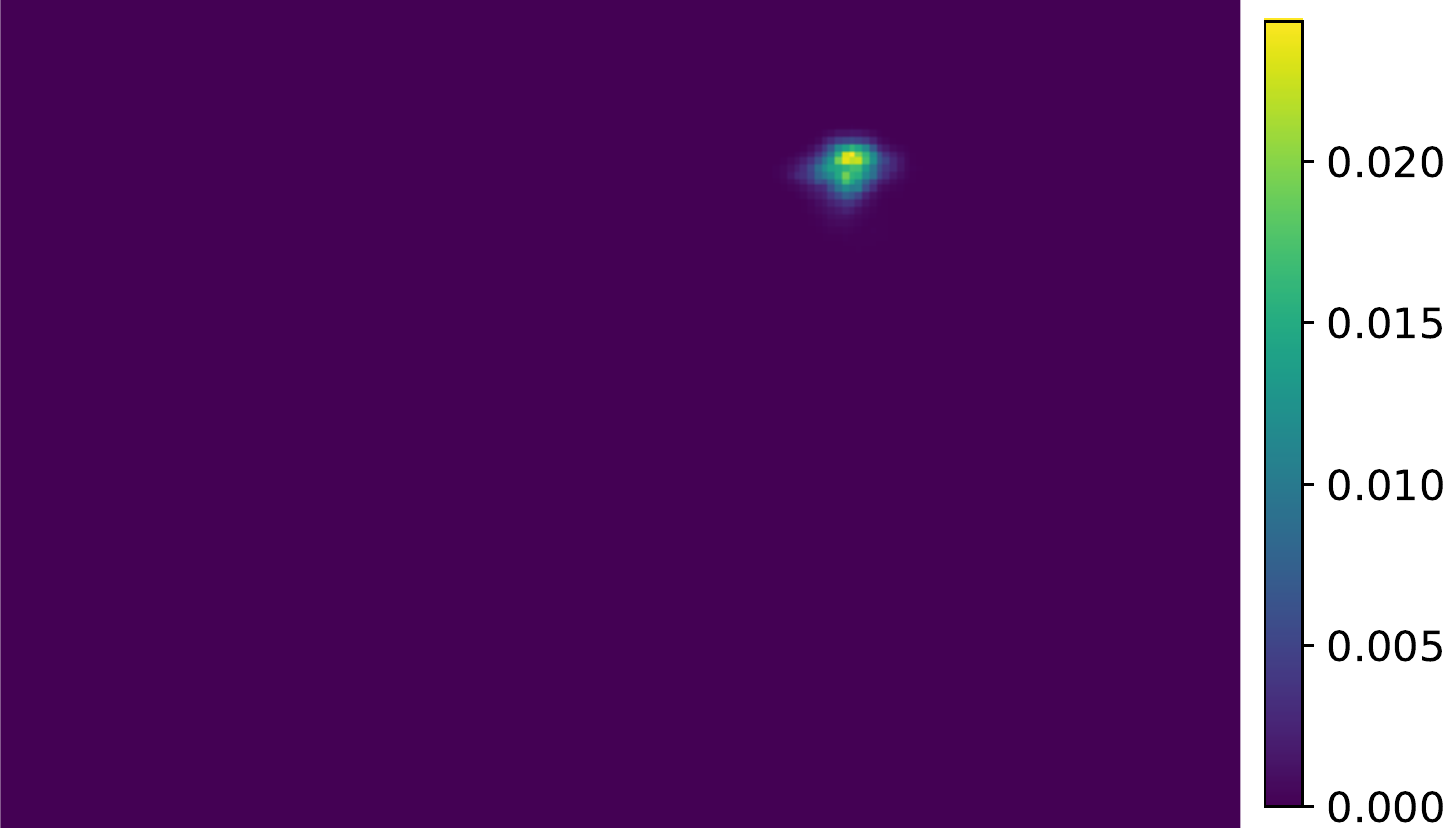} 
  }
  \subfloat{% 
    \includegraphics[height=\picheight\linewidth]{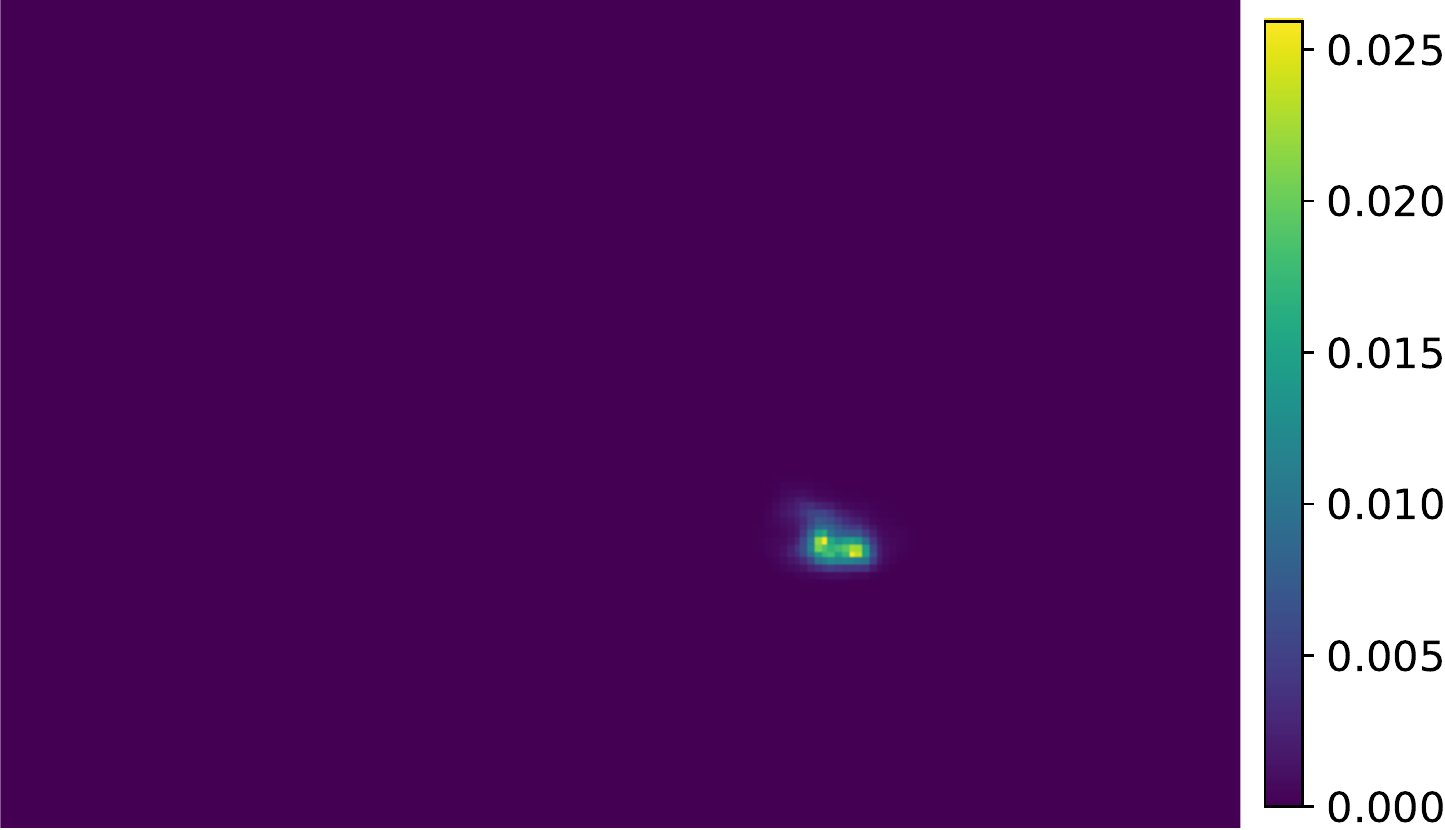} 
  }
  \subfloat{% 
    \includegraphics[height=\picheight\linewidth]{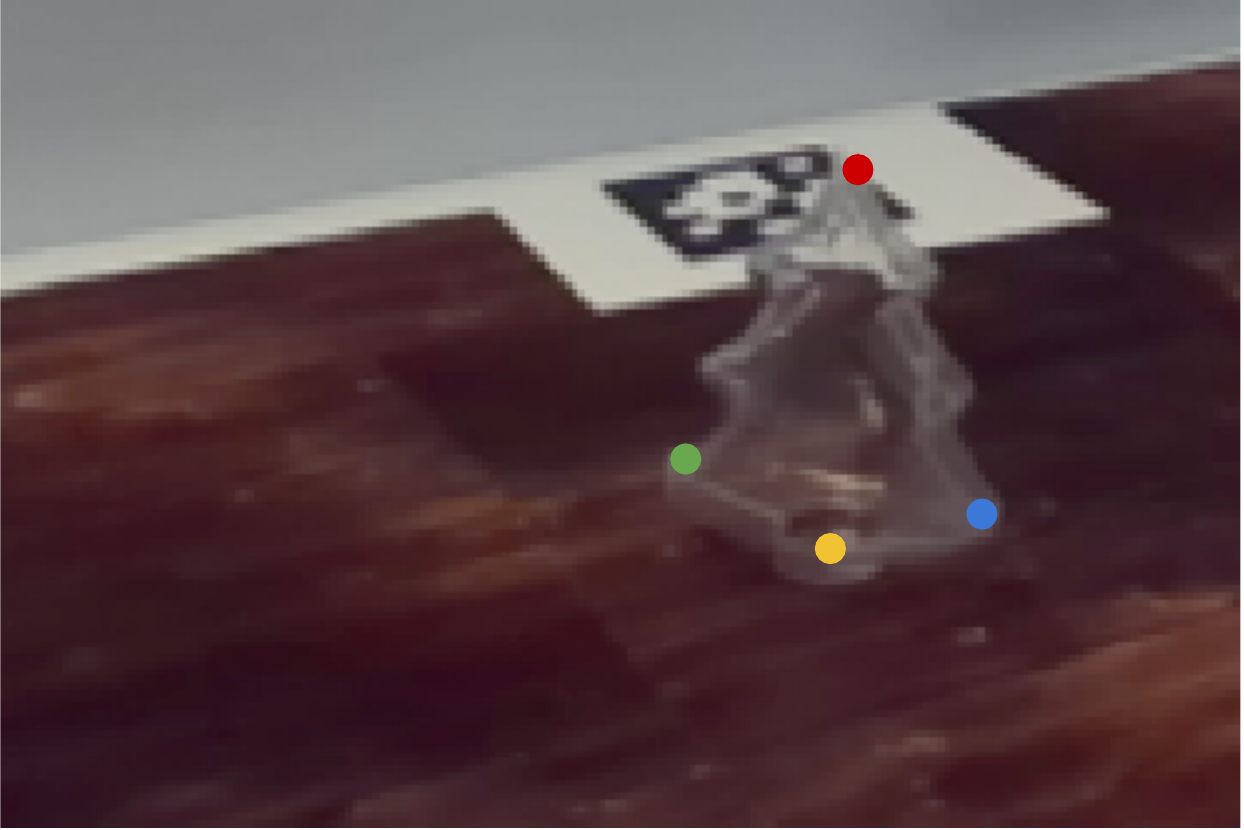} 
  }
  \\
  \vspace{-1ex}
  \caption{Visualization of prediction results on validation set. From left to right in each row: left stereo image with groundtruth keypoints, right stereo image, predicted probability map for first keypoint, predicted probability map for second keypoint, and predicted keypoints. We use \textcolor{viz_red}{red}, \textcolor{viz_yellow}{yellow}, \textcolor{viz_green}{green} and \textcolor{viz_blue}{blue} to mark keypoint \textcolor{viz_red}{1}, \textcolor{viz_yellow}{2}, \textcolor{viz_green}{3} and \textcolor{viz_blue}{4}.} 
  \label{fig:viz}
  \vspace{-1ex}
\end{figure*}

\subsection{Category-Level Pose Estimation}

We defined three categories: bottles (3 objects), bottles and cups (5 objects), and mugs (7 objects). 
For each category, we trained DenseFusion and KeyPose models, leaving out one texture over all objects as the test set.  Thus, this experiment captures how well a category-level model can generalize to any of its members in a new setting.  From the results in Table \ref{tab:single:unseen:texture}, KeyPose surpasses DenseFusion in accuracy by factors of 2 to 5.  Both methods seem to benefit from having larger numbers of samples for training. 

In a second category-level experiment, we held out mug$_0$ for testing; this experiment shows how well the methods generalize to unseen objects.  Given the small number of mugs in the category, the result shows surprisingly good generalization (Table \ref{tab:single:category}).
KeyPose is more accurate than DenseFusion(opaque and real), by a factor of 1.5.   With more objects in the mug category, it is likely both methods would improve.

\subsection{Ablation Studies}
\label{sec:ablation}

To find out which parts of KeyPose are effective, we performed an ablation study on the losses and architecture (stereo vs.\ monocular, early fusion vs.\ late fusion, regression vs.\ integration, projection loss, permutation loss), for a selection of instance and category models.  Results are in Tables \ref{tab:ablation:permutation},
\ref{tab:ablation:arch}.

First, note that \textbf{using stereo improves accuracy over monocular input by a factor of 2}, for both instance and category training.  Although monocular systems can gauge distance by the apparent size of an item, oblique views and different-size objects can make this difficult.  The most telling difference is in the disparity error, where it grows to almost a  pixel, while stereo is at half that.  This clearly demonstrates that stereo input is being used by the network to determine distance, and that keeping the disparity error low is the key to good 3D estimation.

\newcommand\chck{{\color{black}\ding{51}}}
\newcommand\chx{{\color{black}\ding{55}}}
\begin{table}[t]
\small
\centering
% \resizebox{\textwidth}{!}{
\begin{tabular}{l|l| c|c|c|c|c }
\hline
\multicolumn{2}{c|}{stereo} & \chx & \chck & \chck & \chck & \chck \\
\multicolumn{2}{c|}{early fusion} & \chck & \chx & \chck & \chck & \chck \\
\multicolumn{2}{c|}{projection loss} & \chck & \chck & \chx & \chck & \chck \\
\multicolumn{2}{c|}{direct regression} & \chck & \chck & \chck & \chx  & \chck\\
\hline

\multirow{3}{*}{bottle$_0$}
&  3D MAE (mm) &  10.0 &  7.9 &  5.4  &  4.7 &  \textbf{4.6} \\
& UV MAE (px) & 1.62 & \textbf{1.07} & 1.08  & 1.14 & 1.21 \\
& Disp MAE (px) & 0.91 & 0.67 & 0.45 & 0.38 & \textbf{0.36} \\
\hline
\multirow{3}{*}{bottles} 
&  3D MAE (mm) &  10.1 &  10.6 &  9.9  & 6.0 & \textbf{5.8} \\
& UV MAE (px) & 1.23 & 1.38 & \textbf{1.30} & 1.41 & 1.37 \\
& Disp MAE (px) & 0.94 & 0.96 & 0.89 & \textbf{0.47} & 0.48 \\
\hline
\end{tabular}
% }
\vspace{-1ex}
\caption{Ablation study for architecture and loss functions.}
\label{tab:ablation:arch}
% \vspace{-1ex}
\end{table}

\begin{table}[t]
\small
\setlength{\tabcolsep}{4.pt}
\centering
% \resizebox{\textwidth}{!}{
\begin{tabular}% {l| P{2cm} P{2cm} P{2cm} }
{l | ccc}
\hline
crop size & 180x120 & 270x180 & 360x240 \\
\hline
3D MAE (mm) &  \textbf{4.6} &  5.0 &  5.3 \\
\hline
\end{tabular}

\vspace{-1ex}
\caption{Ablation study on size of crop region for bottle$_0$.}
\label{tab:ablation:crop}
\vspace{-1ex}
\end{table}

There is similar but smaller difference between early and late fusion.  Recall that in late fusion (column 2), keypoints are computed for both left and right images, and then their U-values are compared to give the disparity.  Since the U-values have low error, the disparity values do, too.  However, they are higher than in early fusion, which can take advantage of mixing information from both images in the network.  We also observed a much longer error tail for late fusion, with some large metric errors.

The projection loss $\La_\mathit{proj}$ (column 3) helps to keep the disparity error low.  Without it, disparity errors are higher, having 0.09 pixels more in the instance case, and 0.41 pixels more in the category case.  The \textit{UV} error is actually lower when not using the projection loss, but it is less important.  While 0.41 pixels may not seem like a large difference, it can have an outsize effect on metric error.  From stereo geometry, the change of depth for a change of disparity is given by:
\begin{equation}
\frac{\Delta z}{\Delta d} = - \frac{z^2}{fb}
\end{equation}
where $f$ is the focal length and $b$ is the baseline.  At an object distance of 0.8 m, for example, 0.41 pixel error in disparity yields a 5.5 mm error in depth for our stereo system.

The difference between using direct regression to \uvd\ values, vs.\ an integral approach, shows a small bias in favor of regression.  The advantage of the integral approach is the production of \textit{UV} and disparity maps, which is useful for visualization of the network predictions (see Figures \ref{fig:early}, \ref{fig:viz}).

\begin{figure}[t]
    \centering
    \includegraphics[width=0.23\textwidth]{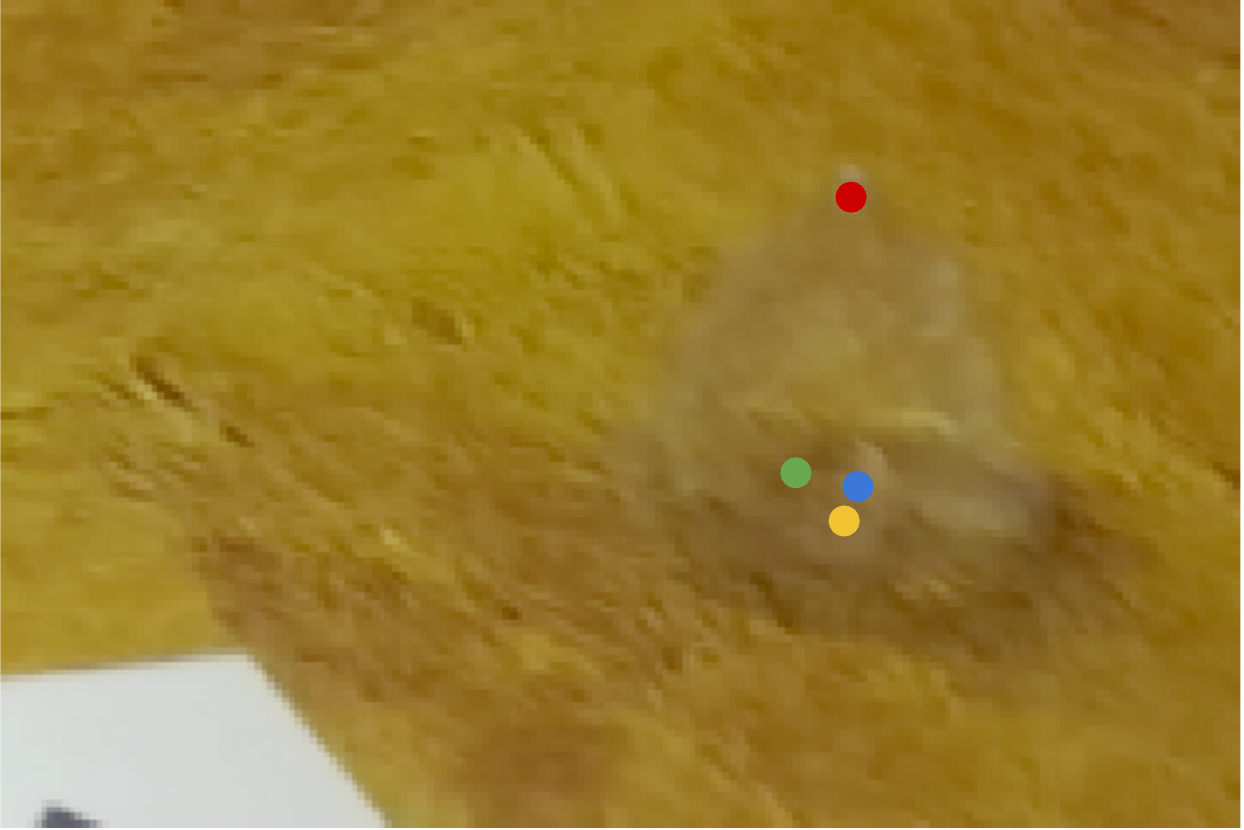}
    \includegraphics[width=0.23\textwidth]{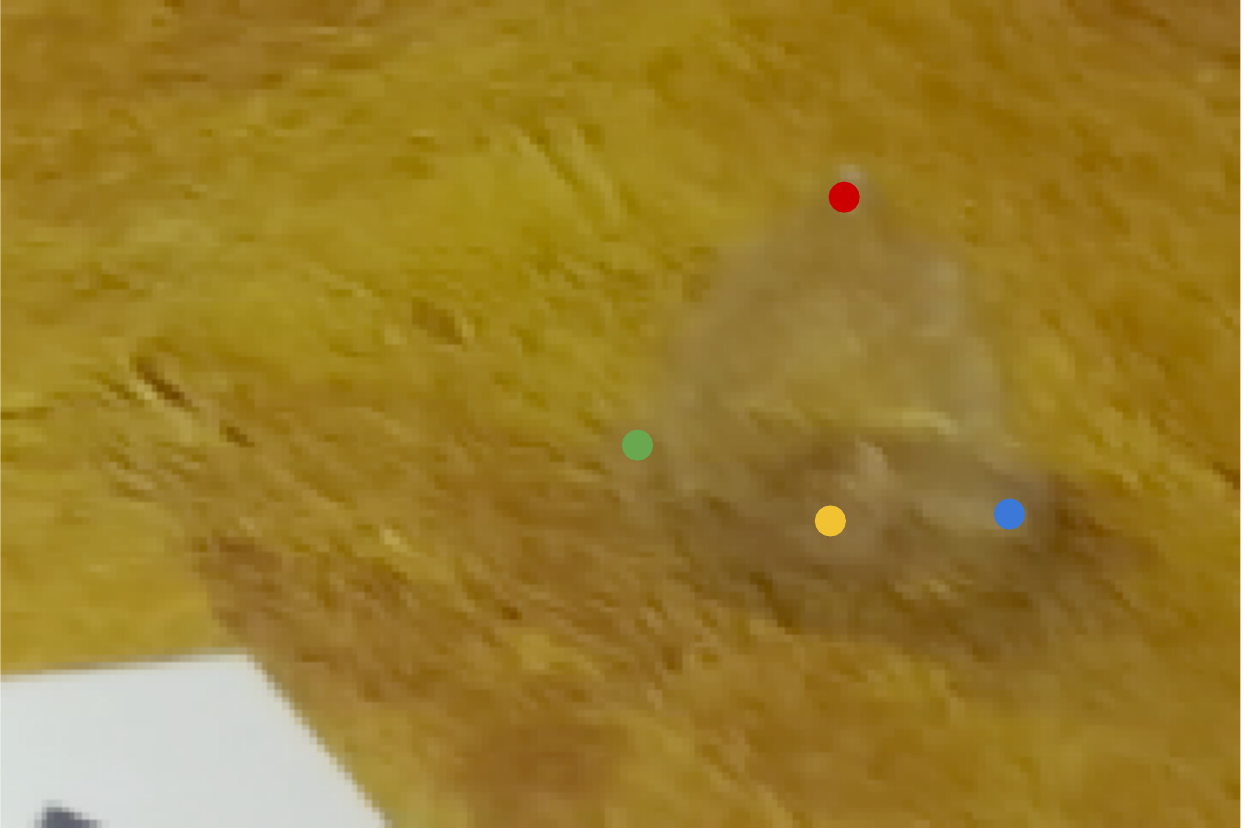}
    \vspace{-1ex}
    \caption{Visualization of ablation study: without (left) vs. with (right) permutation loss. We use \textcolor{viz_red}{red}, \textcolor{viz_yellow}{yellow}, \textcolor{viz_green}{green} and \textcolor{viz_blue}{blue} to mark keypoint \textcolor{viz_red}{1}, \textcolor{viz_yellow}{2}, \textcolor{viz_green}{3} and \textcolor{viz_blue}{4}. Instance tree$_0$ has symmetric keypoints \textcolor{viz_green}{3} \& \textcolor{viz_blue}{4}.}
    \label{fig:permutation}
    % \vspace{-1ex}
\end{figure}

For the permutation loss, results are in Table \ref{tab:ablation:permutation}.  We examined the tree object and turned off the permutation of the side keypoints.  Figure \ref{fig:permutation} shows the effect: since the sides of the tree are symmetric, the choice of which keypoint will be labeled is random.  Training without the permutation loss causes the two keypoints to cluster in the center to minimize loss.  This is reflected in the wide difference between the results.

Finally, we consider whether the results depend on a tight crop of the object.  First note that the crop is generous, especially for small objects (see Figure \ref{fig:viz}).  Then, we dither the location of the object by 20 pixels to make KeyPose robust to bounding box placement.  We also checked larger crops in Table \ref{tab:ablation:crop}, up to 4 times the original area.  The results show minimal degradation, less than any of the loss ablations in Table \ref{tab:ablation:arch}.
Many CNN methods use a tight crop of an object followed by scaling to present the same size to the network.  Here we have chosen the harder problem and used a fixed size crop with no rescaling.  The apparent size of objects varies by a factor of about 2.5, which is reasonable for a lot of applications, such as bin-picking with fixed cameras.  It remains to future work to see if tight crops and scaling would be more accurate.
 
\begin{table}[t]
\small
\setlength{\tabcolsep}{4.pt}
\centering
% \resizebox{\textwidth}{!}{
\begin{tabular}% {l| P{2cm} P{2cm} P{2cm} }
{l | c|ccc}
\hline
\multicolumn{2}{c|}{metrics} & 3D MAE (mm) & UV MAE (px) & Disp MAE (px) \\
\hline
\multirow{2}{*}{\begin{tabular}[c]{@{}c@{}} perm \\ loss \end{tabular}} & {\color{black} \ding{55}} &  26.4 & 11.1 & 1.46 \\
\hhline{~----}
 & {\color{black} \ding{51}} & \textbf{12.8} & \textbf{2.79} & \textbf{1.05} \\
\hline
\end{tabular}
% }
\vspace{-1ex}
\caption{Ablation study on tree$_0$ for permutation loss.}
\label{tab:ablation:permutation}
\vspace{-1ex}
\end{table}

\section{Conclusion and Discussion}

In this paper, we studied the problem of estimating the 3D object pose represented by 3D keypoint locations from stereo images.  By providing an easy-to-use 3D keypoint labeling facility, we have generated TOD, a large-scale labeled dataset of transparent objects, along with registered depth, for training and comparing keypoint pose estimation methods.  The KeyPose model, utilizing early fusion of stereo images, surpasses state-of-the-art on all  benchmark tests in instance and category levels, including when opaque depth is used.  It generalizes across unseen textures and to unseen objects.  The ablation studies validate our emphasis on early fusion and multi-view reprojection losses.  

There are some areas that need further improvement and exploration.  Among these are detecting transparent objects using our heatmap technique, adding more complex backgrounds, varying lighting, and including multi-object samples into the dataset. We will also investigate using mobile robots to capture data in the wild.
Although we concentrated on transparent rigid objects, KeyPose can also be applied to opaque, articulated and deformable objects. 
These directions will be left as future work.

\clearpage

\begin{figure*}[h]
  \centering
  \subfloat{%
  \includegraphics[width=\linewidth]{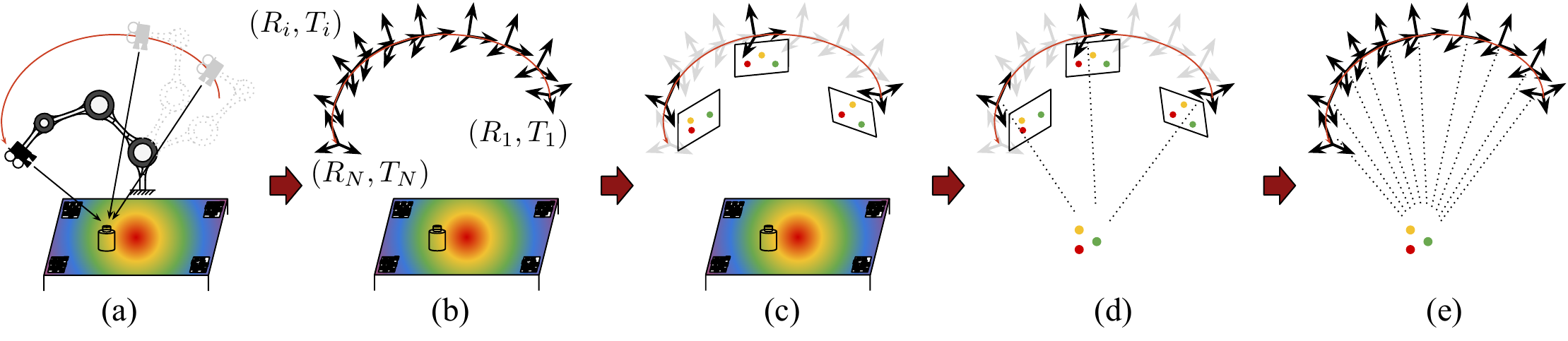} 
  }
  \captionof{figure}{Labeling pipeline. (a) We use robot to scan the object   
from different views and record stereo-RGB and RGBD sequences; (b) AprilTags groundtruth locations are used to calculate the global pose of 
each frame ($1$ through $N$) in the video based on Perspective-n-Point (PnP) algorithm; (c) Only a few key frames selected from the video   
sequence are labeled, where the selection is based farthest point sampling (FPS) of camera poses; (d) From the labeled 2D locations of the  
keypoints, the 3D locations of the keypoints are calculated; (e) The 3D locations are propagated to all frames in the sequence to obtain    
the 2D projected $UV$ location and depth.}
     \label{fig:labeling:pipeline}
\end{figure*}

{\small
\bibliographystyle{ieee_fullname}
\bibliography{egbib}
}

\appendix

\section*{Supplementary}

\begin{figure*}[h]
  \centering
  \subfloat{%
  \includegraphics[width=0.195\linewidth]{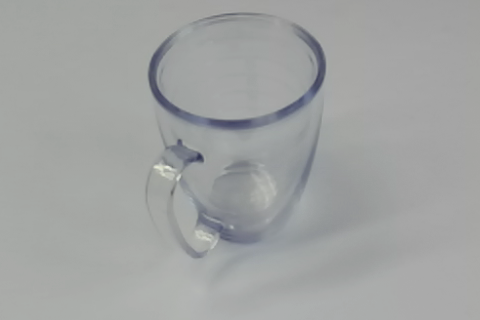} 
  }
  \subfloat{%
  \includegraphics[width=0.195\linewidth]{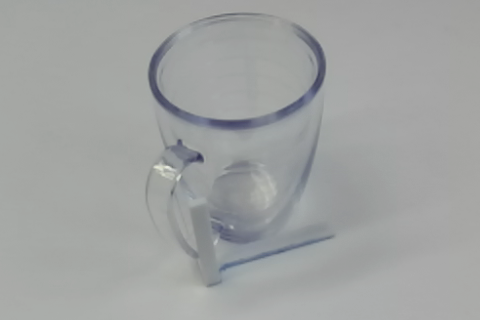}
  }
  \subfloat{%
  \includegraphics[width=0.195\linewidth]{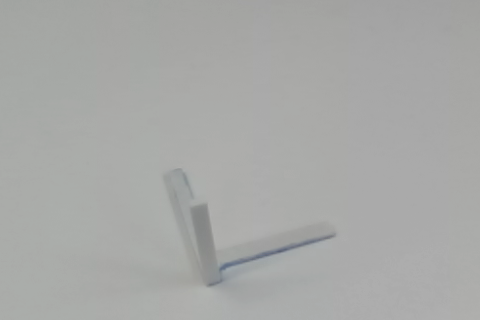}
  }
  \subfloat{%
  \includegraphics[width=0.195\linewidth]{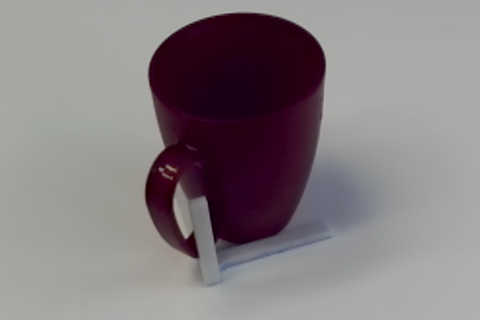}
  }
  \subfloat{%
  \includegraphics[width=0.195\linewidth]{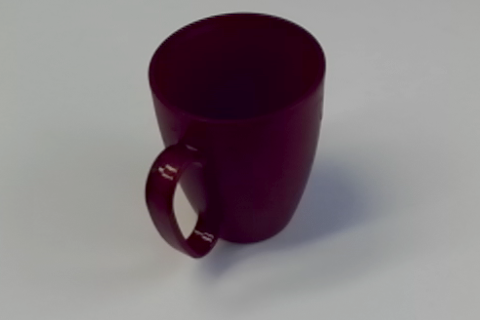}
  }
  \captionof{figure}{Illustration of our method of using marker to replace the transparent objects with its opaque twin. The white plastic marker was produced by 3D printing. }
  \label{fig:replacement}
\end{figure*}

%%%%%%%%% BODY TEXT
\section{Overview}

In this document, we provide additional detail on KeyPose as presented in the main paper. We present object examples in Section \ref{sec:object:examples}. We present details of data capturing pipeline and error analysis in Section \ref{sec:data:capturing}. We also provide details of the architectures, training procedure, and timing in Section \ref{sec:arch}.

\section{Object and Texture Examples}
\label{sec:object:examples}

\textbf{Placement of Object Twins.} 
Capturing the groundtruth depth of the transparent object requires placing its opaque twin at the same pose. 
We proposed an efficient way to accurately do so. 
The process is illustrated in Figure \ref{fig:replacement}, where we show the replacement of mug$_3$.

We first place the transparent object at a desired pose and scan the RGBD and stereo RGB video.
Then we align a specially designed plastic marker closely to the object. 
The plastic marker consists of three sticks orthogonal to each other so that the marker's relative configuration with respect to the object is unique.
After alignment, we remove the transparent object but keep the marker from moving.
Next, we place the opaque twin so that it closely aligns with the marker in the same configuration as the transparent object.
Finally, we remove the marker but keep the opaque twin from moving. In this case, the transparent object and the opaque twin will have exactly the same pose.

\begin{figure}[h]
  \centering
  \subfloat{%
  \includegraphics[width=0.49\linewidth]{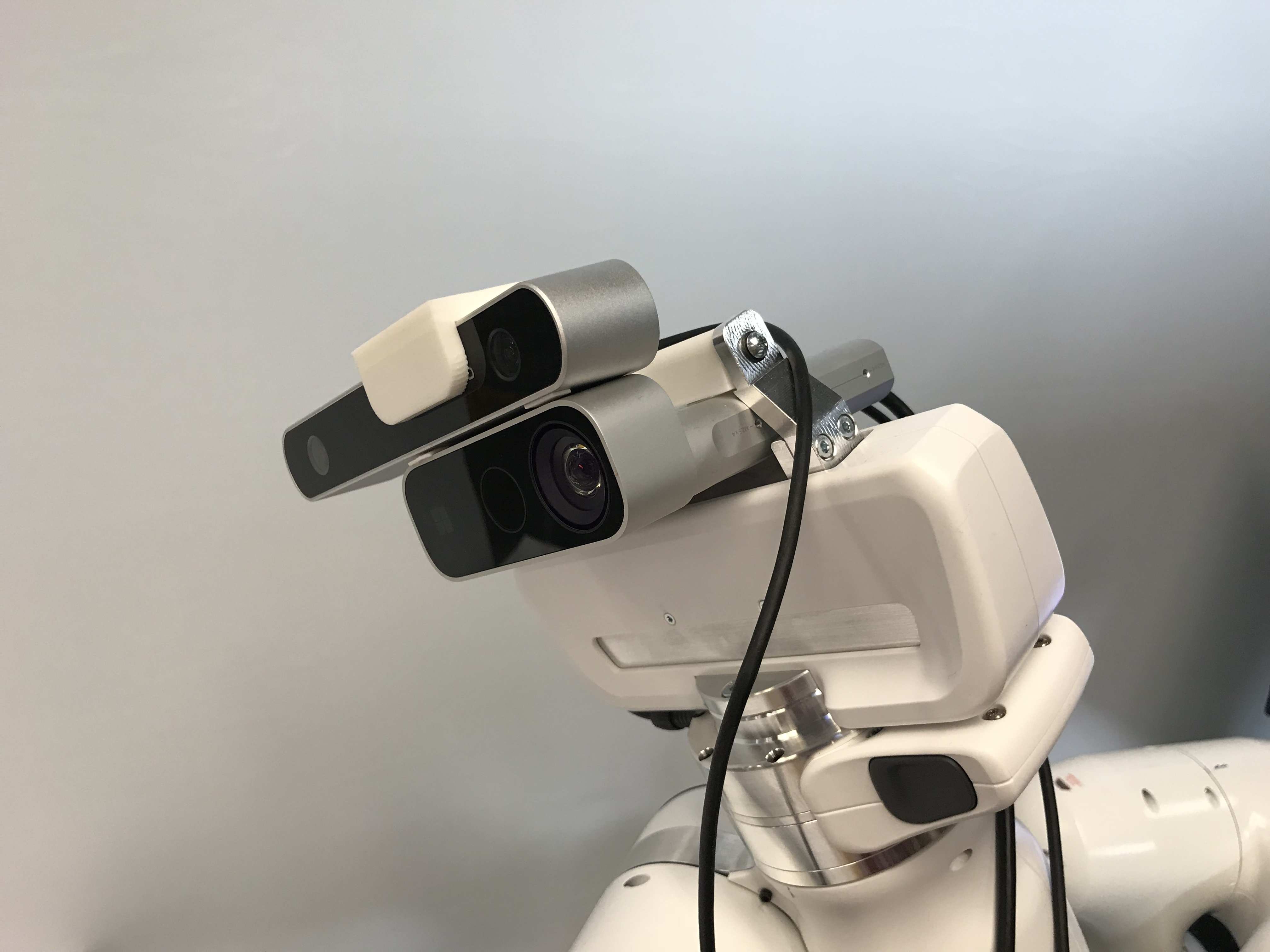} 
  }
  \subfloat{%
  \includegraphics[width=0.49\linewidth]{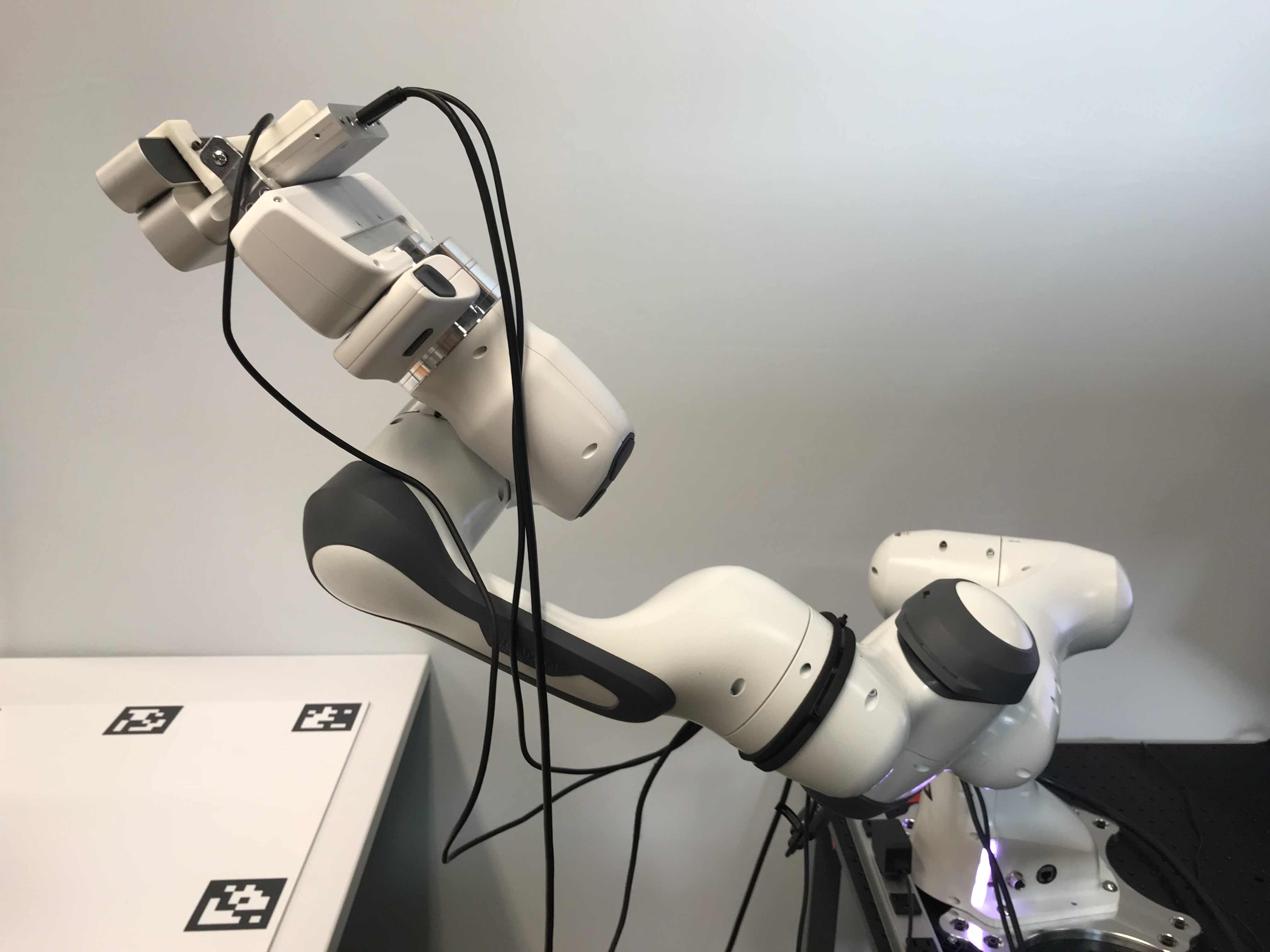} 
  }
  \caption{Robot configuration. Left: we mount both ZED stereo camera (top) and Microsoft Azure Kinect camera (bottom) on the end-effector of the robot, with a plastic fixture produced by 3D printing. Right: Franka Panda arm used to capture data.}
  \label{fig:robot:config}
\end{figure}

\begin{figure*}[h!] 
  \centering
  \subfloat{% 
  \includegraphics[width=0.9\linewidth]{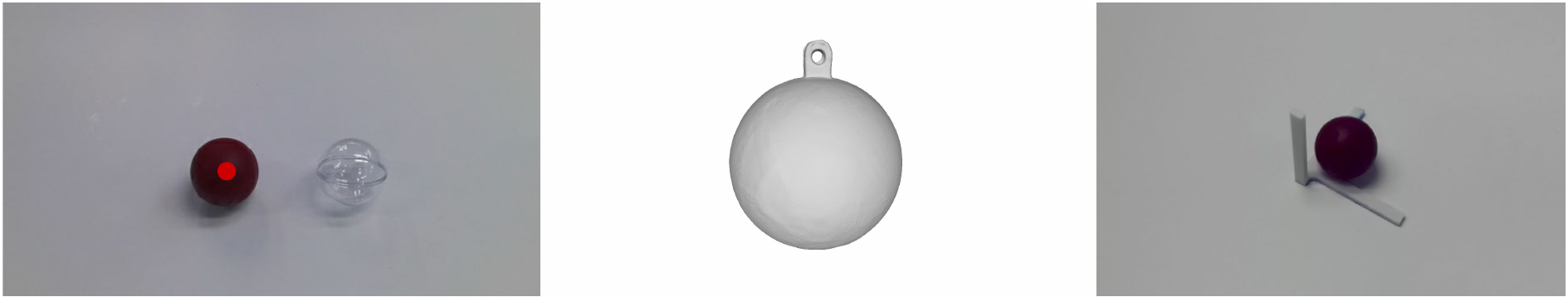} 
  }
  \\
  \subfloat{% 
  \includegraphics[width=0.9\linewidth]{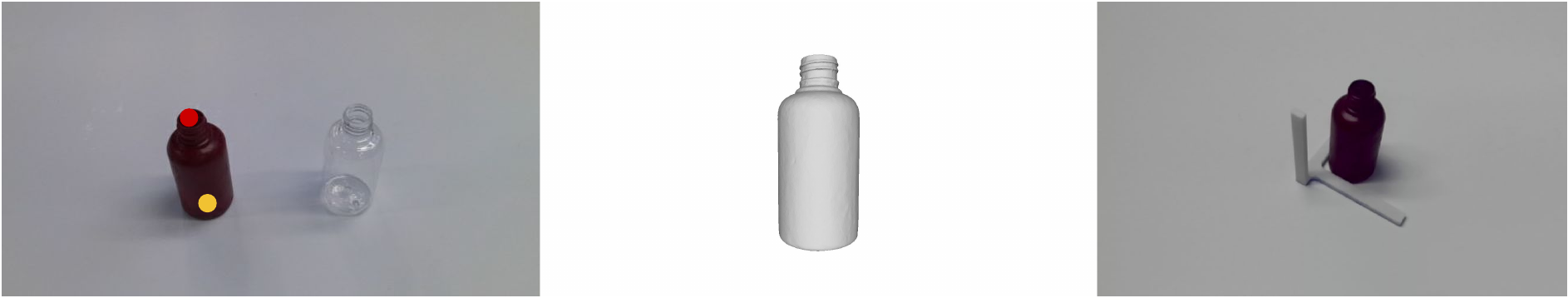} 
  }
  \\
  \subfloat{%
  \includegraphics[width=0.9\linewidth]{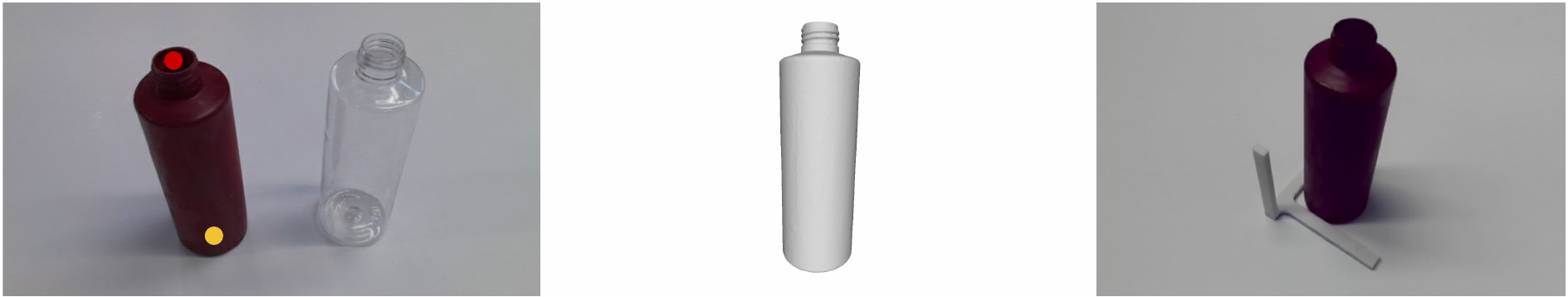} 
  }
  \\
  \subfloat{%
  \includegraphics[width=0.9\linewidth]{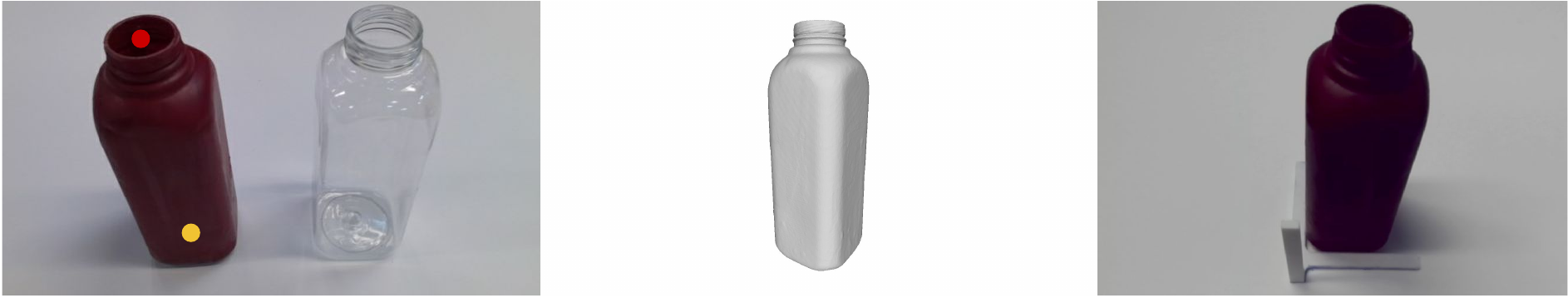} 
  }
  \\
  \subfloat{%
  \includegraphics[width=0.9\linewidth]{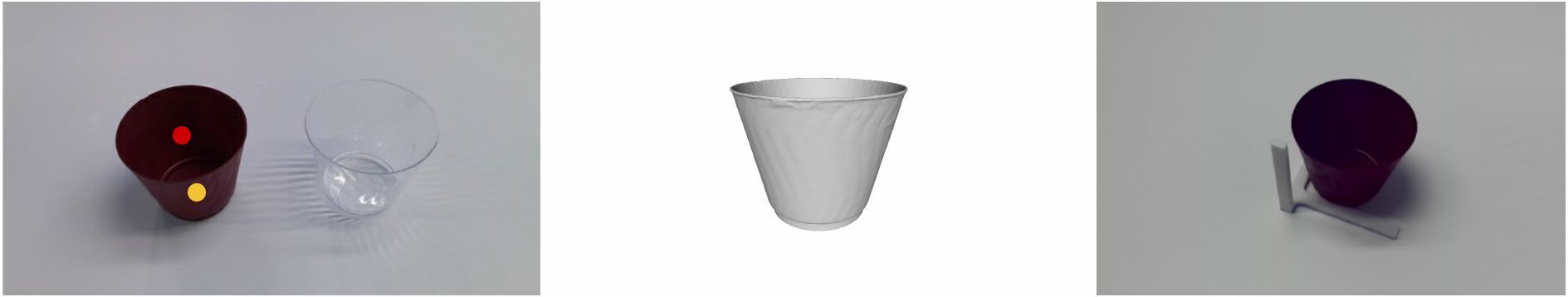} 
  }
  \\
  \subfloat{%
  \includegraphics[width=0.9\linewidth]{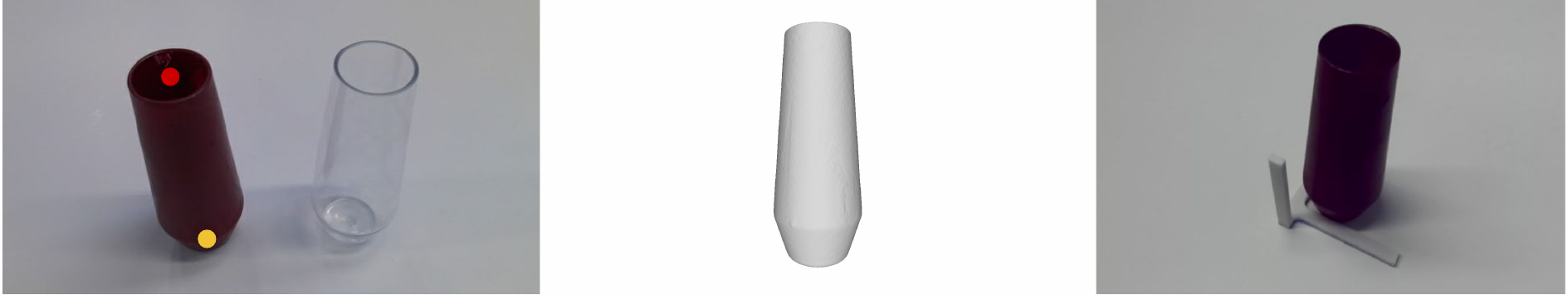} 
  }
  \caption{Visualization of ball$_0$, bottle$_0$, bottle$_1$, bottle$_2$, cup$_0$, cup$_1$ from top to bottom. From left to right in each row: object twin with groundtruth keypoint location, scanned CAD model from opaque object, and aligning the marker to the object. We use \textcolor{viz_red}{red} and \textcolor{viz_yellow}{yellow} to mark keypoint \textcolor{viz_red}{1} and \textcolor{viz_yellow}{2}.}
  \label{fig:viz:object:1}

\end{figure*}

\begin{figure*}[h!] 
  \centering
  \subfloat{%
  \includegraphics[width=0.9\linewidth]{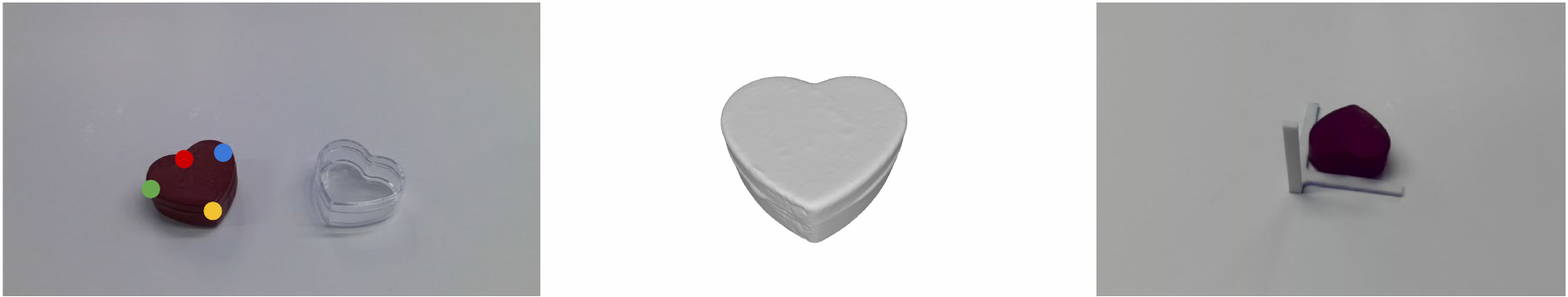} 
  }
  \\
  \subfloat{%
  \includegraphics[width=0.9\linewidth]{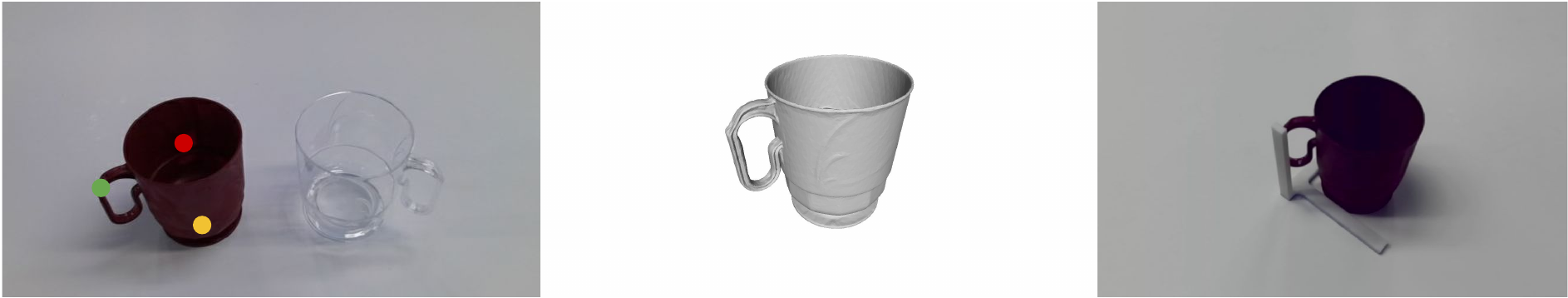} 
  }
  \\
  \subfloat{%
  \includegraphics[width=0.9\linewidth]{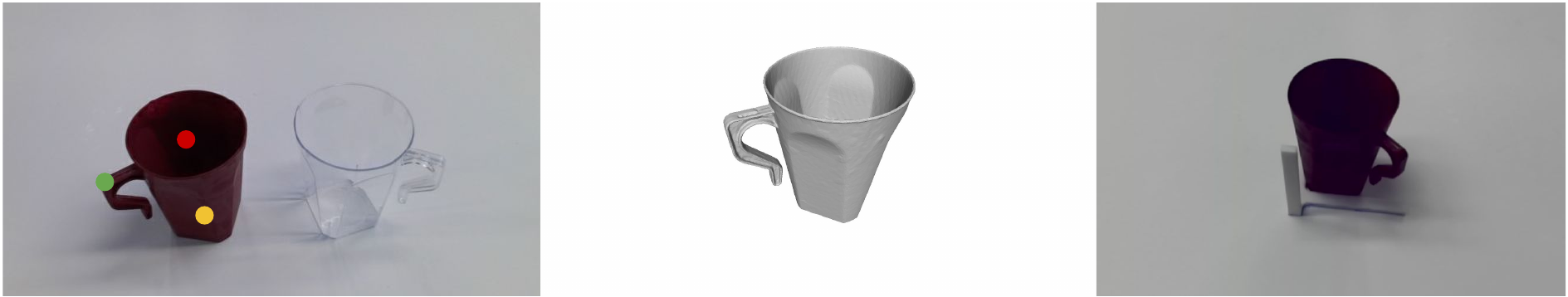} 
  }
  \\
  \subfloat{%
  \includegraphics[width=0.9\linewidth]{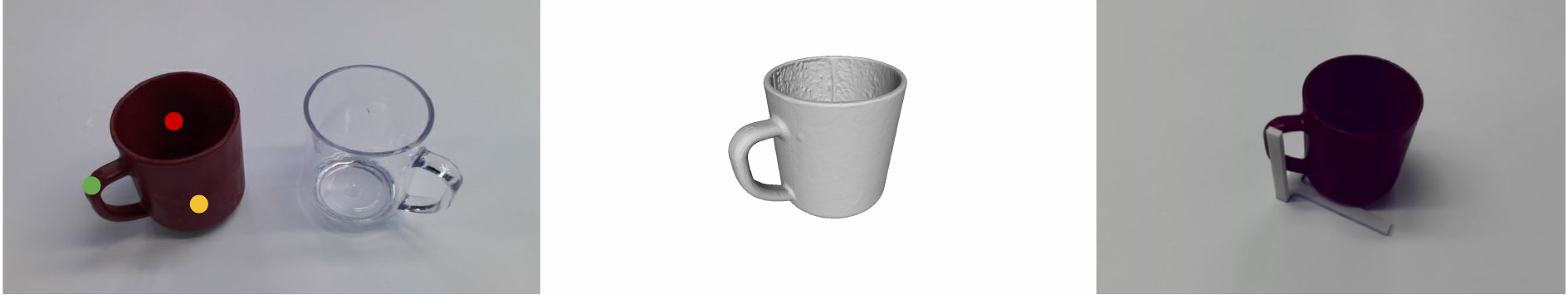} 
  }
  \\
  \subfloat{%
  \includegraphics[width=0.9\linewidth]{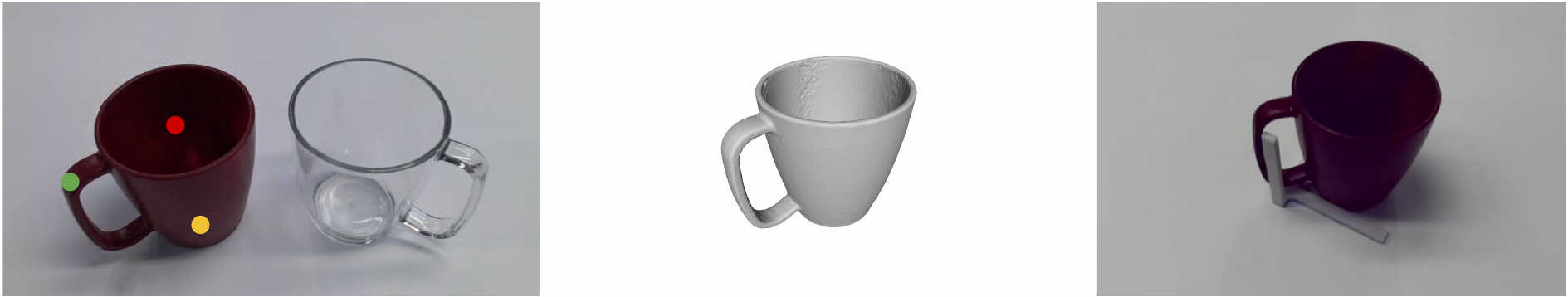} 
  }
  \\
  \subfloat{%
  \includegraphics[width=0.9\linewidth]{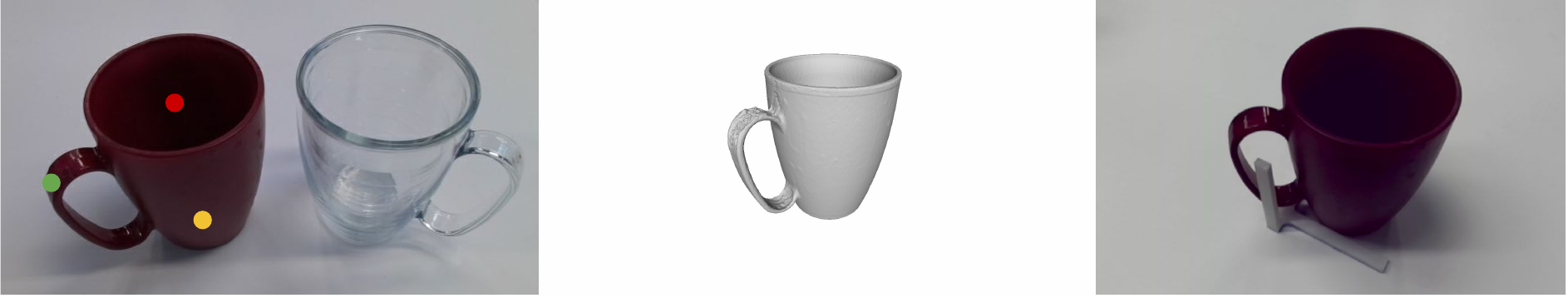} 
  }
  \caption{Visualization of heart$_0$, mug$_0$, mug$_1$, mug$_2$, mug$_3$, mug$_4$ from top to bottom. From left to right in each row: object twin with groundtruth keypoint location, scanned CAD model from opaque object, and aligning the marker to the object. We use \textcolor{viz_red}{red}, \textcolor{viz_yellow}{yellow}, \textcolor{viz_green}{green} and \textcolor{viz_blue}{blue} to mark keypoint \textcolor{viz_red}{1}, \textcolor{viz_yellow}{2}, \textcolor{viz_green}{3} and \textcolor{viz_blue}{4}. For heart$_0$, keypoints \textcolor{viz_green}{3} and \textcolor{viz_blue}{4} are symmetric.}
  \label{fig:viz:object:2}
\end{figure*}

\begin{figure*}[h!]
  \centering
  \subfloat{%
  \includegraphics[width=0.9\linewidth]{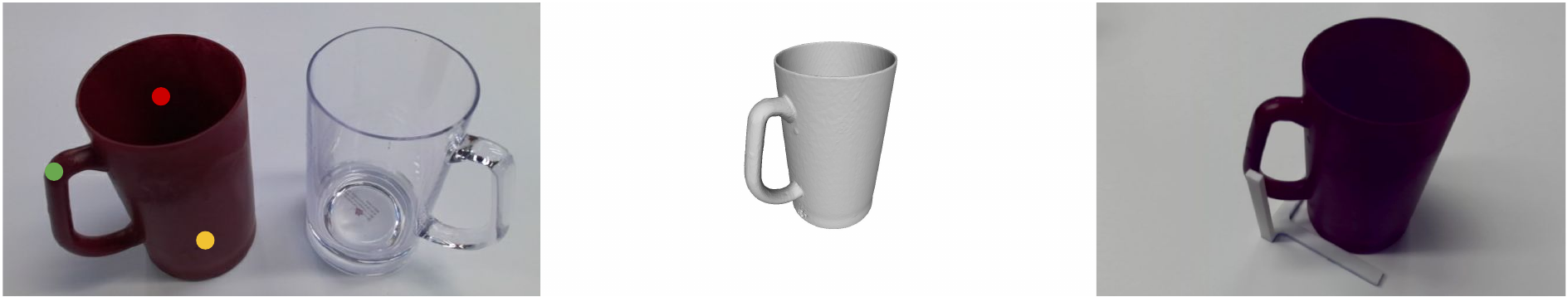} 
  }
  \\
  \subfloat{%
  \includegraphics[width=0.9\linewidth]{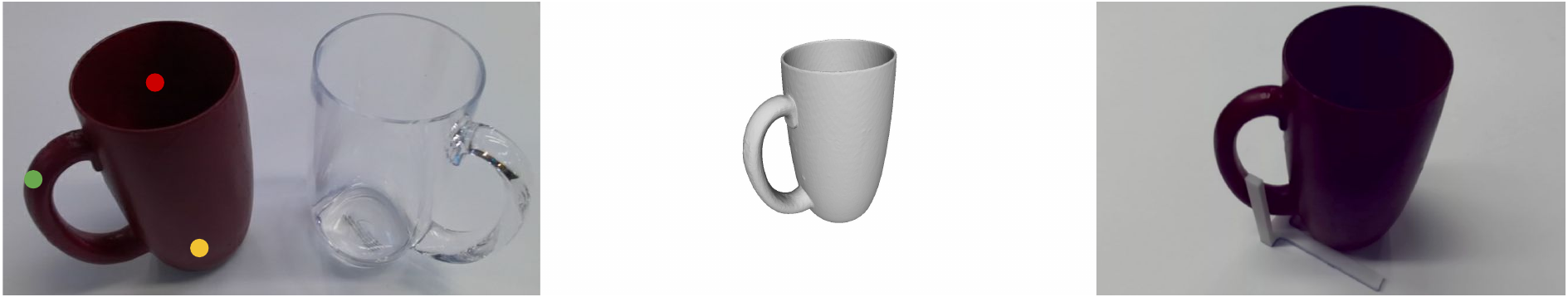} 
  }
  \\
  \subfloat{%
  \includegraphics[width=0.9\linewidth]{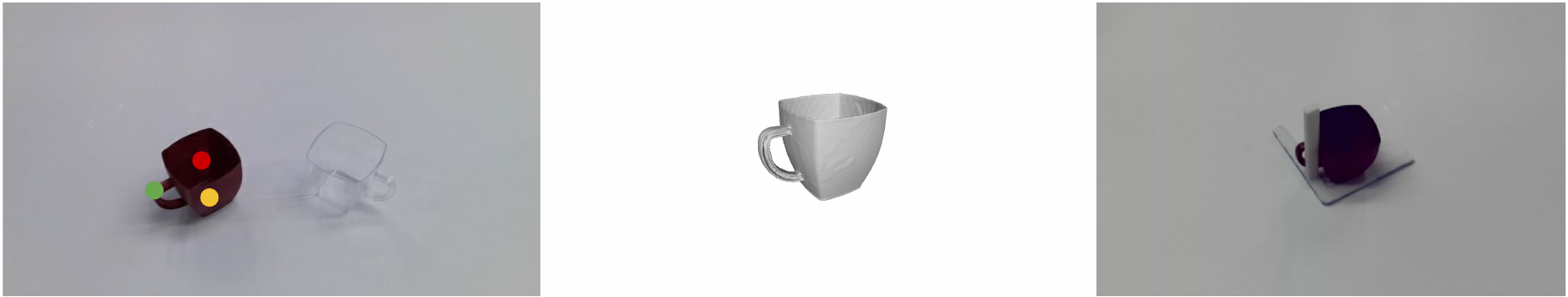} 
  }
  \\
  \subfloat{%
  \includegraphics[width=0.9\linewidth]{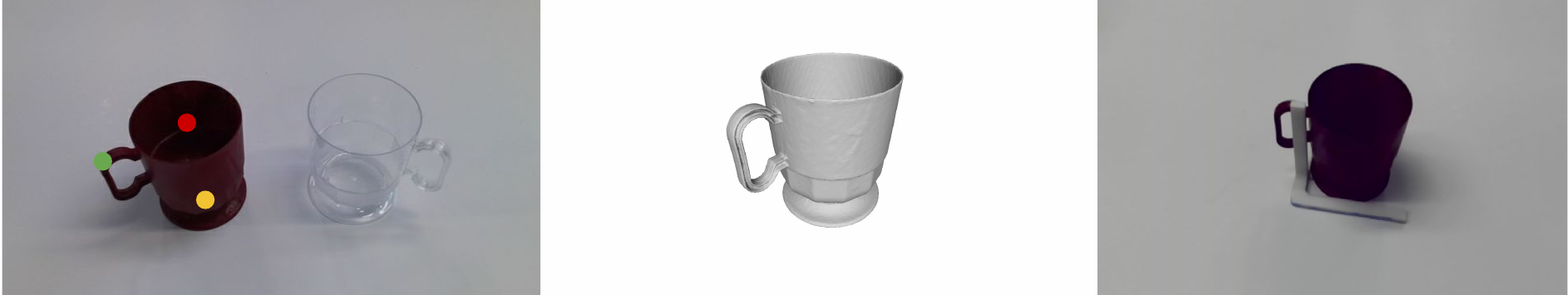} 
  }
  \\
  \subfloat{%
  \includegraphics[width=0.9\linewidth]{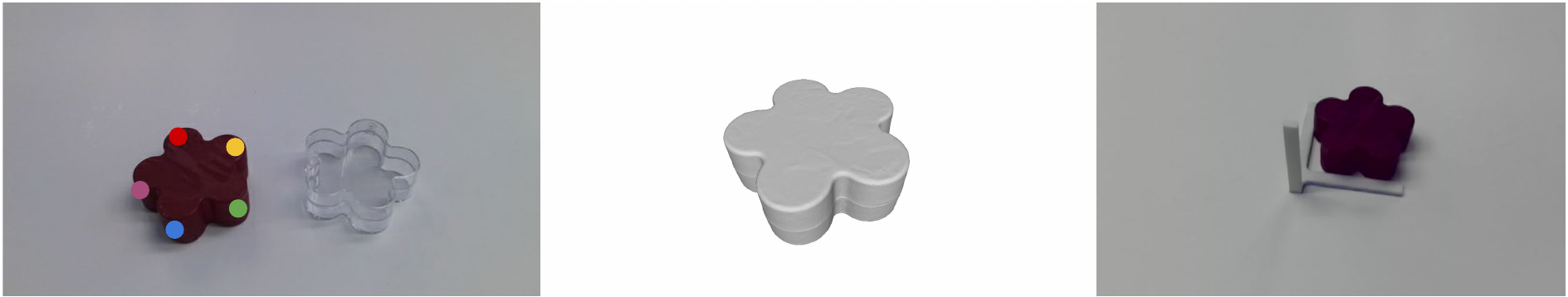} 
  }
  \\
  \subfloat{%
  \includegraphics[width=0.9\linewidth]{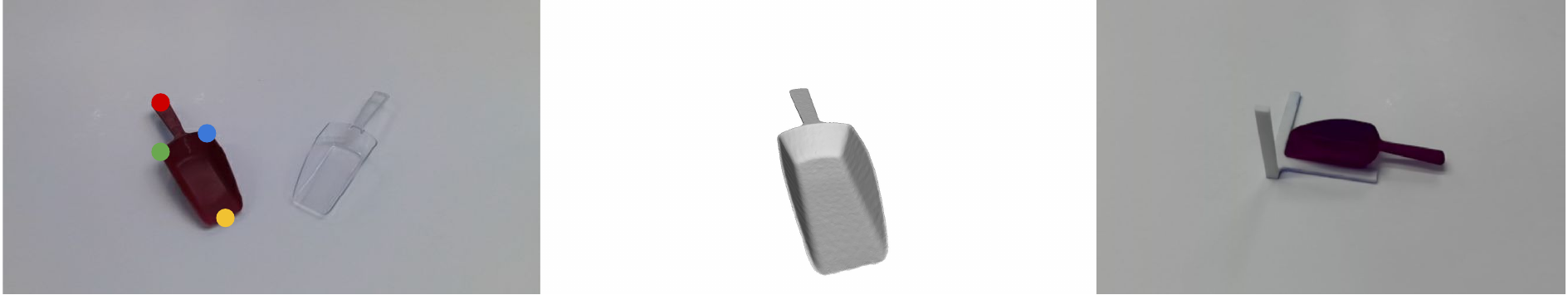} 
  }
  \caption{Visualization of mug$_5$, mug$_6$, mug$_7$, mug$_8$, sakura$_0$, shovel$_0$ from top to bottom. From left to right in each row: object twin with groundtruth keypoint location, scanned CAD model from opaque object, and aligning the marker to the object. We use \textcolor{viz_red}{red}, \textcolor{viz_yellow}{yellow}, \textcolor{viz_green}{green}, \textcolor{viz_blue}{blue} and \textcolor{viz_purple}{purple} to mark keypoint \textcolor{viz_red}{1}, \textcolor{viz_yellow}{2}, \textcolor{viz_green}{3}, \textcolor{viz_blue}{4} and 
  \textcolor{viz_purple}{5}. For sakura$_0$, all its five keypoints are symmetric.}
  \label{fig:viz:object:3}
\end{figure*}

\begin{figure*}[h!]
  \centering
  \subfloat{%
  \includegraphics[width=0.9\linewidth]{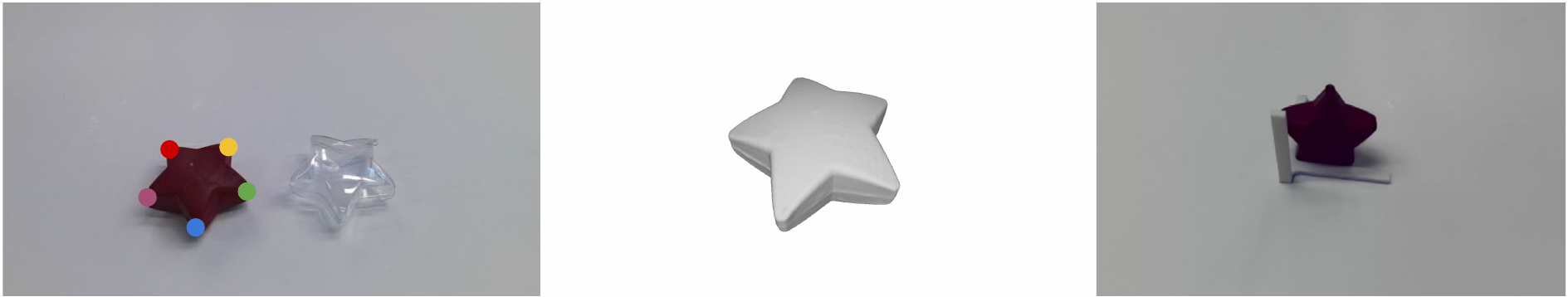} 
  }
  \\
  \subfloat{%
  \includegraphics[width=0.9\linewidth]{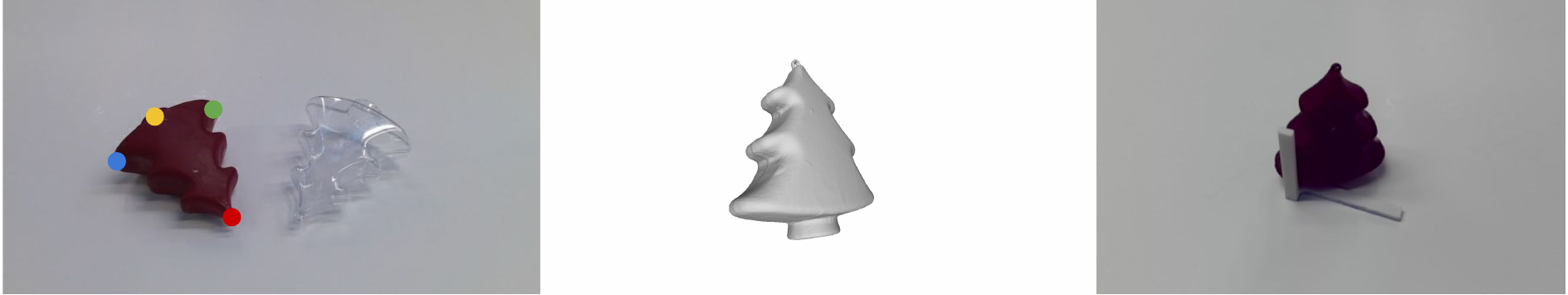} 
  }
  \caption{Visualization of star$_0$ and  tree$_0$ from top to bottom. From left to right in each row: object twin with groundtruth keypoint location, scanned CAD model from opaque object, and aligning the marker to the object. We use \textcolor{viz_red}{red}, \textcolor{viz_yellow}{yellow}, \textcolor{viz_green}{green}, \textcolor{viz_blue}{blue} and \textcolor{viz_purple}{purple} to mark keypoint \textcolor{viz_red}{1}, \textcolor{viz_yellow}{2}, \textcolor{viz_green}{3}, \textcolor{viz_blue}{4} and 
  \textcolor{viz_purple}{5}. For star$_0$, all its five keypoints are symmetric. For tree$_0$, keypoints \textcolor{viz_green}{3} and \textcolor{viz_blue}{4} are symmetric.}
  \label{fig:viz:object:4}
\end{figure*}

\textbf{Objects Used in Our Dataset.}
Our complete dataset consists of 20 object pairs in total, though only 15 object pairs are used in the experiments of the main paper. We illustrate them and the keypoint groundtruth definition in the first column of Figure \ref{fig:viz:object:1}, \ref{fig:viz:object:2}, \ref{fig:viz:object:3} and \ref{fig:viz:object:4}. We also scan the opaque objects and provide the 3D CAD model of the objects for applications that require them. The CAD models of the objects and the alignment of the markers to the objects are also illustrated in Figure \ref{fig:viz:object:1}, \ref{fig:viz:object:2}, \ref{fig:viz:object:3} and \ref{fig:viz:object:4}.

\begin{figure*}[h]
  \centering
  \subfloat{%
  \includegraphics[width=0.195\linewidth]{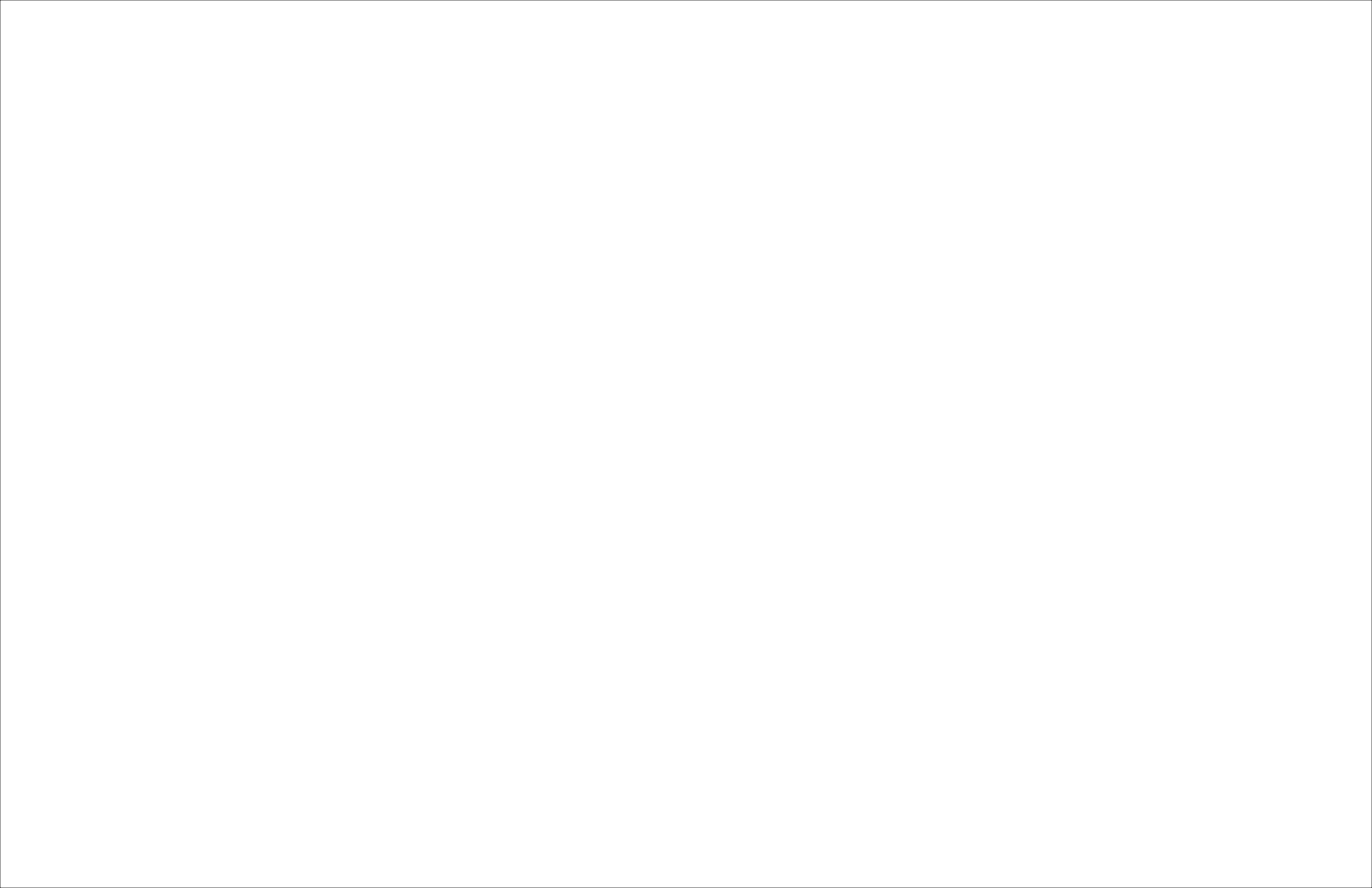} 
  }
  \subfloat{%
  \includegraphics[width=0.195\linewidth]{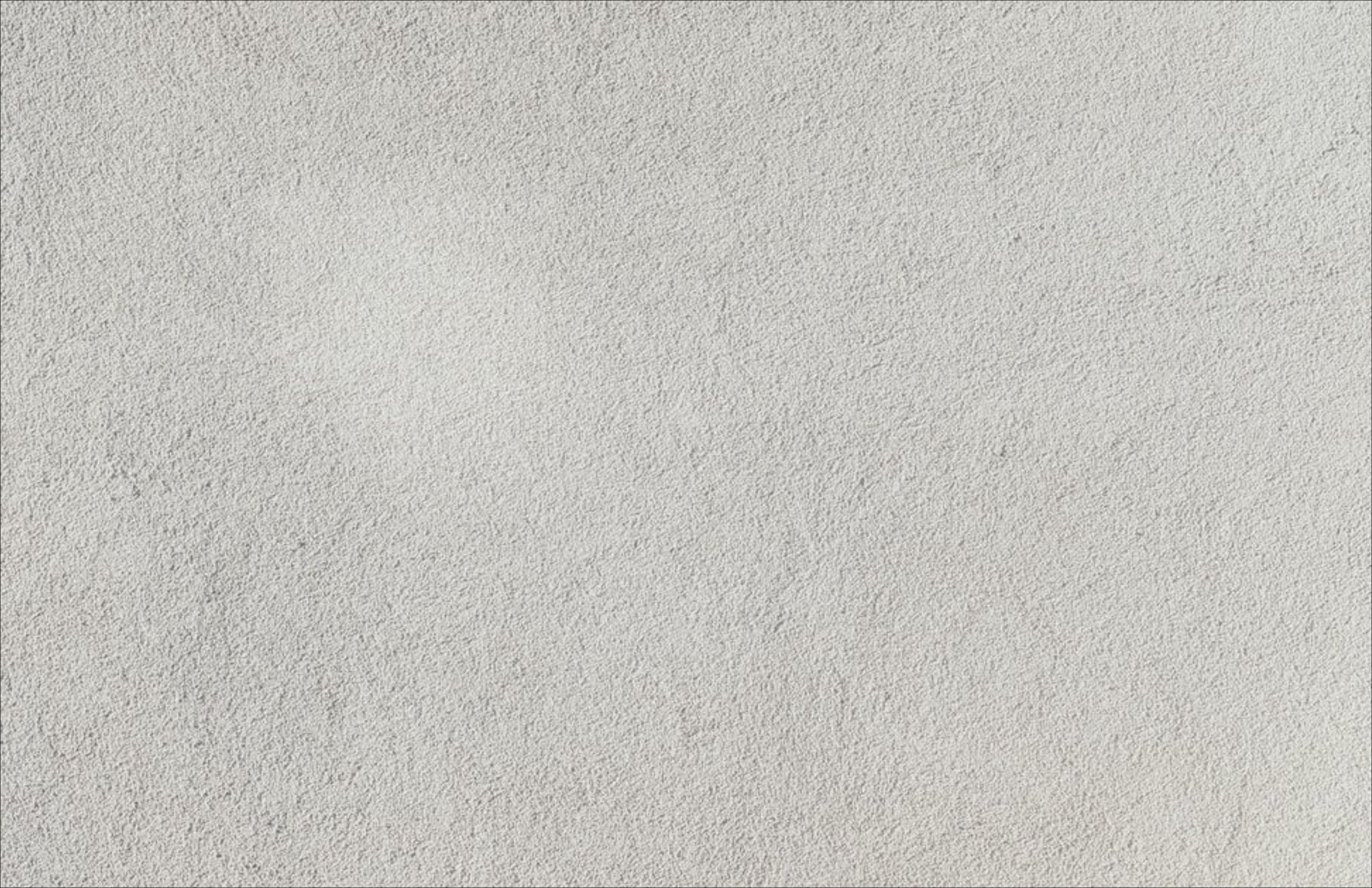} 
  }
  \subfloat{%
  \includegraphics[width=0.195\linewidth]{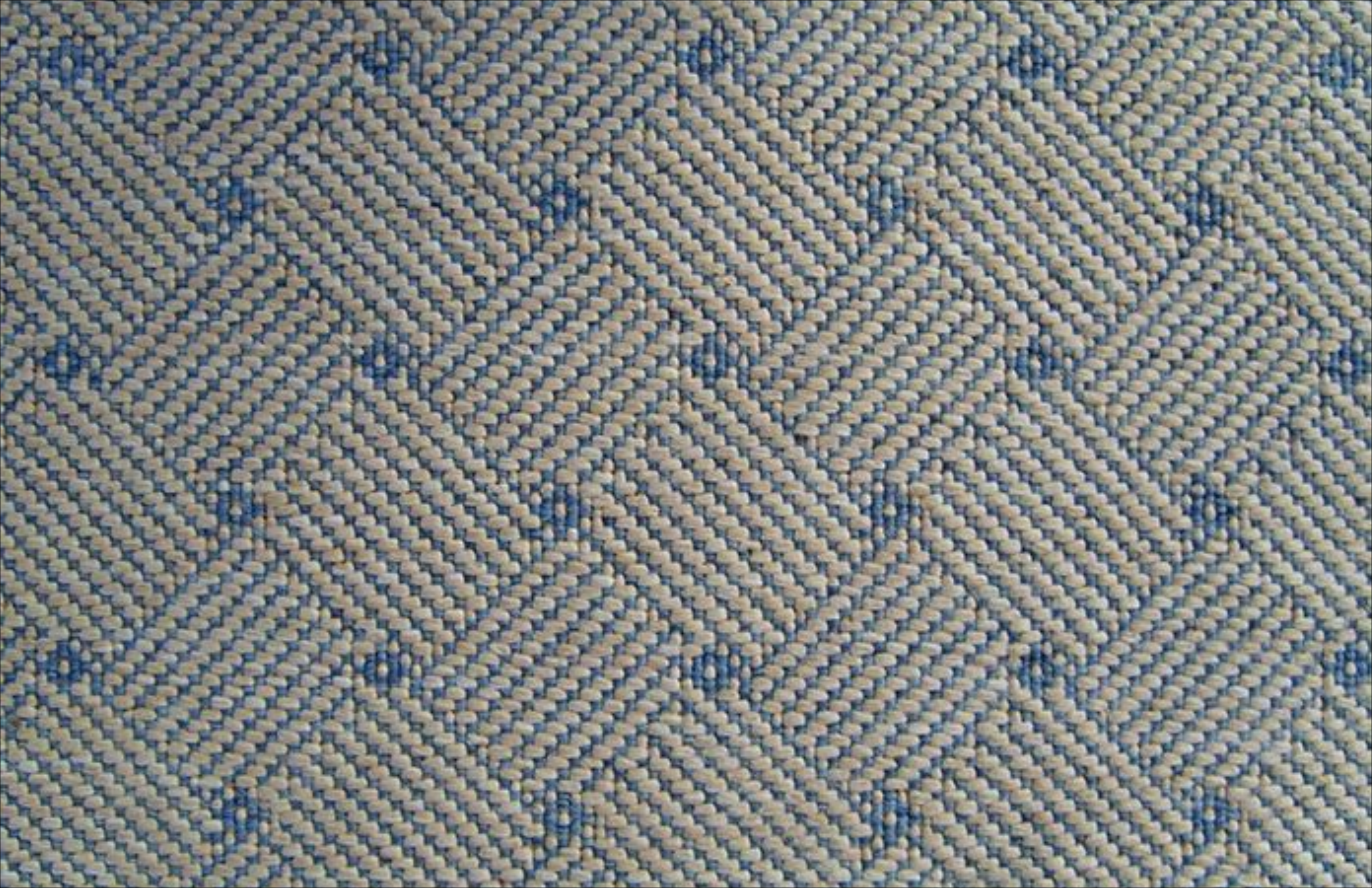} 
  }
  \subfloat{%
  \includegraphics[width=0.195\linewidth]{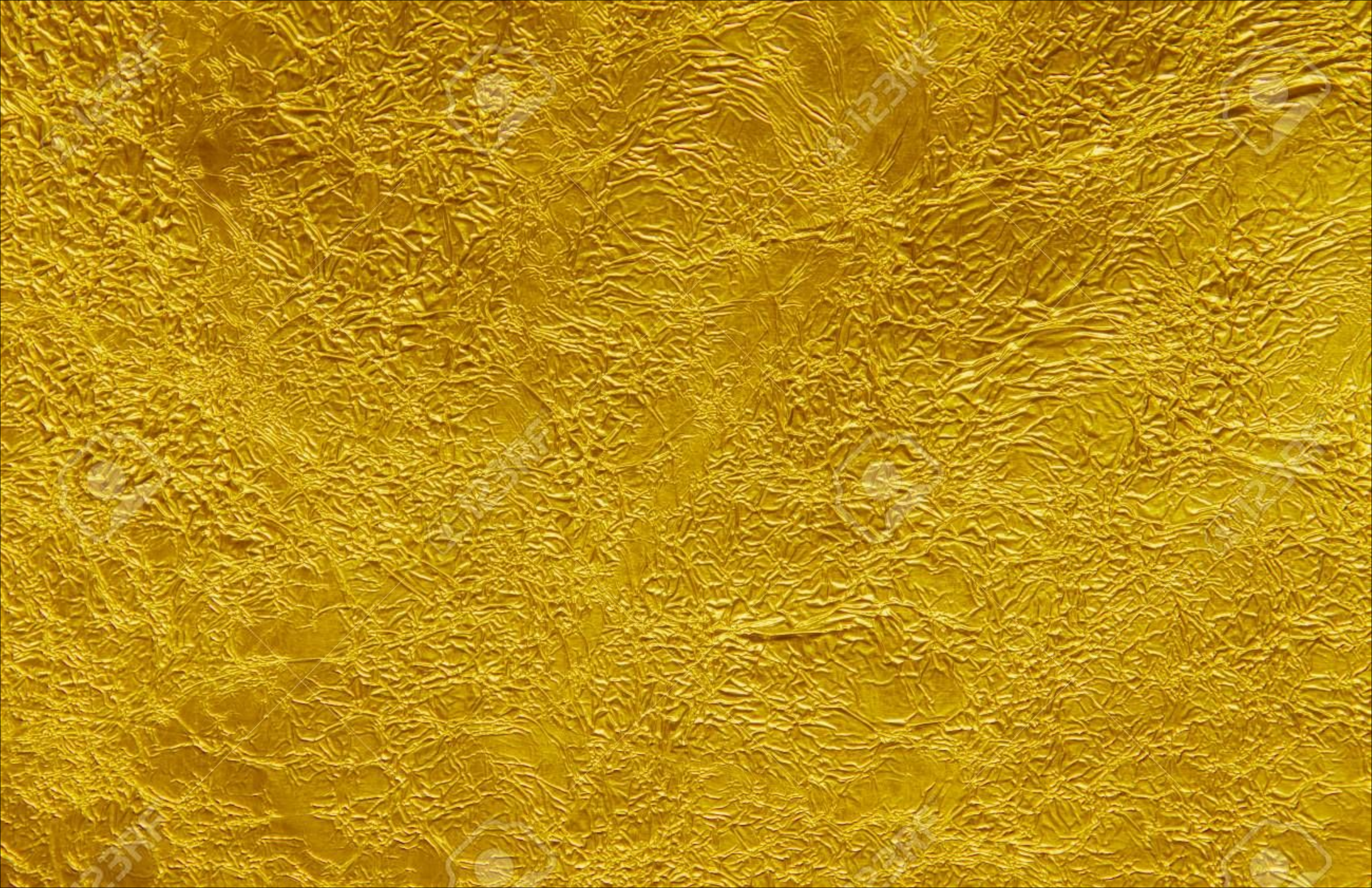} 
  }
  \subfloat{%
  \includegraphics[width=0.195\linewidth]{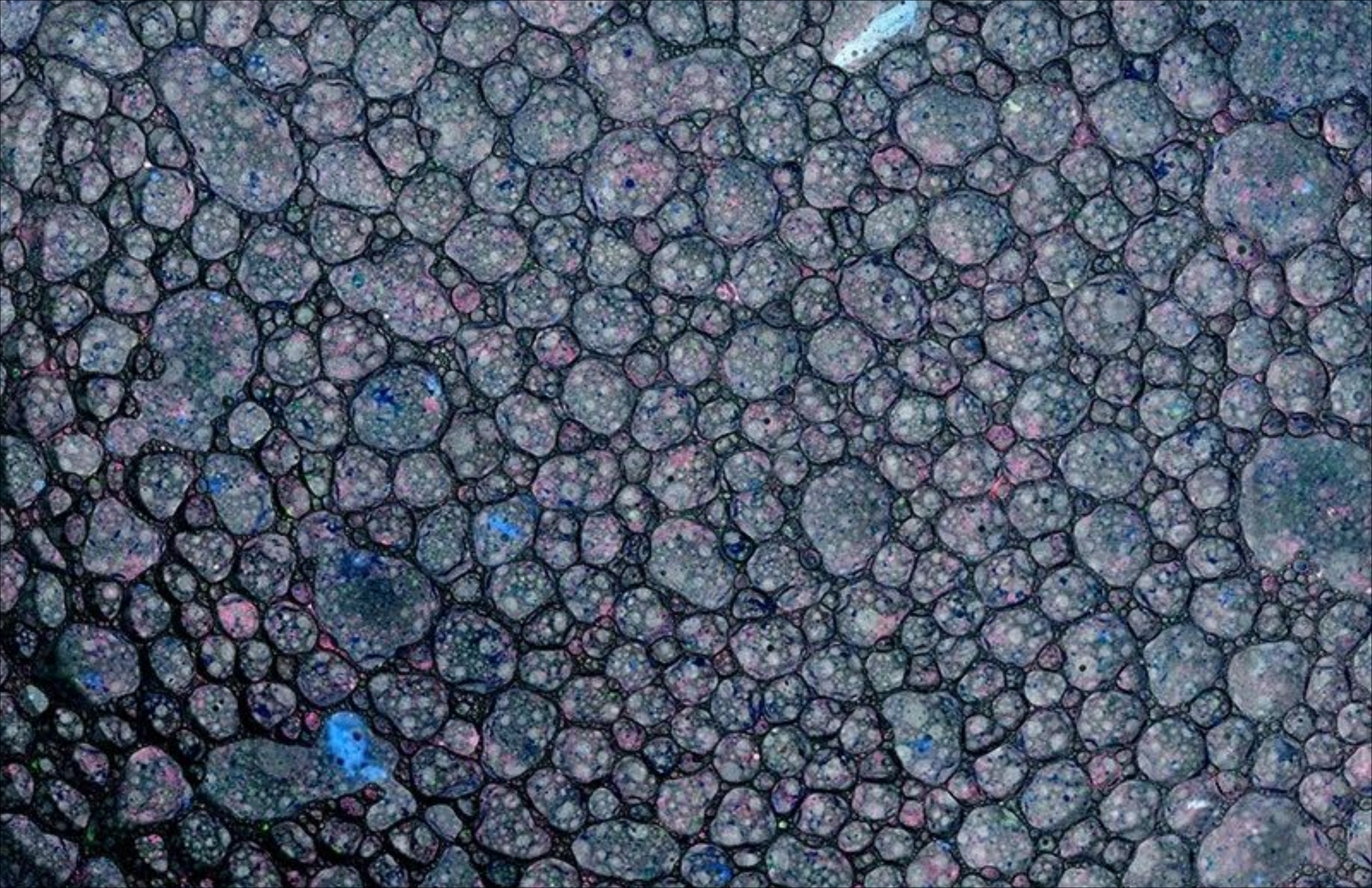} 
  }
  \\
  \subfloat{%
  \includegraphics[width=0.195\linewidth]{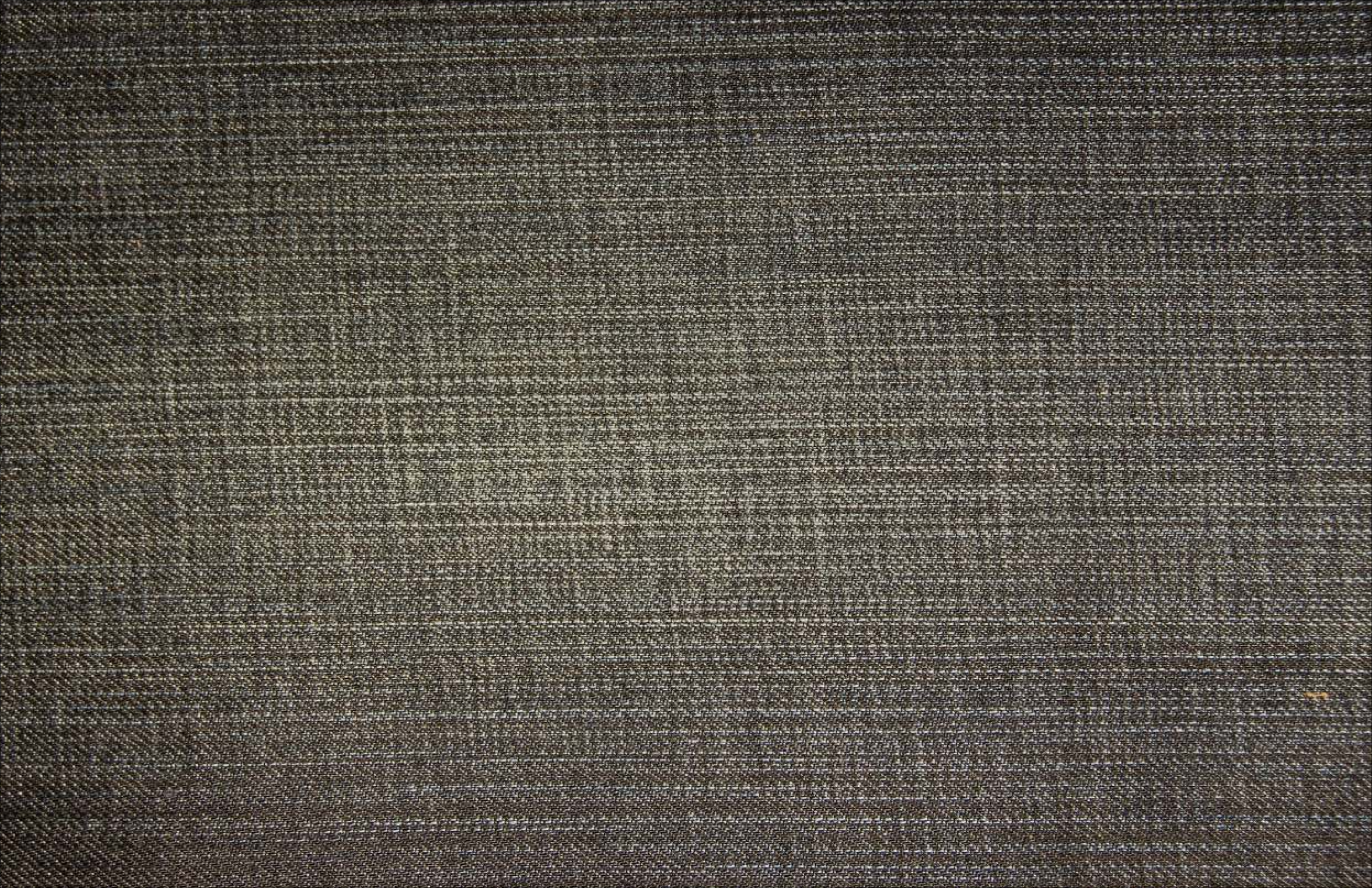} 
  }
  \subfloat{%
  \includegraphics[width=0.195\linewidth]{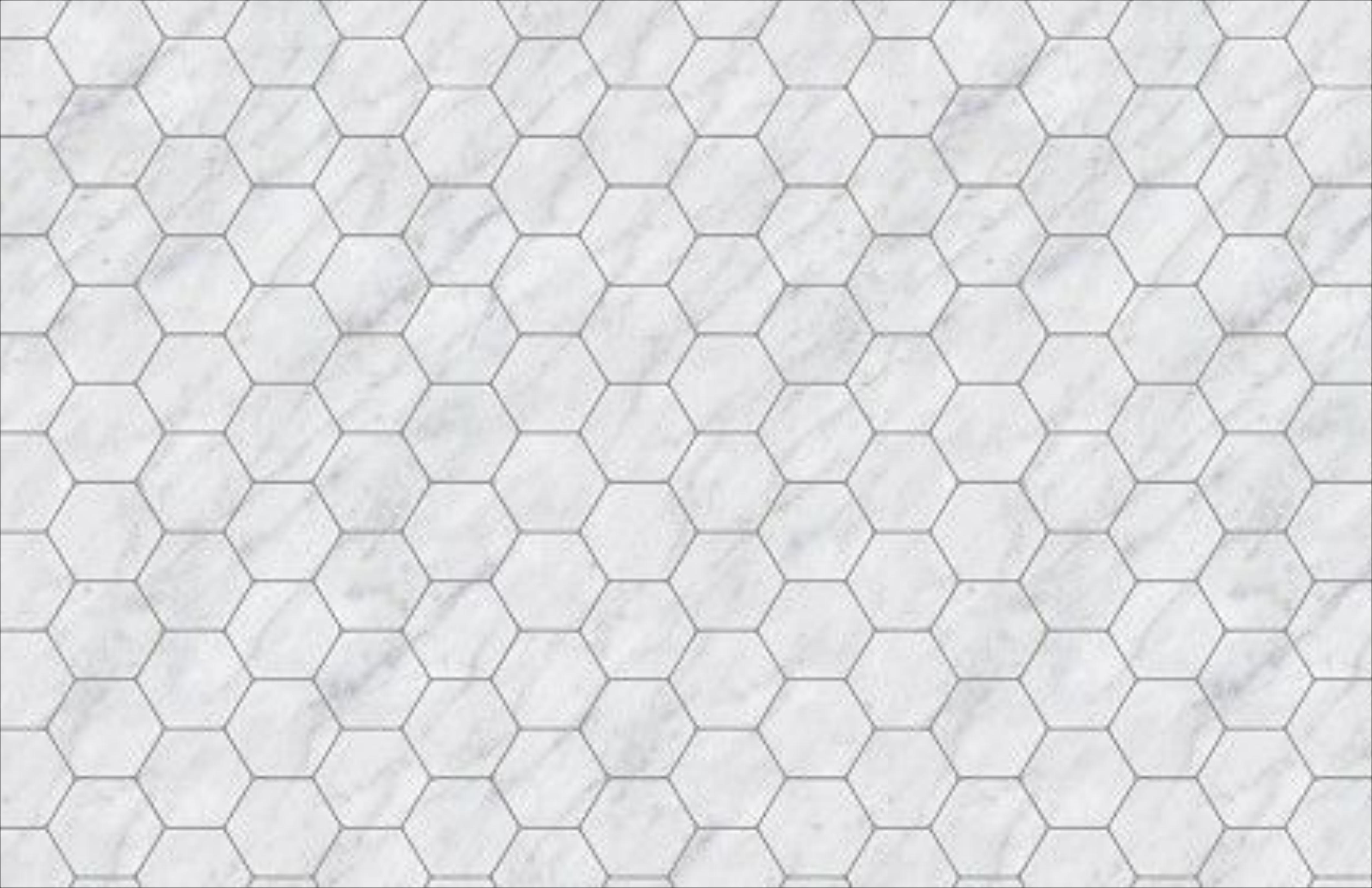} 
  }
  \subfloat{%
  \includegraphics[width=0.195\linewidth]{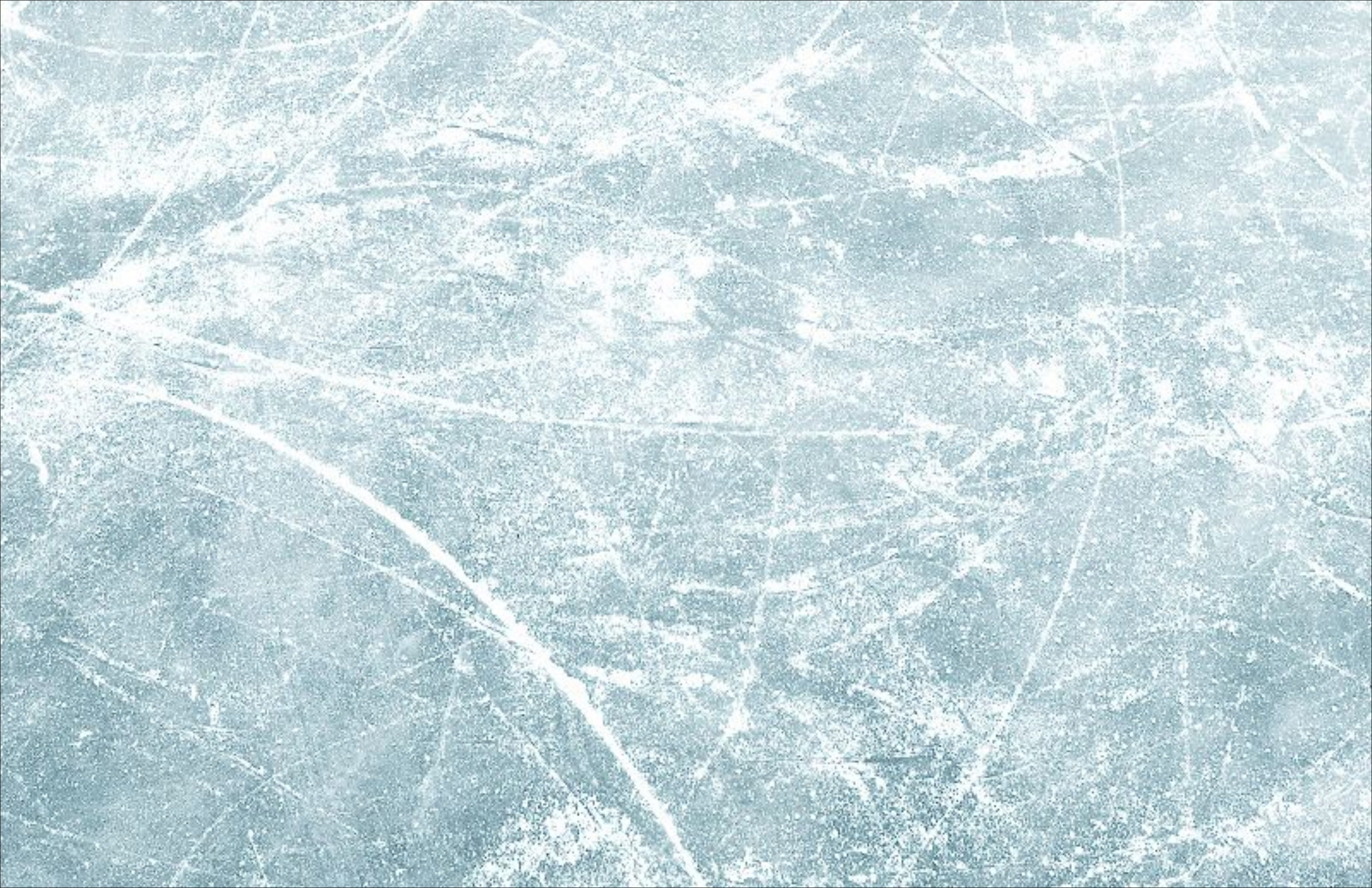} 
  }
  \subfloat{%
  \includegraphics[width=0.195\linewidth]{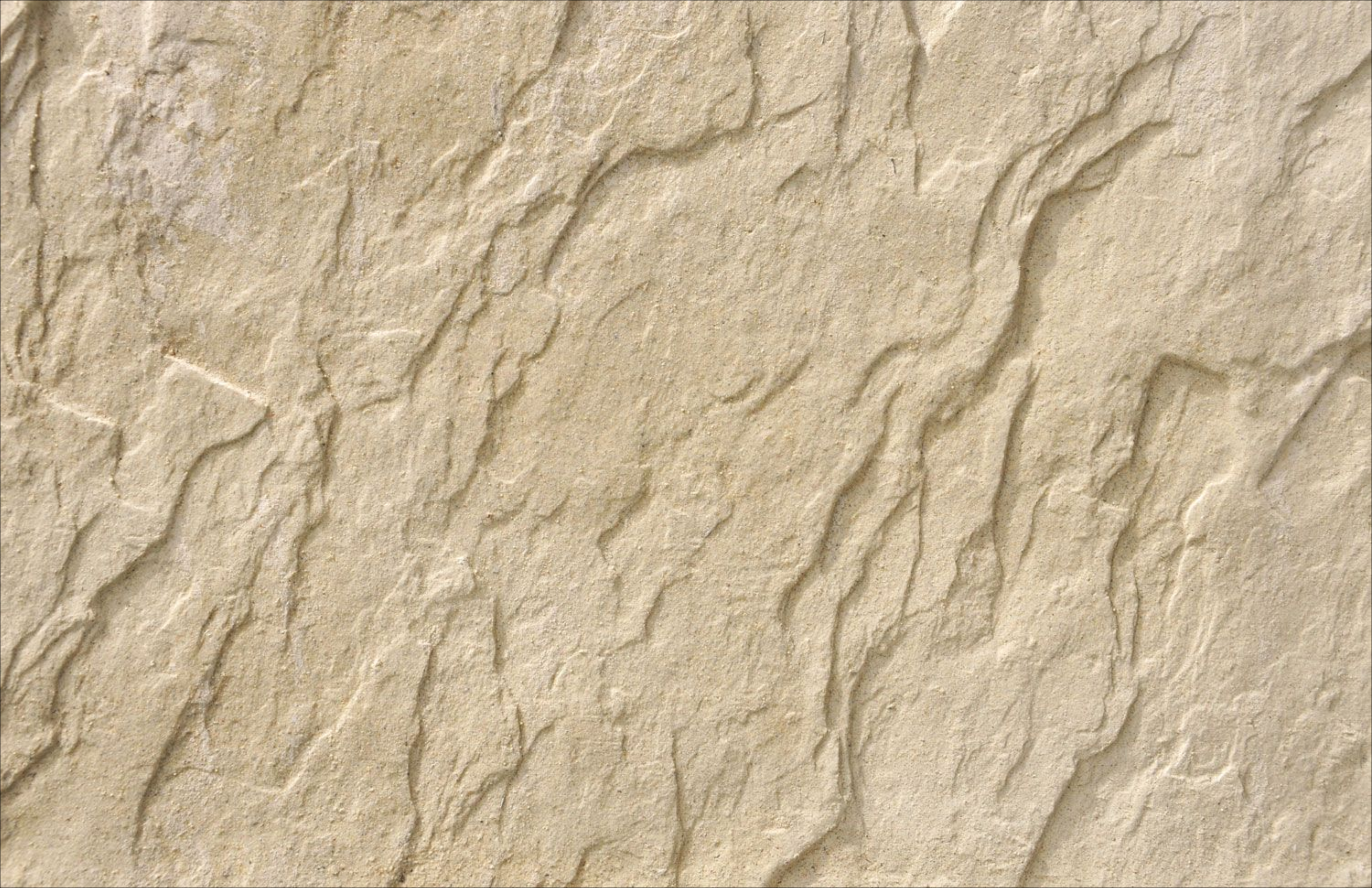} 
  }
  \subfloat{%
  \includegraphics[width=0.195\linewidth]{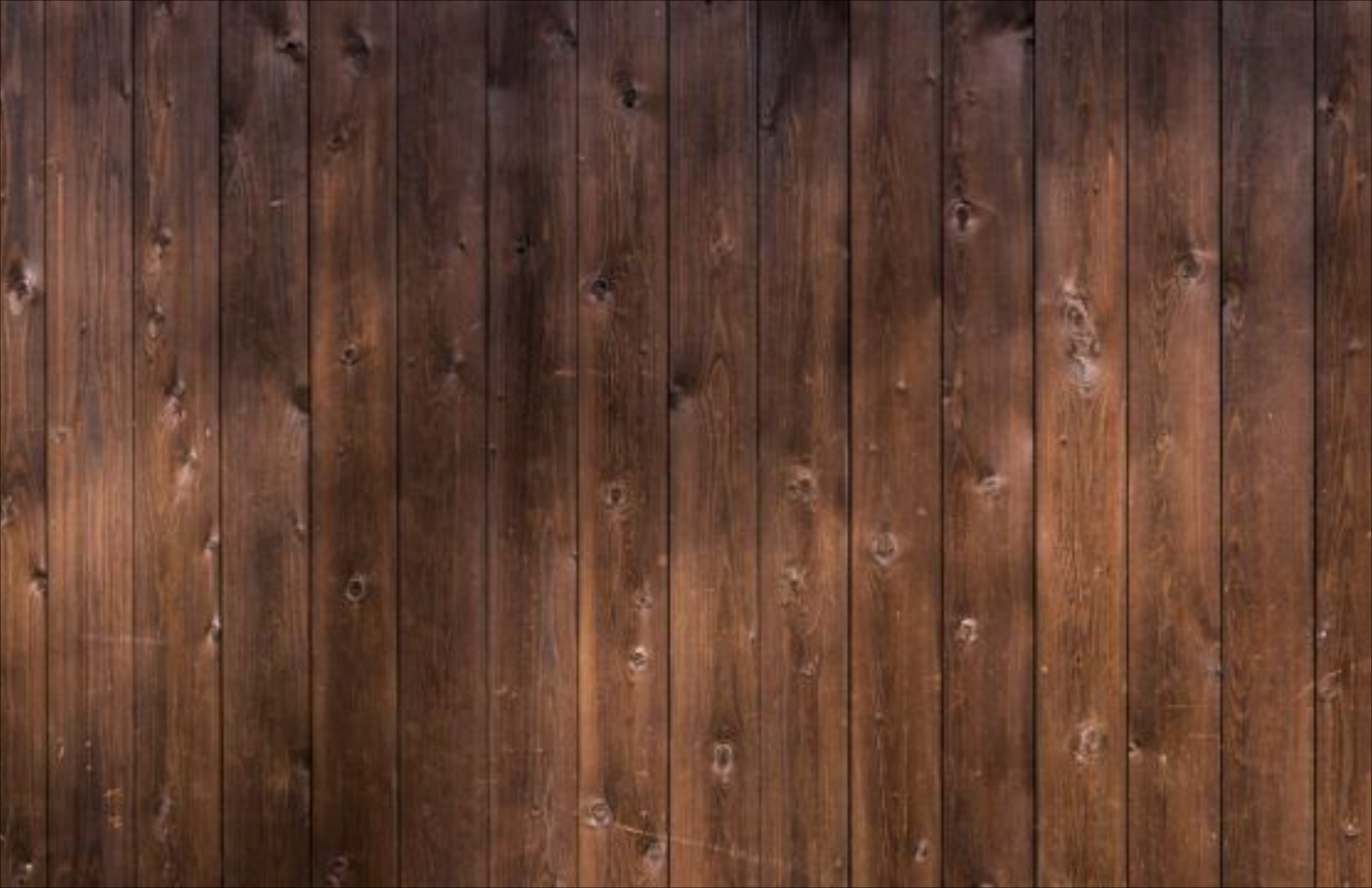} 
  }
  \caption{Background textures used in the dataset.}
  \label{fig:texture}
\end{figure*}

\textbf{Textures Used in Our Dataset.} 
In our dataset, each object in placed on ten diverse background textures, which are illustrated in Figure \ref{fig:texture}. We print the textures on papers and place them beneath the objects. The textures include pebbles, rocks, woods, textile etc.

\section{Data Capture and Error Analysis}
\label{sec:data:capturing}

In this section we provide more detail about the data capture pipeline, as well as an analysis of pose estimation and 3D keypoint errors. The pipeline for capturing and labeling a single object is illustrated in Figure \ref{fig:labeling:pipeline}.

\subsection{Data Capture Pipeline}

Data is captured via a sensor head attached to a Franka Panda arm, illustrated in Figure \ref{fig:robot:config}.  The arm is moved in a trajectory that approaches the object from 0.45 to 1 m, and traverses a solid arc of from approximately $30^\circ$ to $70^\circ$ of elevation, and $-60^\circ$ to $60^\circ$ of azimuth (see video \texttt{data\_capturing.mp4}).  The head stays approximately pointed towards the center of the planar target containing the AprilTags.

The head consists of a Stereolabs ZED stereo camera and a Microsoft Kinect Azure RGBD device.\footnote{Hardware specifications for the ZED are at \url{https://www.stereolabs.com/zed/}, and for the Kinect Azure are in \url{ https://docs.microsoft.com/en-us/azure/kinect-dk/hardware-specification}.}
We use the ZED to capture dual synchronized RGB images at 1280$\times$720 resolution, with a baseline of 0.12 m.  The camera parameters are calibrated at the factory, and correct for distortion and stereo geometry, with the rectified stereo pair having horizontal epipolar lines.  The FOV for the camera is 90(H)$\times$60(V) degrees, fairly wide angle, which introduces perspective distortion at the edges of the images; many other datasets use narrower FOV to avoid this distortion, but we think it is important for the model to deal with it.

The Kinect Azure device is mounted just above the ZED, so the lenses are as close as possible.  It consists of an RGB camera, and a time-of-flight depth camera offset by about 2cm.  We operate the depth camera in wide-angle mode, 120(H)$\times$120(V) degree FOV, with a resolution of 1024$\times$1024.  For the RGB camera, which is synchronized with the depth camera, we capture images at 1280$\times$960, and the FOV is 90(H)$\times$59(V).  As with the ZED, distortion and geometry parameters are calibrated at the factory.

Note that the depth camera has several sources of error, including up to 11 mm of systematic error, and a random error standard deviation of 17 mm.  Additionally, multipath interference, especially in corners, can lead to larger distortions; and low-angle incidence often causes dropouts.  Figures 1 (in the main paper) and 
%\ref{fig:teaser}
\ref{fig:viz:warp} show examples of the depth sensor output.
The two devices are not time-synchronized; alignment of depth to the stereo images is done with the method described in Subsection \ref{subsec:warping}.

Each object is placed in various positions on a background on the planar target board (Figure \ref{fig:viz:board}) and the robot is activated, capturing a video of some 400 images in stereo and 200 images in RGBD (slower frame rate).  Then, the opaque twin is substituted as shown in Figure~\ref{fig:replacement}, and another scan is completed to capture opaque depth.  We save all stereo and RGBD images from these scans, to be processed as described below.

\subsection{Camera Pose Estimation}

To correspond the different camera views, we determine poses from the images of AprilTags on the target board.  First, the image coordinates of the corners of the ArpilTags are extracted using publicly-available software from the University of Michigan\footnote{\url{https://april.eecs.umich.edu/apriltag/}.}.  Given the known 3D positions of the tags on the board, the PnP algorithm of OpenCV is used to compute the camera pose relative to the frame of the target board.

The board contains eight AprilTags along the borders (Figure~\ref{fig:viz:board}).  We reject any images in which fewer than 3 tags are correctly detected.  The mean number of detected tags on a trajectory is 6.1 for the left stereo camera, and 5.9 for the RGB camera of the Kinect (it has a smaller FOV), assuring a robust estimation of the pose.  In reprojecting the target AprilTag points back to the camera at its estimated pose, we can compute the RMSE in pixels for the pose estimation over a trajectory.  Doing this for all 600 trajectories yields the statistics in Table \ref{tab:pose:stats}, with the average RMSE at 1.21 pixels for the left stereo camera and 1.30 pixels for the Kinect.  We use these values in analyzing the errors in depth warping and 3D keypoint estimation.
\begin{table}[t]
    \centering
    \begin{tabular}{c|c|c}
         reprojection error (px) & RMSE mean & RMSE std \\
         \hline
      left stereo & 1.21 & 0.51 \\
      Kinect RGB & 1.30 & 0.387
    \end{tabular}
    \caption{Reprojection errors for camera pose estimates (pixels).  RMSE is computed over a trajectory; average and standard deviation over a set of 600 trajectories.}
    \label{tab:pose:stats}
\end{table}

\subsection{Depth Image Warping}

\newcommand{\trans}[2]{{$^\mathit{#1}T_\mathit{#2}$}}
\label{subsec:warping}

\begin{figure}[t] 
  \centering
  \subfloat{%
  \includegraphics[height=0.44\linewidth]{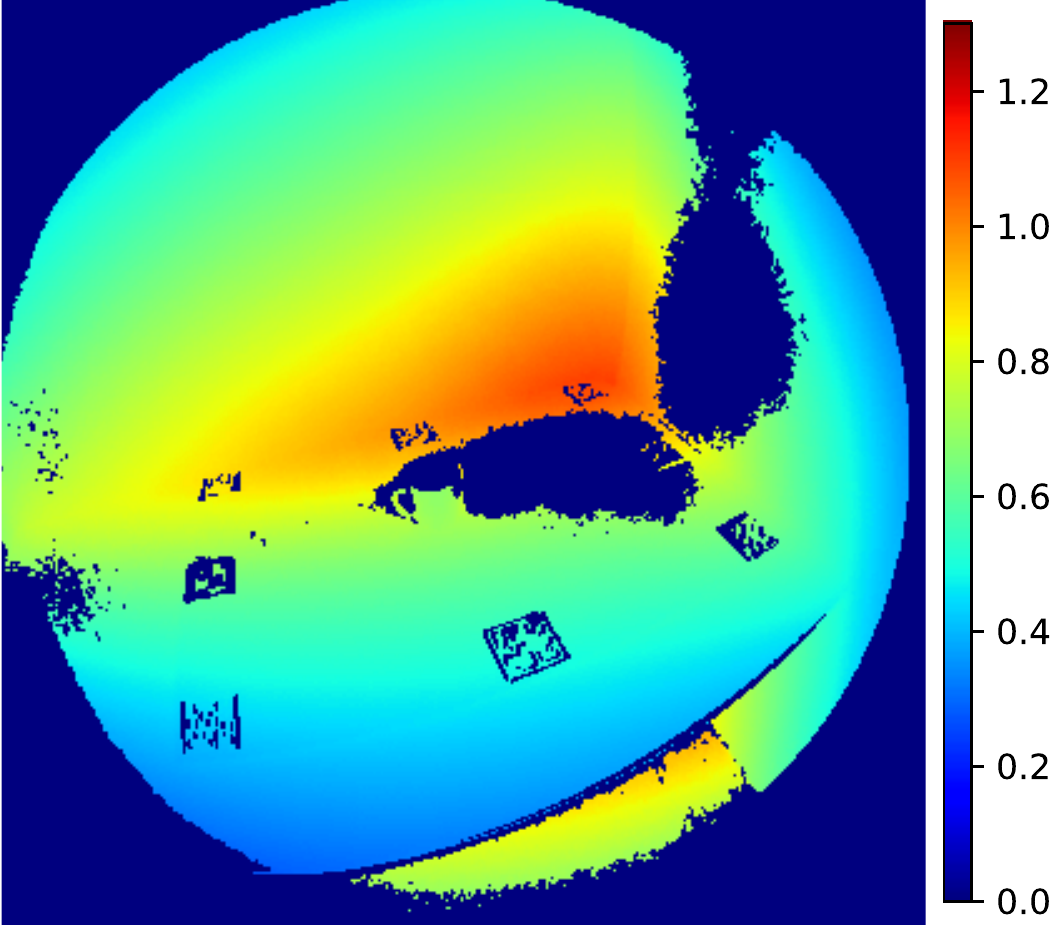} 
  }
  \subfloat{%
  \includegraphics[height=0.44\linewidth]{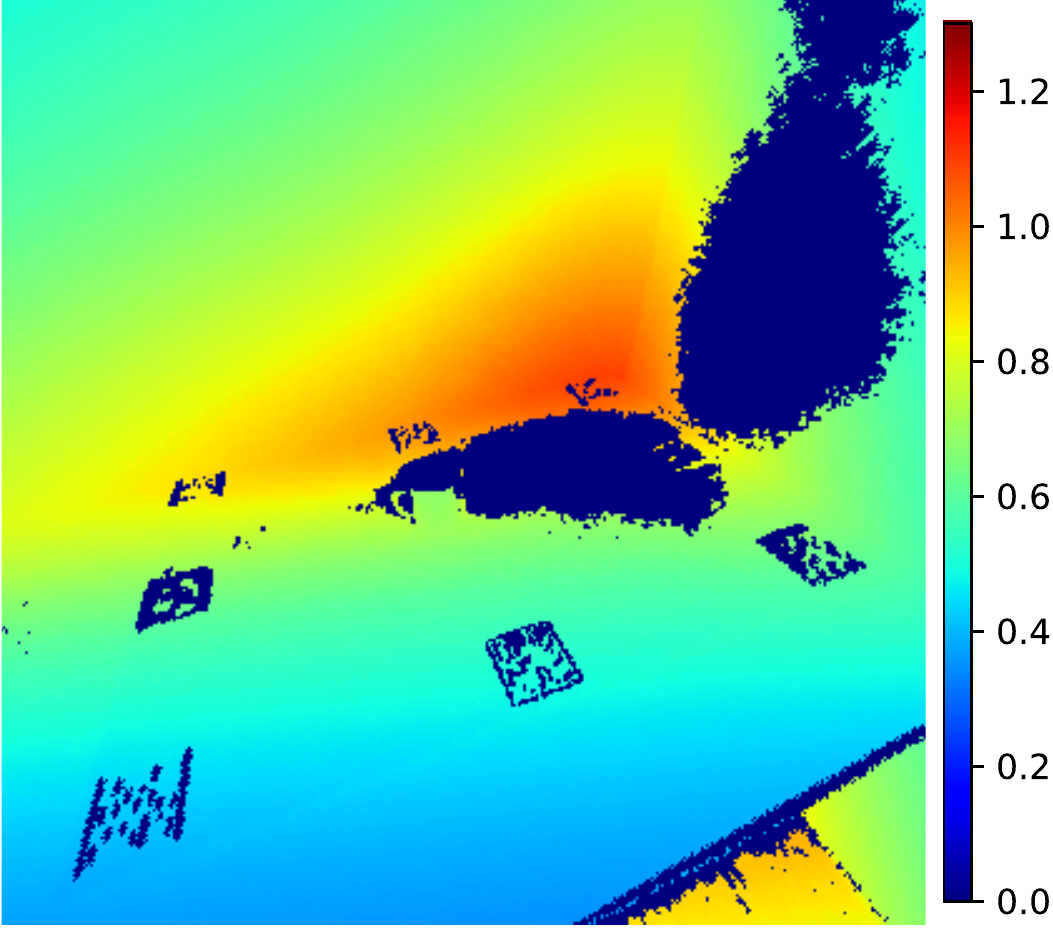} 
  }
  \\
  \subfloat{%
  \includegraphics[height=0.3093\linewidth]{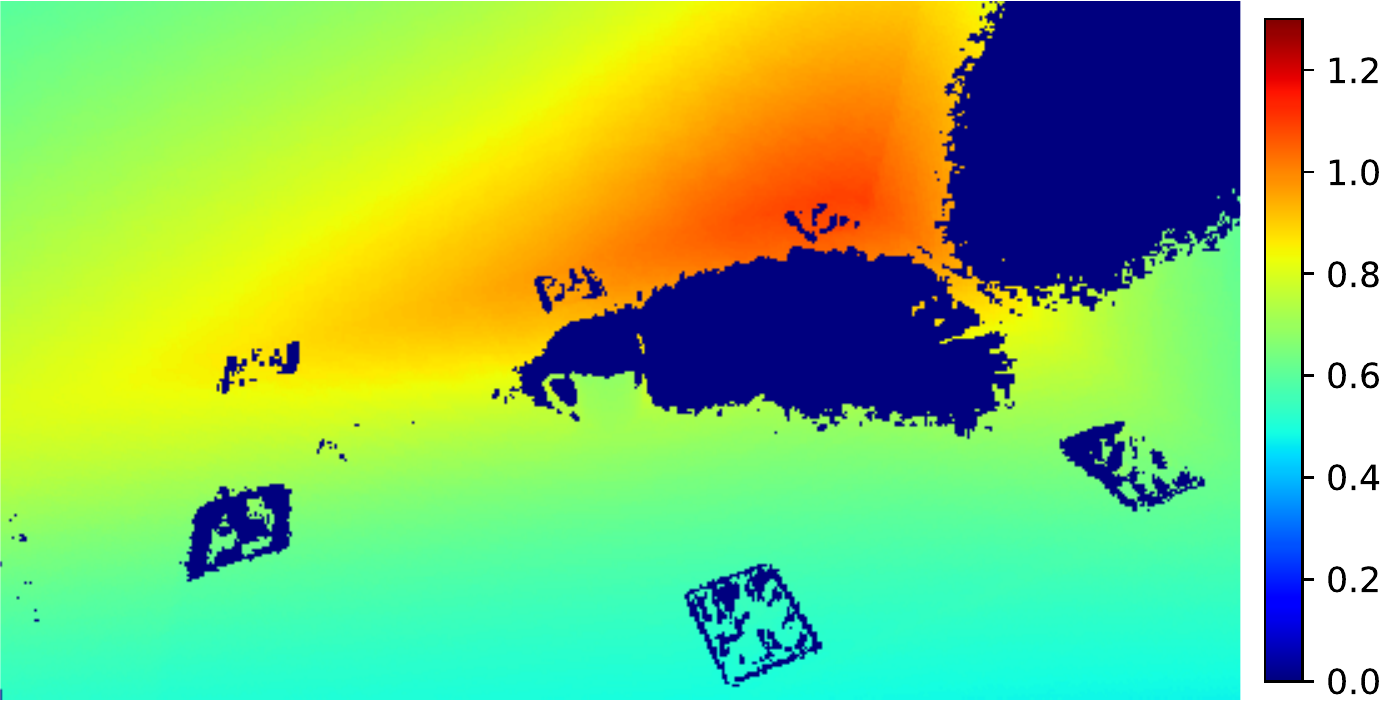} 
  }
  \caption{Left: raw depth image. Right: rectified depth image. Bottom: depth image warped to left stereo image.} 
  \label{fig:viz:warp}
\end{figure}

\begin{figure}[t] 
  \centering
  \includegraphics[width=0.95\linewidth]{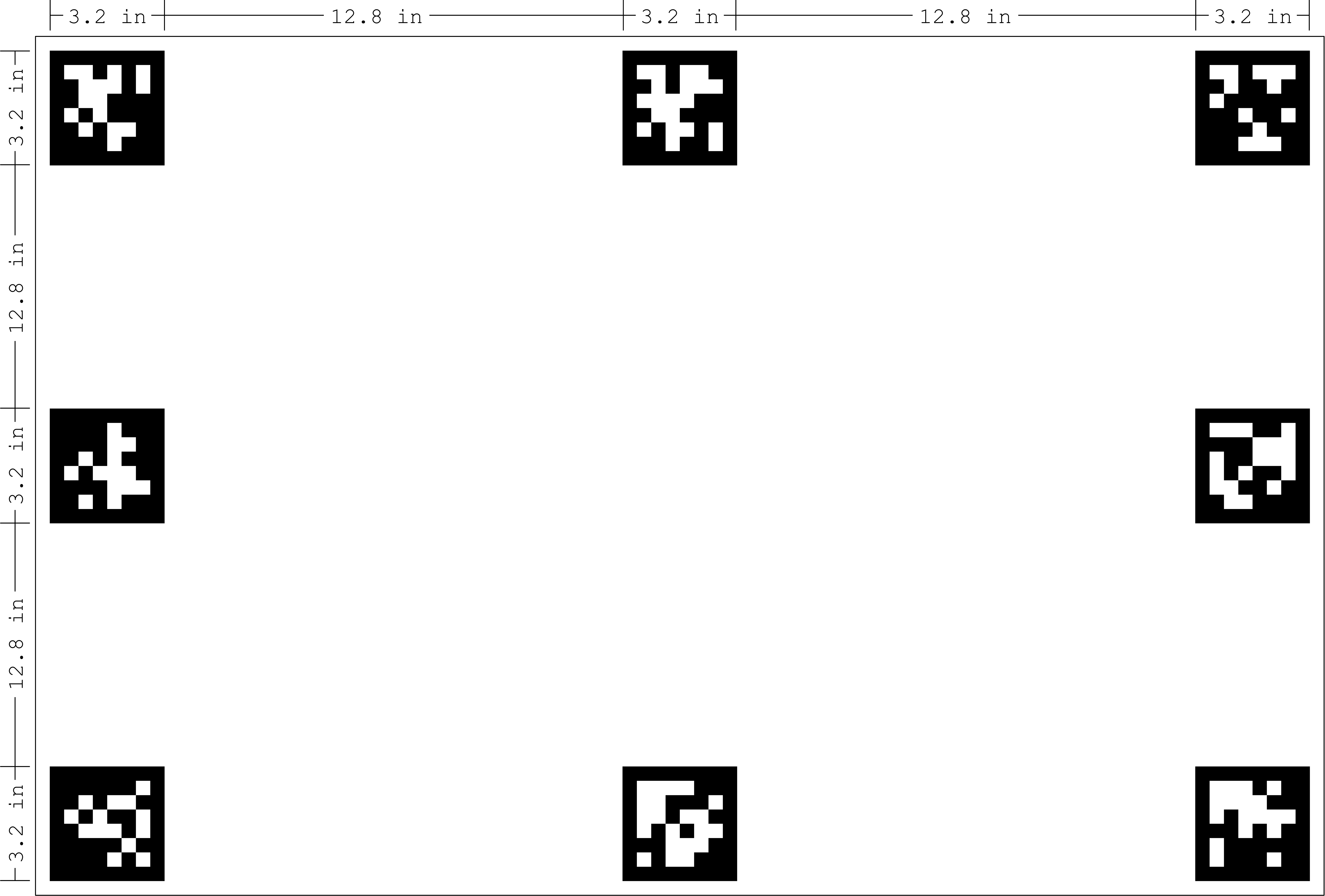} 
  
  \caption{AprilTag board used for capturing data. The positions of AprilTags are fixed therefore global pose can be calculated.} 
  \label{fig:viz:board}
\end{figure}

Since the depth images are acquired from a different viewpoint than the stereo images, they must be warped to register with the latter (we use the left stereo image as the reference image).  The depth image, along with the depth camera parameters, gives a 3D point for every pixel, which can then be transformed to a different camera frame and reprojected to form a new depth image registered with that camera.
There are several steps to warping the depth image:
\begin{enumerate}
\item Remove distortion.  Convert 1024$\times$1024 depth image to a 1024$\times$1024 depth image of an ideal pinhole camera.  This is a standard image warping operation; we use OpenCV's \texttt{undistort} function with the factory-provided calibration parameters.
\item Find the nearest viewpoint.  Since the devices are not synchronized, and the images could have been captured on different scans (opaque vs.\ transparent object), we find the left stereo viewpoint that is closest to the depth image viewpoint.  Since the cameras are all registered to a common world view, the target board, it is possible to do this.  The viewpoints should be close in both position and orientation.  To achieve this, we look at two 3D points, 1 meter ahead of and behind the depth camera.  The left stereo camera whose similar points are closest in the world frame is chosen.
\item Compute transform chain.  There are three relevant poses: depth camera (\trans{depth}{world}), depth camera to Kinect RGB camera, from its known calibration (\trans{rgb}{depth}), and left stereo camera (\trans{left}{world}).  The transform from the depth camera to the left stereo camera is the chain:
 \trans{left}{world} \trans{world}{rgb} \trans{rgb}{depth}. 
\item Warp depth to left image.  Each point in the rectified depth image is converted to a 3D point in the camera frame, then transformed to the left stereo camera frame via the above transform.  Then it is projected onto an image with the same camera parameters as the left image.  Z-buffering assures that closer points overlay further ones.  We also interpolate pixels in the original depth image to eliminate quantization holes in the transformed image.
\end{enumerate}
The results are shown in Figure \ref{fig:viz:warp}.

We can get an idea of the pixel errors in warping the depth data from the camera pose estimation errors (Table \ref{tab:pose:stats}).  The combined RMSE is 1.78 pixels ($ = \sqrt{1.21^2 + 1.30^2}$).  There are additional errors caused by the transform \trans{rgb}{depth}, which we assume to be sub-pixel from factory calibration, and hence small relative to pose estimation error.  The error in depth of the Kinect camera will also result in a projection error, but again, if the left stereo camera and the depth camera viewpoints are close, these should be small and we ignore them.

\subsection{3D Keypoint Labelling and Error Analysis}

Given the pose of multiple cameras looking at the same object on the target in a scan, we can compute 3D keypoints by labeling them on a subset of the 2D images.  Using a Farthest Point Sampling (FPS) algorithm \cite{fps:1985,fast:fps}, we pick 6 images that are farthest from each other in position on the scan.  For each image, we label the keypoints in the image.  These are the projections of the 3D keypoints on the images.  Since we know the camera parameters, a simple least-squares nonlinear estimation finds the 3D keypoints that minimize the squared reprojection errors.  
Once the keypoints are estimated, they are projected back to the labeled views to find the RMSE in pixels.  Any scan that has a keypoint error of more than 5 pixels is rejected.  We collected RMSE statistics for the mug$_2$ object: over all scans, the mean RMSE was 2.28 pixels, and the standard deviation was 0.83 pixels.
Gathering a single scan and labeling it takes about 10 minutes of user work, so it is possible to acquire large sets of real-world data.

How accurate are the keypoints that are computed from 2D annotations?  Unfortunately it is difficult to answer this analytically, because they are computed from two nonlinear optimizations: camera view pose estimation and 3D point estimation from multiple views.  Instead, we use Monte Carlo simulation to run thousands of scenarios that conform to the reprojection error statistics that we gathered for camera and keypoint pose estimates.
In each simulation, we randomly chose 4 to 6 views taken from the Farthest Point Sampling (FPS) \cite{fps:1985,fast:fps} of poses, and calculated the AprilTag corner projections.  We then dithered these projections according to the statistics in Table \ref{tab:pose:stats}, and re-calculated the poses to get estimated poses.  Then, we randomly placed a 3D keypoint in the workspace, and projected it onto the estimated poses.  Finally, we dithered these projections according to the keypoint RMSE statistics, and estimated the keypoint.  The metric distance between the estimated and ground-truth keypoints gives an error measure.  We did this for 10,000 simulations, and calculated the RMSE as 3.4 mm. We compare this to the depth error of Microsoft Azure Kinect in Table \ref{tab:kp:error}, and conclude that our method is \textbf{at least five times more accurate} than the estimation from depth sensors.

\begin{table}[t]
\small
\centering
% \resizebox{\textwidth}{!}{
\begin{tabular}{c|c}
method & 3D error (mm) \\
\hline
Microsoft Azure Kinect depth \tablefootnote{Depth error data for the Microsoft Azure Kinect can be found in \url{ https://docs.microsoft.com/en-us/azure/kinect-dk/hardware-specification}.} & 17 \\

\textbf{Our keypoint labelling} & \textbf{3.4} \\
\end{tabular}
% } 
\vspace{-1ex}
\caption{ Comparison of keypoint labelling error. }
\label{tab:kp:error}
% \vspace{-2ex}
\end{table}

\section{Architecture and Training Details}
\label{sec:arch}

\subsection{Keypose Architecture}

As noted in the paper, the KeyPose architecture is derived from KeyPointNet \cite{keypointnet}.  
Table \ref{tab:arch} lists the layers and their parameters.   

Stereo RGB images at a resolution of 180$\times$120 are stacked and fed into a set of exponentially-dilated 3$\times$3 convolutions \cite{dcnn} that expands the context for predicting keypoints, while keeping the resolution constant.  The exponential series expands the context for each pixel to an area of 64$\times$64 \cite{dcnn}.  Repeating this sequence twice ensures an even wider context.
After each dilated convolution, we apply batch normalization followed by leaky  RELU activation (alpha of 0.1).  We also insert $L2$ regularization with a factor of 0.001.

\uvd\ coordinates are extracted either by direct regression to numeric values, or with an integral image.  For direct regression, we add two 1x1 convolutions with 64 features and $L2$ regularization, again followed by batch normalization and leaky RELU activation.  Then, we have one 1x1 convolution with $3N$ features, where $N$ is the number of keypoints.  The output of this convolution is taken directly as the \uvd\ values for the keypoints.

For the integral image technique, we add a 3x3 convolution with $N$ features and $L2$ regularization.  The output is processed by a spatial softmax to convert it to a probability, and then integrated to find the centroid (and hence the \textit{UV} coordinates), as in IntegralNet \cite{integral:net}.  A disparity heatmap is computed by a 3x3 convolution with $N$ features and $L2$ regularization, convolved with the probability map, and the centroid predicts the disparity.

We experimented with various other architectures, including UNet \cite{unet} and adding an explicit correlation operator.  These did not do better than the dilated CNN.

\begin{table}[t]
\small
\setlength{\tabcolsep}{4.5pt}
\centering
% \resizebox{\textwidth}{!}{
\begin{tabular}{c|c|c|c|c}
\hline
layer \# & kernel size & dilations & stride & \# of channels \\
\hline
1--7 & 3 & 1,1,2,4,8,16,32 & 1 & 48 / 64\\
8--15 & 3 & 1,1,2,4,8,16,32 & 1 & 48 / 64\\
prob & 3 & 1 & 1 & $N$ \\
disp & 3 & 1 & 1 & $N$ \\
regress & 1 & 1,1,1 & 1 & 64, 64, $3N$ \\
\hline
\end{tabular}
% } 
\vspace{-1ex}
\caption{Architecture of Keypose Early Fusion.  The number of channels is 48 for instance models, and 64 for category models.  $N$ is the number of keypoints.}
\label{tab:arch}
% \vspace{-2ex}
\end{table}

\subsection{Training}

As described in the paper, we trained with a batch size of 32 and about 300 epochs for instance training, and 200 epochs for category training, which has more training samples.  We used the ADAM optimizer in TensorFlow, with a learning rate of $1\times 10^{-3}$, and successively reducting to $5\times 10^{-6}$ by the end.  We did not do a systematic hyperparameter search, which might be able to improve results.  Based on experience with the training, we use a curriculum that introduces the projection loss when 1/3 of the training is done, and ramps it up fully by 2/3 of the training.  The coefficient for projection loss was set at 2.5; more did not improve the results, while less tended to cause higher error in the disparity.

The network has a tendency to overfit unless care is taken during training in augmenting the data.  We performed both geometric and photometric augmentations.  For geometry, rotating the view around the camera center yields a new realistic view of the object, and can be performed via 2D warping operations on the image, without knowing the 3D geometry of the scene.  However, we are constrained by stereo geometry, as we want to keep the epipolar constraints along horizontal lines, allowing the network to determine correlation between the left and right images.  Rotation around the camera $X$-axis, scaling and shear along the image $Y$-axis, and flipping the image $Y$ axis are operations that preserve this constraint.  We implemented $X$-axis rotation, using a random value in the interval $[-5^\circ, +5^\circ]$. We have not yet tried scaling and shear operations.  

\begin{figure}[t] 
  \centering
  \includegraphics[width=0.48\linewidth]{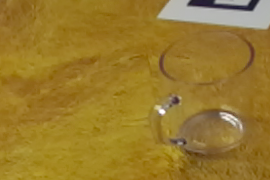}
  \includegraphics[width=0.48\linewidth]{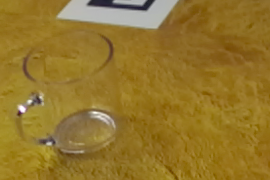}
  \\
  \includegraphics[width=0.48\linewidth]{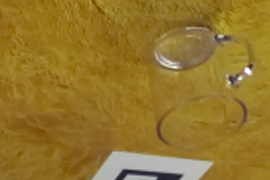}
  \includegraphics[width=0.48\linewidth]{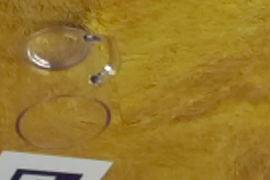}
  \\
  \includegraphics[width=0.48\linewidth]{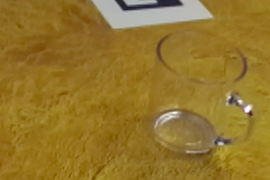}
  \includegraphics[width=0.48\linewidth]{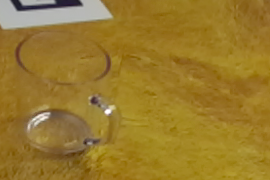}
  \\

  \caption{Mirroring stereo data.  Top: original left/right images.  Middle: Image pair when stereo camera is rotated 180 degrees.  Left and right are turned upside down and switched.  Bottom: Each image is flipped vertically to give an upright stereo pair, that looks like a mirrored version of the original, preserving epipolar geometry.} 
  \label{fig:mirror}
\end{figure}

Another transformation that doubles the size of the dataset is \textit{mirroring}.  In this operation, the right and left images are swapped, while preserving epipolar geometry.  Note that, if we rotate the stereo camera 180 degrees around the center between the two cameras, turning it upside-down, the new left camera will now see an upside-down version of the right camera image, and the new right camera an upside-down image of the left camera image.  Since we prefer to deal with upright images, we can flip the two new images vertically while preserving epipolar geometry.  Figure \ref{fig:mirror} shows a typical example.

For photometric augmentations, we used tensorflow operations to randomize hue, saturation, contrast, and brightness.  For hue, we chose a \texttt{max\_delta} of 0.1.  For saturation, the bounds were between 0.6 and 1.2.  For contrast, the bounds were between 0.7 and 1.2.  For brightness, we chose a  \texttt{max\_delta} of 32/255.  We also drop out random elliptical portions of the input images and replace them with background images.
Finally, images were normalized by subtracting a mean value and scaling by a standard deviation.  
These values were taken from ImageNet training: the mean values for RGB were $[0.485, 0.456, 0.406]$, and the standard deviation values were $[0.229, 0.224, 0.225]$.

A good measure to track during training and testing is Mean Absolute Error (MAE): the error in metric distance between the predicted and labeled 3D keypoints, in meters.  We use the average error rather than mean square error to alleviate undue influence from outliers.  
Even with all augmentations, the network will overfit, in that training MAE typically goes to around 5mm, while the testing MAE can be several times that, depending on the type of test data (instance vs. category, held-out texture vs. held-out object).  Future work will be to find models and training that generalize better.

\subsection{Pose From Keypoints}
\label{sec:model:pose}

3D keypoints are a flexible way to describe the pose of an object, although they are not always a minimal parameterization.  For example, a rigid object without symmetry has 6 DOF, while a minimum of 3 non-collinear keypoints (9 parameters) are needed.  But keypoints have the advantage of being able to describe articulated and deformable objects, such as the human hand, although we do not take advantage of that capability in this paper.

\begin{figure}[t] 
  \centering
  \includegraphics[width=1.0\linewidth]{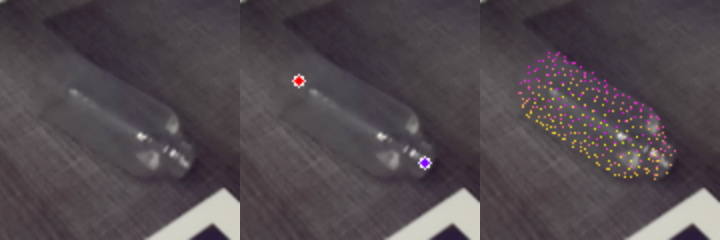} \\
  \includegraphics[width=1.0\linewidth]{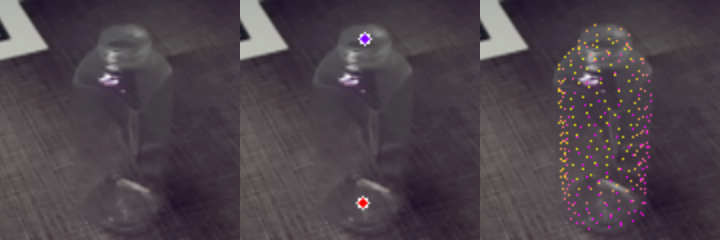} \\
  \includegraphics[width=1.0\linewidth]{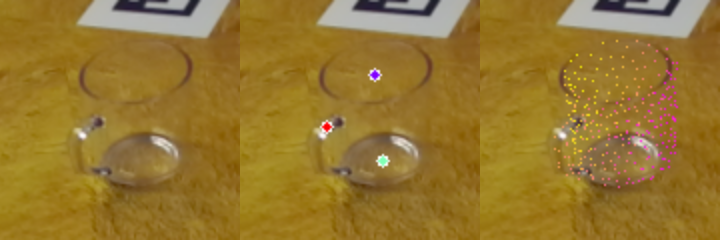} \\
  \includegraphics[width=1.0\linewidth]{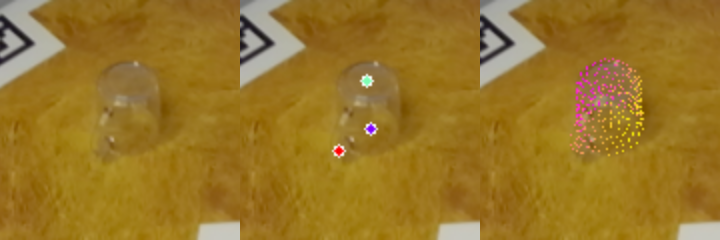}   
  
  \caption{Left: Input left image.  Middle: Predicted 3D keypoints.  Right: Projected points from the CAD model, for bottle$_2$ and mug$_2$.} 
  \label{fig:viz:pose}
\end{figure}

Since we have CAD models of the transparent objects, we can use the predicted 3D keypoints to align the models to the camera view, and project them into the image.  In the CAD models, we have labeled the 3D keypoints so they correspond to the ones labeled in the dataset.  Aligning two sets of 3D keypoints, with known correspondence, can be done using the orthogonal Procrustes algorithm \cite{procrustes}.  Figure \ref{fig:viz:pose} shows the result, for a bottle and mug.

\subsection{Model Run Time}
\label{sec:model:run:time}

The dilated CNN architecture performs inference efficiently, even though it stays at the same resolution as the input, since the number of features does not expand.  Typical runtimes for inference on a single sample, using a NVidia Titan V GPU and an i7 desktop, is 3 ms.  This does not include the time it would take to find a bounding box for a full detection and pose estimation pipeline.

\subsection{Baseline Method Details}

We use a variation of DenseFusion \cite{densefusion} as the baseline to compare with our KeyPose. There are two differences between the variation model and the original DenseFusion model. First, when extracting point clouds in the first stage, the original DenseFusion model assumes knowing the object segmentation masks in RGB images, while the variation model only assumes knowing the same rough detection bounding boxes as KeyPose in order to fairly compare with KeyPose. Therefore, the extracted point clouds are different. Second, instead of regressing 6DoF poses, the variation model directly predicts the locations of the 3D keypoints. We use similar permutation loss to train the variation DenseFusion model.

\end{document}